\documentclass[lettersize,journal]{IEEEtran}
\usepackage{hyperref}
\usepackage[hyphenbreaks]{breakurl}
\usepackage[nocompress]{cite}
\usepackage{amsmath,amssymb,amsfonts}
\usepackage{algorithm}
\usepackage{algorithmicx}
\usepackage{algcompatible}
\usepackage{graphicx}
\usepackage{textcomp}
\usepackage{xcolor}
\usepackage{microtype}
\usepackage{subcaption}
\usepackage{cleveref}
\usepackage[british]{babel}
\usepackage{tikz}
\usepackage{booktabs}
\usepackage[numbers]{natbib}
\usepackage{pifont}
\usepackage{enumitem}
\usepackage{multirow}
\usepackage{threeparttable}
\usepackage{wasysym}
\usepackage{pict2e}
\usepackage{diagbox}
\usepackage{colortbl}

\newcommand{\Foreach}[1]{\FOR{#1}}

\begin{document}
\title{\texttt{FLClear}: Visually Verifiable Multi-Client Watermarking for Federated Learning}
\author{\IEEEauthorblockN{Chen Gu\thanks{The authors are with the School of Computer Science and Information Engineering, Hefei University of Technology, Hefei, China (e-mail: guchen@hfut.edu.cn; yysun@mail.hfut.edu.cn; yifan.she@mail.hfut.edu.cn; hudh@hfut.edu.cn)}, Yingying Sun, Yifan She, Donghui Hu\thanks{Corresponding author: Donghui Hu}}
}

\maketitle

\begin{abstract}
Federated learning (FL) enables multiple clients to collaboratively train a shared global model while preserving the privacy of their local data. Within this paradigm, the intellectual property rights (IPR) of client models are critical assets that must be protected. In practice, the central server responsible for maintaining the global model may maliciously manipulate the global model to erase client contributions or falsely claim sole ownership, thereby infringing on clients’ IPR. Watermarking has emerged as a promising technique for asserting model ownership and protecting intellectual property. However, existing FL watermarking approaches remain limited, suffering from potential watermark collisions among clients, insufficient watermark security, and non-intuitive verification mechanisms.
In this paper, we propose \texttt{FLClear}, a novel framework that simultaneously achieves collision-free watermark aggregation, enhanced watermark security, and visually interpretable ownership verification. Specifically, \texttt{FLClear} introduces a transposed model jointly optimized with contrastive learning to integrate the watermarking and main task objectives. During verification, the watermark is reconstructed from the transposed model and evaluated through both visual inspection and structural similarity metrics, enabling intuitive and quantitative ownership verification. Comprehensive experiments conducted over various datasets, aggregation schemes, and attack scenarios demonstrate the effectiveness of \texttt{FLClear} and confirm that it consistently outperforms state-of-the-art FL watermarking methods.
\end{abstract}

\begin{IEEEkeywords}
Watermarking, federated learning, visualizable verification, security  
\end{IEEEkeywords}

\section{Introduction}
\label{sec:introduction}
Federated learning (FL)~\cite{mcmahan2017communication} is a distributed machine learning paradigm that facilitates collaborative training of a global deep neural network (DNN) while preserving the privacy of locally sensitive data. In each training round, clients download the global model from a central server and update it locally using their private datasets to produce local models (or gradients). After local training, each client transmits its model updates to the server for aggregation. The server aggregates the received updates using a predefined algorithm (e.g., FedAvg~\cite{mcmahan2017communication}) and redistributes the updated global model to all clients for the next round. This iterative process continues until the global model converges or a predefined number of communication rounds is reached.
FL training demands substantial data and computational resources distributed across client devices, resulting in significant commercial and practical value~\cite{10.1145/3298981,LI2020106854}. For instance, in healthcare, FL enables collaborative training of predictive models for early disease detection by integrating decentralized patient data from multiple hospitals~\cite{10288131}. Other  notable applications include autonomous driving systems~\cite{taik2022clustered} and credit scoring models~\cite{kong2024asia}.

Because the global model is collaboratively trained by multiple clients, the intellectual property rights (IPR) associated with individual client contributions represent critical assets~\cite{yang2024fedgmark}. However, the global model remains susceptible to unauthorized manipulation and misappropriation, exposing clients to significant IPR risks~\cite{9847383}. For instance, a malicious server could falsely claim exclusive ownership of the aggregated model for commercial gain, disregarding the substantial contributions of participating clients. To mitigate such risks, embedding verifiable ownership information within the global model is essential to ensure fairness and accountability in FL systems.

\begin{figure}[t]
    \centering
    \includegraphics[width=0.95\columnwidth]{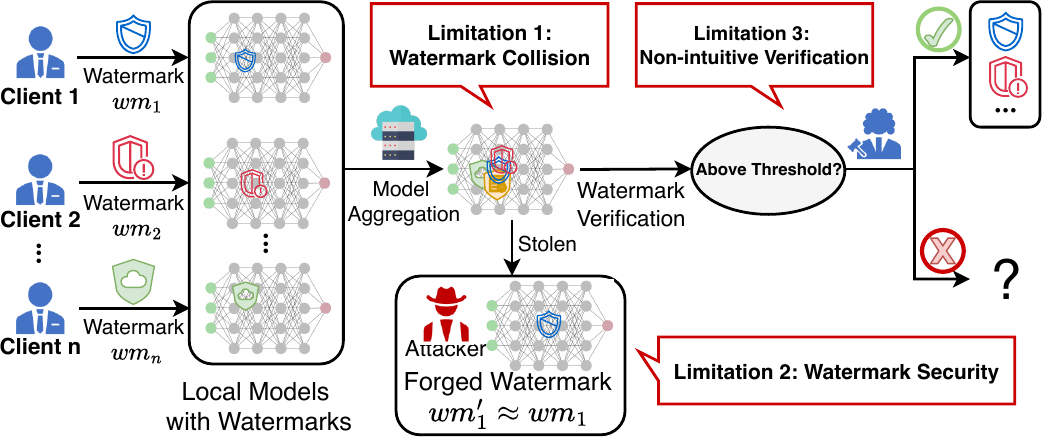}
    \caption{Limitations of existing FL watermarking}
    \label{fig:limit}
\end{figure}

\noindent \textbf{Research Gap.}
Watermarking~\cite{10.1145/3078971.3078974,10.5555/3277203.3277324,2021A} has emerged as a promising approach for safeguarding DNN model IPR, as it allows the embedding of verifiable ownership information without significantly degrading model performance. 
FL watermarking approaches typically embed watermarks on the server side to assert ownership of the global model~\cite{9603498,cao2024fedsmw,10888286,CHEN2023103504}. This design  mirrors conventional DNN watermarking, since the server maintains full control of the training pipeline. In contrast, because no individual client maintains full control over model training in FL, client-side watermarking becomes substantially more challenging, rendering traditional DNN watermarking schemes unsuitable for federated environments \cite{9603498,CHEN2023103504}.
Existing multi-client watermarking approaches can be broadly categorized into three types: (i) \emph{weight-based} methods~\cite{10630980,liang2023fedcipfederatedclientintellectual,xu2024robwe}, which embed ownership information directly into model parameters; (ii) \emph{trigger-based} methods~\cite{10.1016/j.eswa.2024.123776,li2025infightingdarkmultilabelbackdoor,luo2025harmlessbackdoorbasedclientsidewatermarking,gai2025mfl}, which establish ownership through distinctive input–output mappings (e.g., backdoor triggers); and \emph{hybrid} methods~\cite{9847383,10840113,10504977}, which combine both weight-based and trigger-based strategies. Nonetheless, all these approaches exhibit inherent limitations, as illustrated in \Cref{fig:limit}, which motivates the need for a more interpretable and secure watermarking mechanism.

\begin{itemize}[left=0pt]
	\item Limitation 1: \emph{Watermark Collision.} Collisions in weight-based watermarking methods arise from competitive embedding within constrained parameter spaces~\cite{10630980,liang2023fedcipfederatedclientintellectual,9847383}. A similar issue occurs in trigger-based watermarking when multiple clients adopt comparable trigger patterns but assign distinct output labels \cite{10.1016/j.eswa.2024.123776,luo2025harmlessbackdoorbasedclientsidewatermarking,li2025infightingdarkmultilabelbackdoor}. 
	Although recent studies attempt to alleviate this issue, they often compromise practicality by restricting the number of participating clients~\cite{xu2024robwe,liang2023fedcipfederatedclientintellectual,9847383} or assuming an honest server~\cite{luo2025harmlessbackdoorbasedclientsidewatermarking,gai2025mfl}. 
    \item Limitation 2: \emph{Watermark Security.} Modification attacks, which remove or distort legitimate watermarks, have been widely investigated. These attacks are commonly used to evaluate watermark \emph{robustness}, typically through pruning~\cite{han2015learning} or fine-tuning~\cite{lukas2022sok}. In contrast, forgery attacks~\cite{wang2019neural,10630980} fabricating counterfeit watermarks to undermine watermark \emph{security} have received limited attention. Most multi-client watermarking schemes remain susceptible to such forgery attempts~\cite{wang2019neural}. The method proposed in~\cite{10630980} provides resistance to forgery, but requires collaboration among clients. 
    \item Limitation 3: \emph{Non-intuitive Verification.} Existing watermark verification methods rely on statistical thresholds (e.g., bit error rate or classification accuracy), which are inherently non-intuitive and lack interpretability for human evaluation. Weight-based methods~\cite{10630980, xu2024robwe} depend on fixed thresholds for verification, while trigger-based methods~\cite{li2025infightingdarkmultilabelbackdoor,luo2025harmlessbackdoorbasedclientsidewatermarking,gai2025mfl} assess ownership through the classification accuracy of trigger samples.
\end{itemize}

\noindent \textbf{Challenges.}
These limitations motivate critical research challenges for designing an effective FL watermarking framework, leading to the following question:

\emph{How can a client-level FL framework simultaneously achieve collision-free, secure, and visually interpretable watermarking?}

To address this question, three core design challenges must be resolved (C1-C3). \emph{C1:} Client watermarks should be preserved and distinguishable after model aggregation. The solution should ensure that each client’s watermark avoids overlap or structural similarity with others. \emph{C2:} The solution should enhance watermark resilience against modification and forgery attacks. Since existing anti-forgery techniques~\cite{yao2025hashed,qiu2024belt} are designed for centralized training, new mechanisms tailored to the decentralized nature of FL are required. \emph{C3:} The verification process should produce visually verifiable results~\cite{8695386,10.5555/3698900.3699195} that enable reliable ownership authentication without dependence on arbitrary statistical thresholds. Importantly, the new approach should address all three challenges cohesively and integrate seamlessly into the FL training pipeline.

\noindent \textbf{Our Solution.} 
We propose \texttt{FLClear}, a FL watermarking framework that integrates a transposed model with contrastive learning. Each client constructs the transposed model by generating multiple input vectors derived from its unique watermark extraction vector, augmented through a vector augmentation mechanism, and assigns a human-perceptible watermark image as the designated output. The transposed model and the main-task model are jointly optimized through parameter sharing to enable simultaneous learning of both objectives.
During verification, the watermark is reconstructed from the transposed model and validated via both visual inspection and quantitative structural similarity metrics, providing transparent and reliable ownership authentication.

The experimental evaluation demonstrates that \texttt{FLClear} achieves high watermark fidelity and strong task performance across diverse datasets and aggregation schemes. Model accuracy remains comparable to baseline FL training, while the structural similarity index (SSIM) of reconstructed watermarks consistently exceeds 0.9. The framework also exhibits resilience against severe model-modification attacks including pruning~\cite{han2015learning}, fine-tuning~\cite{lukas2022sok}, quantization~\cite{tartaglione2021delving}, overwriting~\cite{2021A}, and reduces forgery attack success rates by nearly 100\%. When compared with state-of-the-art baselines such as FedTracker~\cite{10504977}, FedIPR~\cite{9847383}, and ClearStamp~\cite{10.5555/3698900.3699195}, \texttt{FLClear} achieves high main-task performance and strong watermark accuracy.

Our initial motivation was to design a visually verifiable watermarking scheme addressing the challenge of non-intuitive verification. Notably, we found that \texttt{FLClear}, through its integration of a transposed model and contrastive learning, also mitigates the challenges of collision and security, distinguishing it from prior watermarking methods: (i) a vector-based watermark key that inherently ensures low inter-client similarity, thereby preventing watermark collisions during model aggregation (addressing C1); (ii) a contrastive learning strategy that strengthens watermark security against forgery attacks (addressing C2); and (iii) a transposed model training process that enables intuitive, human-perceptible watermark verification without reliance on statistical thresholds (addressing C3).

The main contributions of this paper are summarized as follows:
\begin{itemize}[left=0pt]
	\item We propose the \emph{first} visually verifiable multi-client FL watermarking framework that supports both qualitative and quantitative ownership verification.
    \item We integrate a transposed model with contrastive learning to jointly optimize watermark and main-task objectives, achieving intuitive visual verification, conflict-free watermarking, and enhanced security.
    \item We implement and rigorously evaluate \texttt{FLClear} across diverse datasets, aggregation schemes, and attack scenarios. Results show that \texttt{FLClear} is compatible with multiple aggregation schemes, achieves 0\% forgery success rate, and consistently outperforms state-of-the-art watermarking baselines.
\end{itemize}

\noindent \textbf{Availability.} The source code for implementing \texttt{FLClear} is available at \url{https://github.com/Chen-Gu/FLClear}. We hope that this work can help advance FL watermark research and inspire further progress in the community.

\section{Background}
\label{sec:prel}

\noindent \textbf{Federated Learning.}
The FL framework contains a central server and $n$ clients $\{ \mathcal{C}_1, \mathcal{C}_2, \ldots,\mathcal{C}_n\}$, where each client $\mathcal{C}_i$ holds a local dataset $D_i$ of size $|D_i|$. Each client $\mathcal{C}_i$ computes its model parameters $\omega_i$ by minimizing the following objective function:
\begin{equation}
	\mathcal{F}_i(\omega)=\frac{1}{|D_i|} \sum_{\xi \in D_i} \mathcal{L}_i(\omega;\xi),
\end{equation}
where $\mathcal{L}_i(\cdot)$ denotes the empirical loss function for data sample $\xi \in D_i$. The global objective of FL is to obtain optimal model parameters by minimizing the overall loss function:
\begin{equation}
	\min _\omega \mathcal{F}(\omega) = \sum_{i=1}^n p_i \mathcal{F}_i(\omega),
\end{equation}
where $p_i = \frac{\left|D_i\right|}{|D|}$ represents the relative weight of client $\mathcal{C}_i$, and $\sum_{i=1}^n p_i=1$.
A typical FL training process consists of three iterative steps:
\begin{itemize}[left=0pt]
	\item Step I: Local Training: Each client performs local training on the received global model, computes local parameter updates, and transmits them to the server.
	\item Step II: Model Aggregation: The server aggregates all client updates to produce a new global model.
	\item Step III: Model Broadcasting: The updated global model is broadcast to all clients to initiate the next training round.
\end{itemize}

These steps repeat iteratively until convergence, i.e., when the global optimization objective approaches the optimal solution.

\noindent \textbf{Transposed Model.}
\label{subsec:transposed_model}
A transposed model~\cite{10.5555/3698900.3699195} approximates a reverse mapping of the original model by reversing the correspondence between its input and output spaces. It comprises a sequence of transposed layers, each functionally paired with its counterpart in the original model to invert data flow. The construction rules depend on the underlying layer type:
\begin{itemize}[left=0pt]
    \item Linear Layer: A standard linear layer computes $y = xW^T + b$, where $W$ denotes the weight matrix, $b$ is the bias, and $T$ represents the transpose operation. Its transposed counterpart reverses this mapping as $y = (x - b)W$.
    \item Convolutional Layer: A transposed convolutional layer reverses the spatial transformation of a standard convolutional layer~\cite{5539957}. Because pooling layers also affect spatial dimensions, their inverses are achieved by adjusting stride and padding parameters within the transposed convolution, thereby unifying the inversion process for convolutional and pooling operations.
    \item Batch Normalization: Since batch normalization (BN) preserves input dimensionality, we reuse the original BN operation in the transposed model, following the same computation~\cite{ioffe2015batch}: $ y = \frac{x - E(x)}{\sqrt{\text{Var}(x) + \epsilon}} \times \gamma + \beta $, where $\gamma$ and $\beta$ are learnable parameters, and $\epsilon$ ensures numerical stability.
    \item Other Layers: Activation and dropout layers neither alter dimensionality nor contain learnable parameters; hence, they remain unchanged during transposition.
\end{itemize}

\noindent \textbf{Contrastive Learning.}
\label{sec:cl}
Contrastive learning~\cite{khosla2020supervised} is a self-supervised representation learning paradigm that trains models to structure feature embeddings such that semantically similar (positive) pairs are drawn closer, and dissimilar (negative) pairs are pushed apart within the latent space. Given an anchor sample $x$, a positive sample $x^+$, and a negative sample $x^-$, an encoder $f(\cdot)$ maps each input to a latent embedding $z = f(x)$. 
The margin-based contrastive loss is defined as follows:
\begin{equation}
	\mathcal{L} = y \cdot D(x, x^+)^2 + (1 - y) \cdot \max(0, m - D(x, x^-))^2,
\end{equation}
where $D(\cdot)$ denotes a distance metric, $y = 1$ for positive pairs, $y = 0$ for negative pairs, and $m > 0$ specifies the enforced margin between embeddings of negative pairs.

\section{Threat Model}
\label{sec:tm}

\noindent \textbf{Adversary's goal.}
The adversary aims to manipulate client-embedded watermarks in the aggregated model to compromise ownership evidence while maintaining model utility. Specifically, the adversary may: (i) perform watermark modification to weaken or erase a legitimate client’s watermark, thus impeding reliable ownership verification; or (ii) execute watermark forgery to fabricate counterfeit ownership evidence falsely attributed to an adversary-controlled entity. All attacks are constrained by utility requirements, since adversaries typically seek to maintain acceptable model accuracy.

\noindent \textbf{Adversary's knowledge.}
We consider a \emph{white-box} adversary with access to the global model, including both read and write capabilities over its parameters. However, the adversary has no direct access to any client’s local model and therefore cannot reconstruct client-specific batch normalization (BN) statistics~\cite{li2021fedbn} or local data distributions. Moreover, we assume that the adversary may have partial knowledge of a client’s watermark. The adversary is further assumed to have complete knowledge of the proposed watermarking framework, including its training procedure and verification protocol. For example, the adversary understands the transposed-model training mechanism, including both input and output configurations. Leveraging this information, the adversary can launch attacks aimed at compromising the legitimate watermark. These assumptions define a strong adversarial setting that maximizes the attack feasibility against the proposed watermarking framework.

\noindent \textbf{Adversary's capacity.}
A white-box adversary enables multiple categories of attacks targeting watermark robustness and security. Specifically, the adversary may either manipulate the global model to remove existing watermarks or generate new forged ones. Depending on whether the global model is altered, attacks are categorized into two primary types:

\begin{itemize}[left=0pt]
	\item Manipulation Attacks: These attacks compromise watermark \emph{robustness} by modifying the global model (e.g., through pruning, fine-tuning, or quantization), which perturbs model parameters and reduces similarity to the original model \cite{10.1145/3078971.3078974,han2015learning}. These manipulations are generally applied after model aggregation. We also consider the overwriting attack~\cite{2021A}, in which the adversary replicates the \texttt{FLClear} training process to inject a new watermark into the model.
	\item Forgery Attacks: Instead of removing existing watermarks, forgery attacks compromise watermark \emph{security} by generating counterfeit watermark evidence that invalidates ownership verification. Here, the adversary does not alter the global model but exploits the transposed model $\mathcal{T}_\theta$ to optimize an input vector $\mathbf{v}_{atk}$, such that the generated watermark image $\mathbf{wm}_{atk}=\mathcal{T}_\theta(\mathbf{v}_{atk})$ approximates a target watermark $\mathbf{wm}$ \cite{wang2019neural,pang2025modelshield}. The forgery objective is formalized as:
\begin{equation}
	\min_{\mathbf{v}_{atk}} \left \| \mathcal{T}_\theta (\mathbf{v}_{atk})-\mathbf{wm} \right \|_2.
\end{equation}

Depending on whether the true watermark $\mathbf{wm}$ is known, forgery attacks are classified as either \emph{untargeted} or \emph{targeted}. In an untargeted attack, the adversary selects a random image as a pseudo-target $\mathbf{vm}$ and optimizes $\mathbf{v}_{atk}$ to establish a mapping between them~\cite{fan2019rethinking}. In contrast, a targeted forgery has knowledge of the true watermark $\mathbf{wm}$ and seeks to construct an extraction vector $\mathbf{v}_{atk}$ that reproduces $\mathbf{wm}$ (i.e., $\mathbf{v}_{atk}$ should resemble the genuine extraction vector $\mathbf{v}$).
\end{itemize}

\section{Design of \texttt{FLClear}}
\label{sec:theory}

\begin{figure}[t]
    \centering
    \includegraphics[width=0.95\columnwidth]{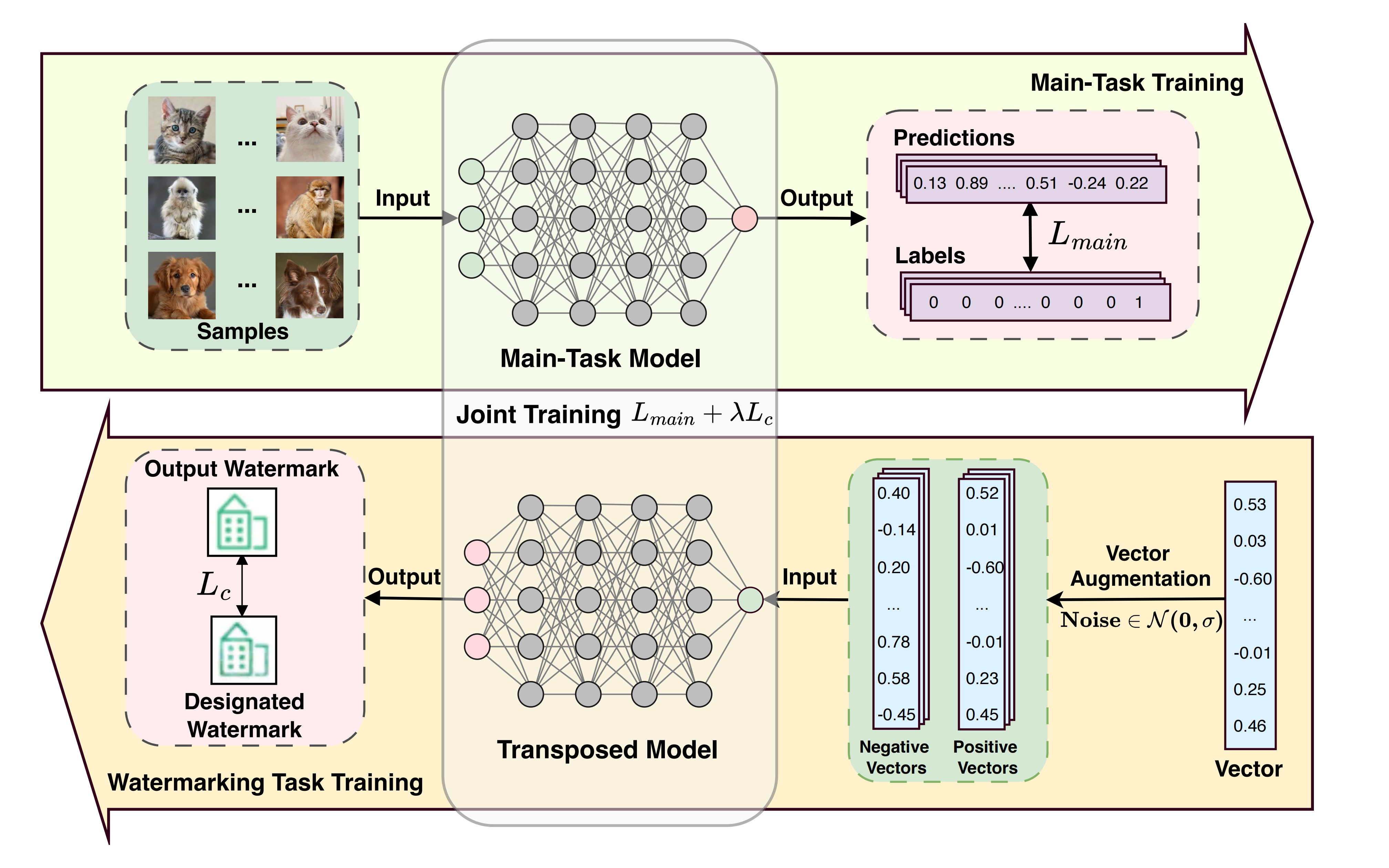}
    \caption{\texttt{FLClear} framework}
    \label{fig:ov}
\end{figure}


\subsection{Overview}
The overall design of \texttt{FLClear} is illustrated in \Cref{fig:ov}. \texttt{FLClear} employs a transposed model for watermark embedding that shares parameters with the main-task model to achieve joint feature representation. Specifically, the architecture of the transposed model is derived from the main-task model based on transposition rules. The framework uses a watermark extraction vector as a watermark key and applies a vector augmentation mechanism to generate diverse transposed model inputs. To enable the transposed model to generate the designated watermark image, it is jointly optimized with the main-task model under a contrastive loss objective.
During verification, each client uses its watermark extraction vector to reconstruct a human-recognizable watermark image, supporting both intuitive visual and quantitative verification. \texttt{FLClear} effectively addresses challenges of watermark collisions, security, and interpretability, providing an efficient and practical framework for intellectual property protection in FL.

\subsection{Watermarking Task Training}
\noindent \textbf{Step 1: Input Vectors Generation.}
Since the main-task model maps input images to classification vectors, the client constructs its reverse counterpart, the transposed model by applying predefined transposition rules~\cite{10.5555/3698900.3699195}. The transposed model performs the reverse mapping, generating a watermark image for verification from an input vector.
In our framework, a watermark extraction vector $\mathbf{v}$ serves as the input to the transposed model, whose dimensionality corresponds to the number of classes in the main task. The use of such a vector naturally ensures low inter-client similarity, thereby preventing potential watermark collisions during model aggregation. Moreover, the vector $\mathbf{v}$ also functions as a secret key for ownership verification. 
However, training the transposed model $\mathcal{T}_{\theta}$ with only a single vector–image pair often leads to overfitting, as the model tends to memorize the fixed input–output mapping rather than learning a generalized representation. This overfitting impairs the generalization capability of the generated watermark images.

\begin{figure}[t]
    \centering
    \begin{subfigure}{0.48\columnwidth}
        \centering
        \includegraphics[width=\textwidth]{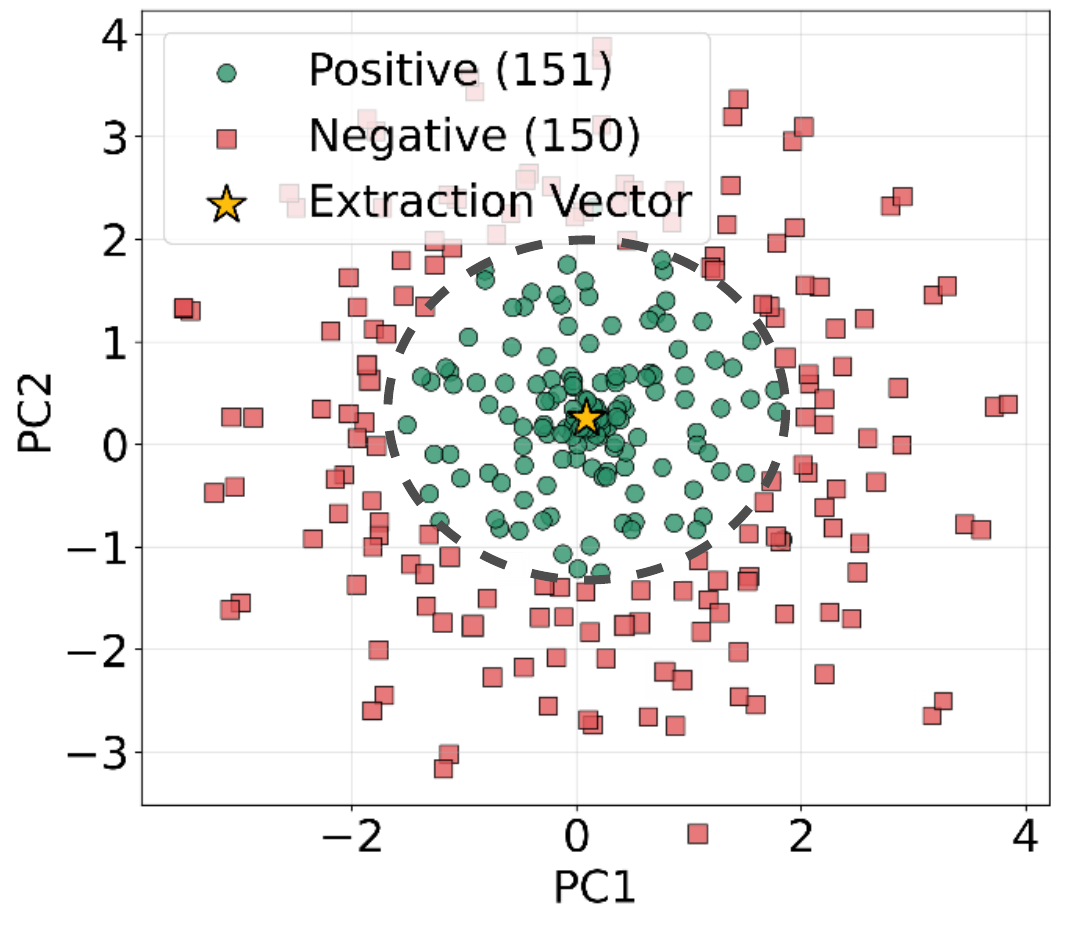}
        \caption{\centering{Visualization of Augmented Vectors}}
        \label{fig:va_a}
    \end{subfigure}
    \begin{subfigure}{0.48\columnwidth}
        \centering
        \includegraphics[width=\textwidth]{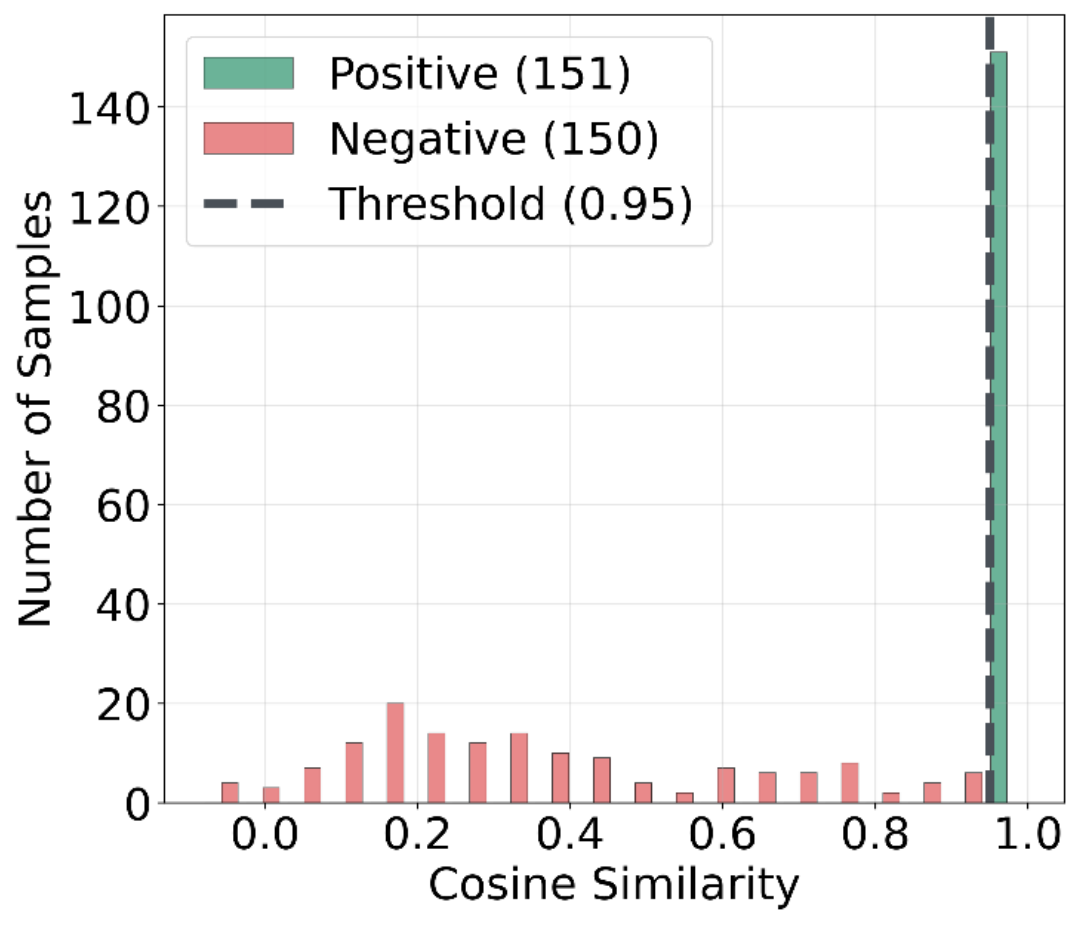}
        \caption{\centering{Similarity Distribution of Vectors}}
        \label{fig:va_b}
    \end{subfigure}
    \caption{Vector augmentation. \Cref{fig:va_a} visualizes PCA projection of the augmented vectors, showing a balanced number of positive and negative vectors around the extraction vector. \Cref{fig:va_b} illustrates the distribution of negative and positive vectors separated by the similarity threshold.}
    \label{fig:va}
\end{figure}

We introduce a vector augmentation mechanism, illustrated in \Cref{fig:va}, to improve the transposed model’s robustness and discriminativeness. The mechanism generates diverse vector samples $\mathbb{D}_{v} = \{\mathbf{v}_1, \mathbf{v}_2, \ldots, \mathbf{v}_n \}$ by perturbing $\mathbf{v}$ with Gaussian noise:
\begin{equation}
	\mathbf{v}_i = \mathbf{v} + \mathcal{N} (0,\sigma^{2}),
\end{equation}
where $\sigma$ controls the deviation between $\mathbf{v}_i$ and $\mathbf{v}$. Samples with cosine similarity below a predefined threshold are labeled as negative vectors, while those above are positive. Balanced sampling ensures robustness and discrimination in the learned space. 

\noindent \textbf{Step 2: Output Designation.}
Each client specifies a predefined, human-recognizable watermark image, denoted as $\mathbf{wm}$, which serves as the designated output of the transposed model. The dimensions of $\mathbf{wm}$ are identical to those of the input images used in the main task, ensuring compatibility with the model architecture. Unlike conventional watermarking methods that embed imperceptible or statistically verifiable signals, \texttt{FLClear} explicitly associates a visually interpretable image with each client, enabling direct and intuitive ownership verification.

\noindent \textbf{Step 3: Transposed Model Training.}
We employ contrastive learning~\cite{khosla2020supervised} to train the transposed model. The positive loss encourages $\mathcal{T}_{\theta}(\mathbf{v}^+)$ to resemble the true watermark $\mathbf{wm}$, while the negative loss forces $\mathcal{T}_{\theta}(\mathbf{v}^-)$ to diverge. The contrastive loss $\mathcal{L}_{c}$ is defined as:
\begin{equation}
\label{eq:cl}
\begin{aligned}
    \mathcal{L}_{c} = &\ y \left(1 - \text{SSIM}(\mathcal{T}_{\theta}(\mathbf{v}^+), \mathbf{wm})\right) \\ & +(1-y) \max \left(0, (m-\text{SSIM}(\mathcal{T}_{\theta}(\mathbf{v}^-), \mathbf{wm}))\right),
\end{aligned}
\end{equation}
where $m$ denotes the similarity margin. The SSIM is adopted as the similarity measure to align reconstructed and reference images in both pixel and perceptual feature spaces. This formulation drives the model to distinguish genuine watermark vectors from adversarial or random ones. During training, the transposed model is optimized to reconstruct this designated image from the client’s vectors, thereby embedding both semantic distinctiveness and visual traceability into the model.

\subsection{Joint Optimization}
The training of the watermarking task and the main task is performed through a joint optimization strategy, formulated as a multi-task learning problem~\cite{zhang2021survey} that leverages shared model parameters. Specifically, the watermarking task is optimized via a transposed model that shares all learnable parameters with the main-task model. This parameter sharing enables the watermark to be inherently embedded within the main model’s parameters, ensuring tight coupling between task learning and watermark representation.
Only learnable parameters are shared between the two models. Non-learnable parameters, such as the running mean and variance in normalization layers, are maintained independently in each client model to accommodate their distinct input data distributions. The overall joint objective is:
\begin{equation}
\label{eq:ojo}
	\mathcal{L} = \mathcal{L}_{main} + \lambda \cdot \mathcal{L}_{c},
\end{equation}
where $\mathcal{L}_{main}$ denotes the loss of the main task. The hyperparameter $\lambda$ balances the contributions of both tasks. 
 To ensure effective watermark embedding and reliable extraction, each client locally store two types of key data: its unique watermark extraction vector and the BN statistics associated with the watermarking task. Separating and preserving these BN statistics is essential because the data distributions of the two tasks differ fundamentally. While the main task involves image classification, the watermarking task performs a vector-to-image reconstruction. Storing task-specific BN statistics ensures stable training and robust watermark recovery.

\subsection{\texttt{FLClear} Implementation}

\begin{algorithm}[t]
	\caption{\texttt{FLClear} Implementation}
	\label{alg:FLClear}
	\begin{algorithmic}[1]
	\Statex \textbf{Input:}  A set of clients $\{C_1, \ldots, C_n\}$; predefined number of training rounds $T$.
	\Statex \textbf{Output:} Global model $\mathcal{G}^{T}$.
	\Foreach{each round $t = 1, \ldots, T$}
		\State \underline{\emph{On each client $C_i$:}}
		\State Construct the transposed model $\mathcal{T}_i$ according to the transposed rules;
		\State Initialize the watermarking task training using augmented vectors and the designated watermark image;
		\State Jointly optimize the main-task model $\mathcal{M}_i(\theta)$ and the transposed model $\mathcal{T}_i(\theta)$; \quad \text{// Using \Cref{eq:ojo}}
		\State Upload the updated main-task model $\mathcal{M}_i(\theta)$ to the server;
	
		\State \underline{\emph{On the server:}}
		\State Aggregate received models $\{ \mathcal{M}_1(\theta), \ldots, \mathcal{M}_n(\theta) \}$  using a specific aggregation scheme;
		\State Update the global model $\mathcal{G}^{t}$ and distribute it to all clients;
	\ENDFOR
	\State \textbf{Return} $\mathcal{G}^{T}$.
	\end{algorithmic}
\end{algorithm}

We integrate watermark training with the main-task model on each client and outline the implementation of \texttt{FLClear} in \Cref{alg:FLClear}. At each communication round, every client $C_i$ constructs a transposed model $\mathcal{T}_i$ derived from its main-task model according to the transposition rules. 
Each client utilizes augmented vectors and the designated watermark image for training the watermarking task.
Both the main-task model $\mathcal{M}_i(\theta)$ and the transposed model $\mathcal{T}_i(\theta)$ are jointly optimized, sharing the same parameter set $\theta$ to ensure alignment between the main-task learning and the embedded watermark representation. Upon completion of local training, each client uploads the updated parameters of $\mathcal{M}_i(\theta)$ to the central server. The server aggregates all received updates $\{\mathcal{M}_1(\theta), \mathcal{M}_2(\theta), \ldots, \mathcal{M}_n(\theta)\}$ according to a specified aggregation scheme (e.g., FedAvg) to obtain the updated global model, which is then redistributed to the clients for the next round. This iterative process continues until convergence or a predefined number of rounds $T$ is reached, yielding a final global model $\mathcal{G}^{T}$ that preserves both main-task performance and watermark fidelity.

\subsection{Watermark Verification Method}
During verification, the verifier reconstructs a transposed model from the main model using transposition rules and BN statistics. The watermark extraction vector $\mathbf{v}$ is fed into the transposed model to generate the reconstructed watermark $\mathbf{wm}^{\prime}$.
Verification can be performed in two complementary ways: (i) \emph{visual inspection}, where the reconstructed image $\mathbf{wm}^{\prime}$ is visually compared against the designated watermark $\mathbf{wm}$; and (ii) \emph{quantitative evaluation}, which computes a similarity metric such as SSIM between $\mathbf{wm}$ and $\mathbf{wm}^{\prime}$. Verification is successful when $\texttt{SSIM}(\mathbf{wm}, \mathbf{wm}^{\prime}) \ge \tau$, where $\tau$ is a verification threshold.
The primary advantage of visual verification lies in its inherent robustness. An adversary must reproduce a human-recognizable image, not merely achieve a high numerical score.

\subsection{Discussion: Why does \texttt{FLClear} Work?}
\label{sec:theory-discussion}
In this section, we provide explanations of why our proposed solution is effective by answering the following core questions.

\begin{figure}[t]
    \centering
    \begin{subfigure}{0.51\columnwidth}
        \centering
        \includegraphics[width=\textwidth]{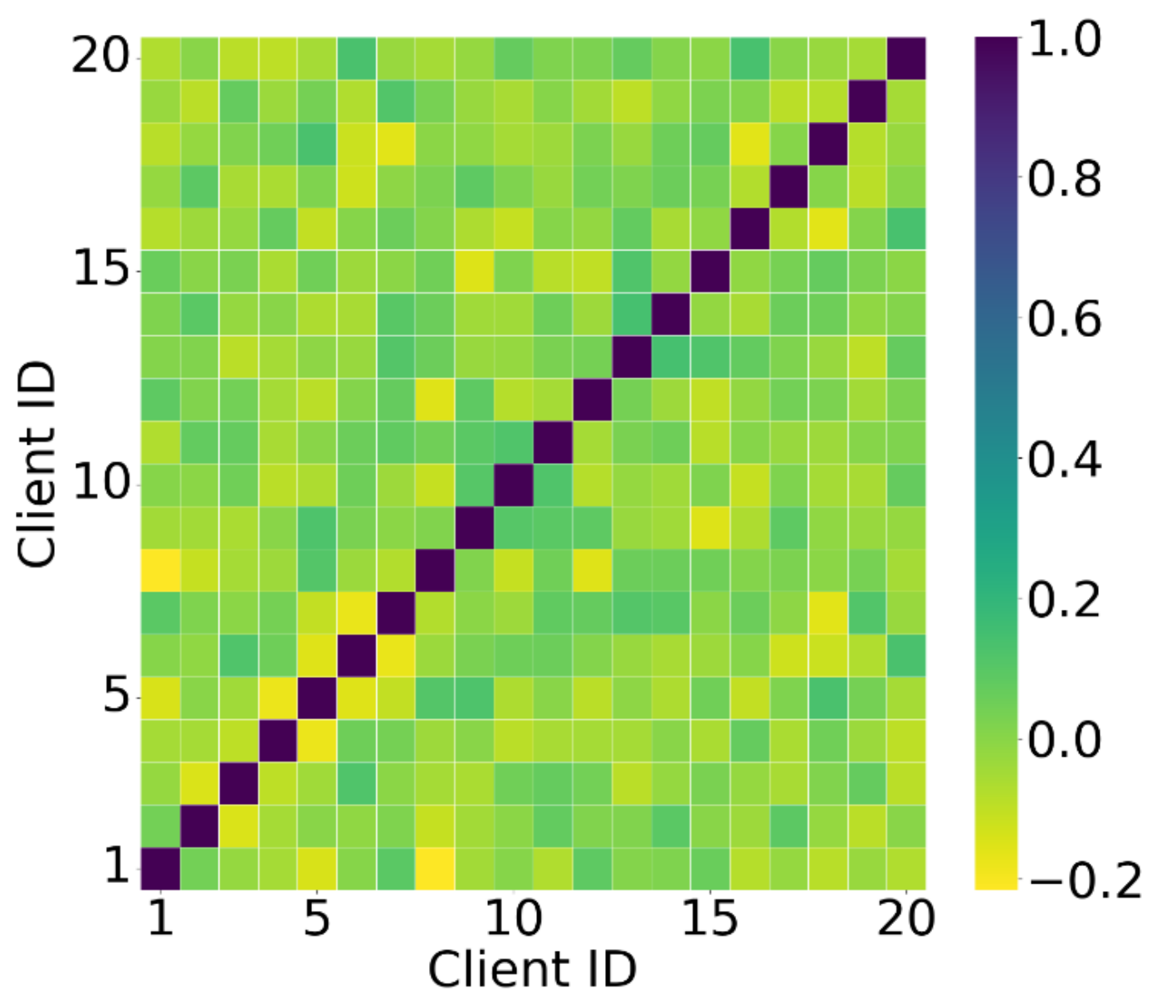}
        \caption{\centering{Cosine-similarity matrix of clients’ watermark vectors}}
        \label{fig:agg_a}
    \end{subfigure}
    \begin{subfigure}{0.47\columnwidth}
        \centering
        \includegraphics[width=\textwidth]{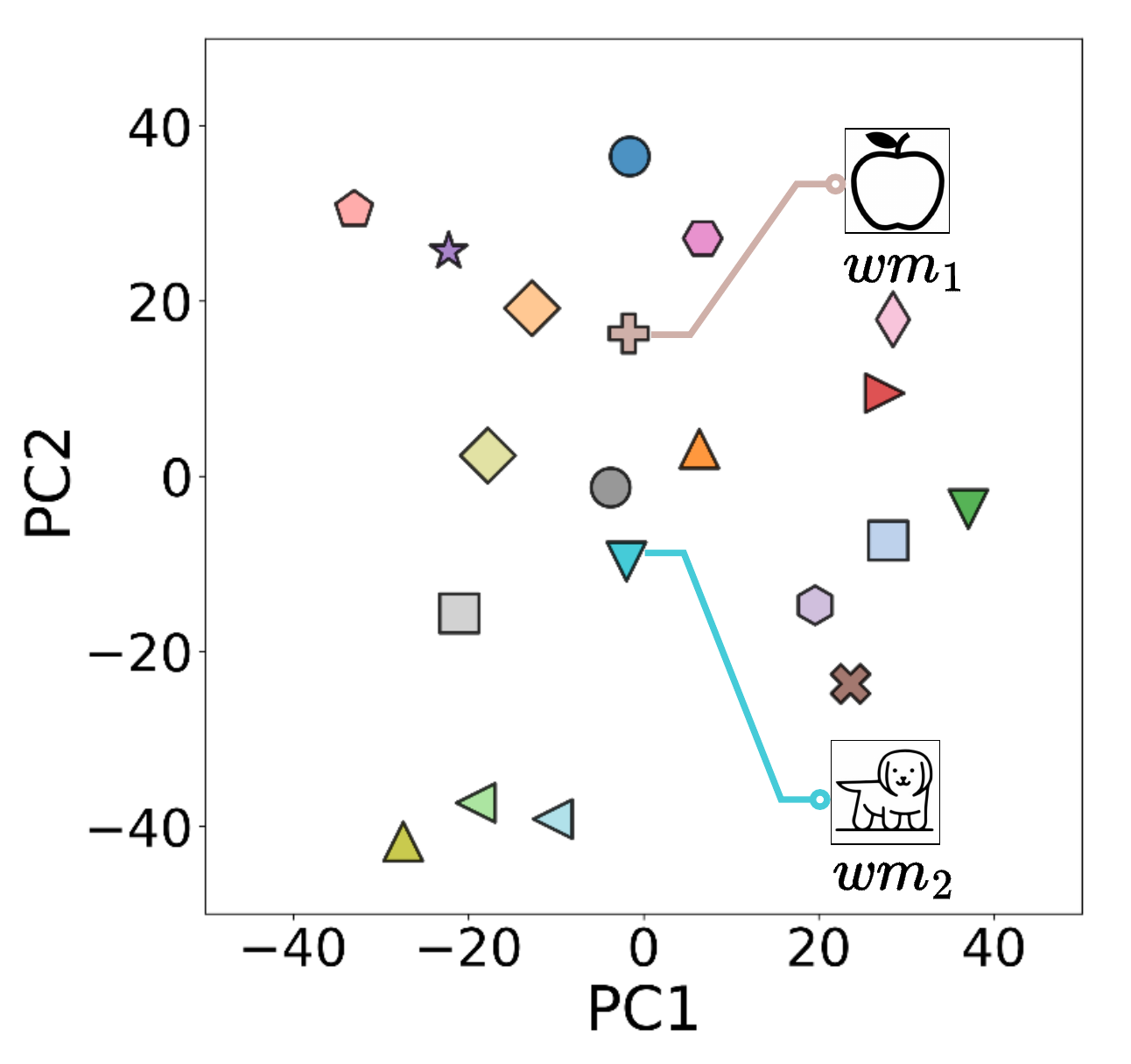}
        \caption{\centering{Visualization of watermark feature vectors}}
        \label{fig:agg_b}
    \end{subfigure}
    \caption{Analysis of watermark non-collision across clients. Each parameter in the vector is randomly sampled from range $[-1, 1]$.}
    \label{fig:agg}
\end{figure}

\noindent \textbf{Q1: Why do watermarks avoid collision after model aggregation?}
The robustness of \texttt{FLClear} against watermark collisions arises from the design of its watermark extraction vectors, which function as unique keys for each client. These vectors exhibit low inter-client similarity, ensuring that watermarks from different clients are independently mapped within the feature space. By encoding watermarks in the transposed model’s feature representation rather than directly in the model weights, \texttt{FLClear} minimizes the risk of overlapping watermark signals during aggregation. As shown in \Cref{fig:agg_a}, the pairwise cosine similarities among the 20 initialized watermark extraction vectors remain low, confirming that each vector occupies a distinct region of the feature space. Furthermore, \Cref{fig:agg_b} illustrates that the corresponding 20 watermarks are clearly separated, demonstrating that they are embedded in non-overlapping feature subspaces. This design effectively prevents watermark collisions during model aggregation and ensures reliable ownership verification.
 
\noindent \textbf{Q2: Why are watermarks securely preserved against forgery attacks?}
This security stems from the intrinsic difficulty of forging a genuine watermark extraction vector when contrastive loss is applied. As described in \Cref{sec:tm}, attackers cannot directly access the true watermark extraction vector $\mathbf{v}$. Consequently, both targeted and untargeted forgery attacks must attempt to generate a forged vector $\mathbf{v}_{atk}$ that closely resembles $\mathbf{v}$ so that the transposed model produces a forged watermark $\mathbf{vm}_{atk}$ visually similar to the authentic watermark $\mathbf{vm}$. However, the contrastive loss effectively constrains this process by driving most forged vectors into the negative sample space, where they are encouraged to produce dissimilar outputs. As a result, the transposed model generates $\mathbf{vm}_{atk}$ that significantly differ from the genuine $\mathbf{vm}$, thereby ensuring strong resistance against forgery attempts.

\begin{figure}[t]
    \centering
    \newcommand{\lcimgheight}{3.5cm}
    \begin{subfigure}[b]{0.38\columnwidth}
        \centering
        \includegraphics[height=\lcimgheight]{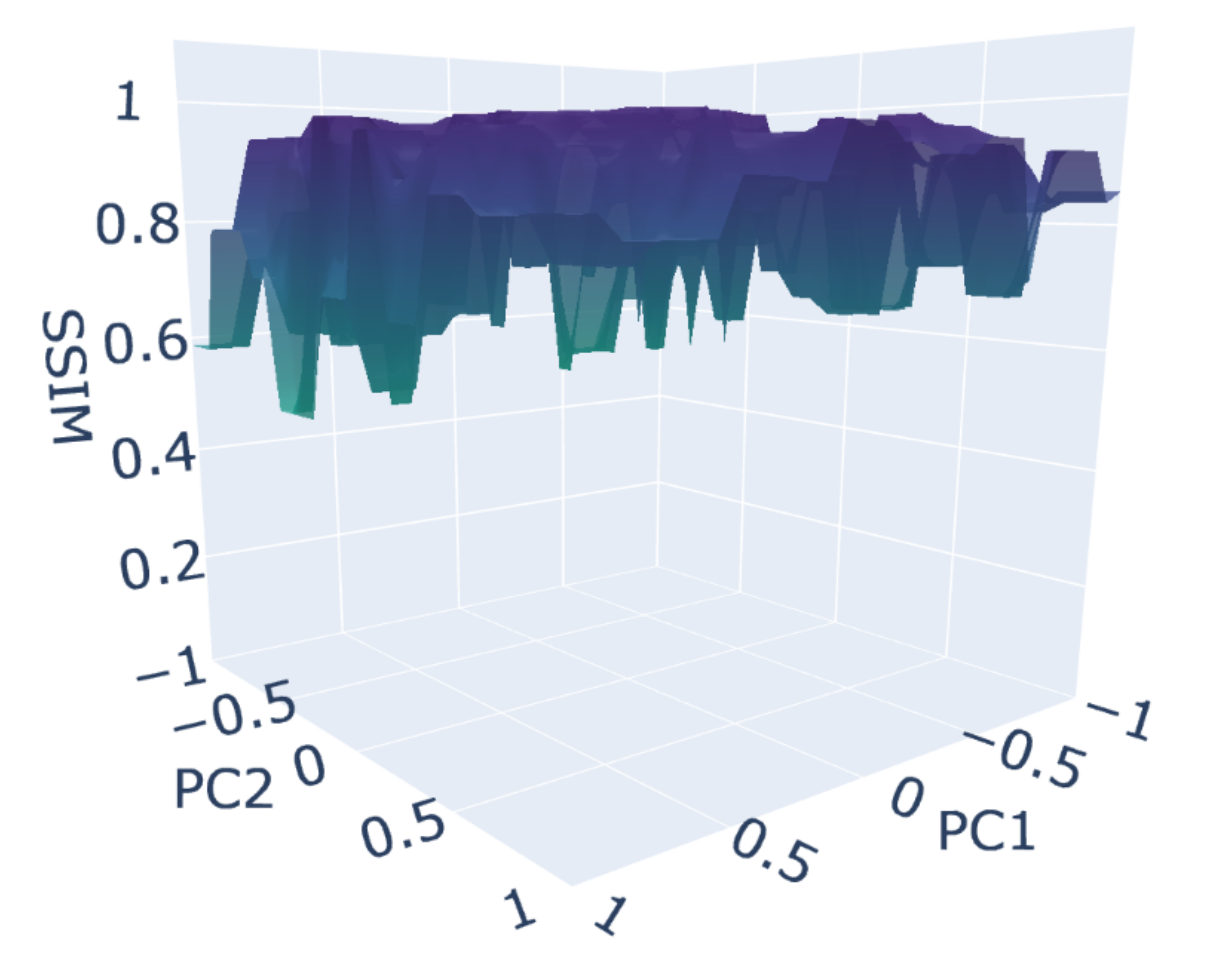}
        \caption{w/o Contrastive Loss}
        \label{fig:lc_a}
    \end{subfigure}
    \hfill
    \begin{subfigure}[b]{0.5\columnwidth}
        \centering
        \includegraphics[height=\lcimgheight]{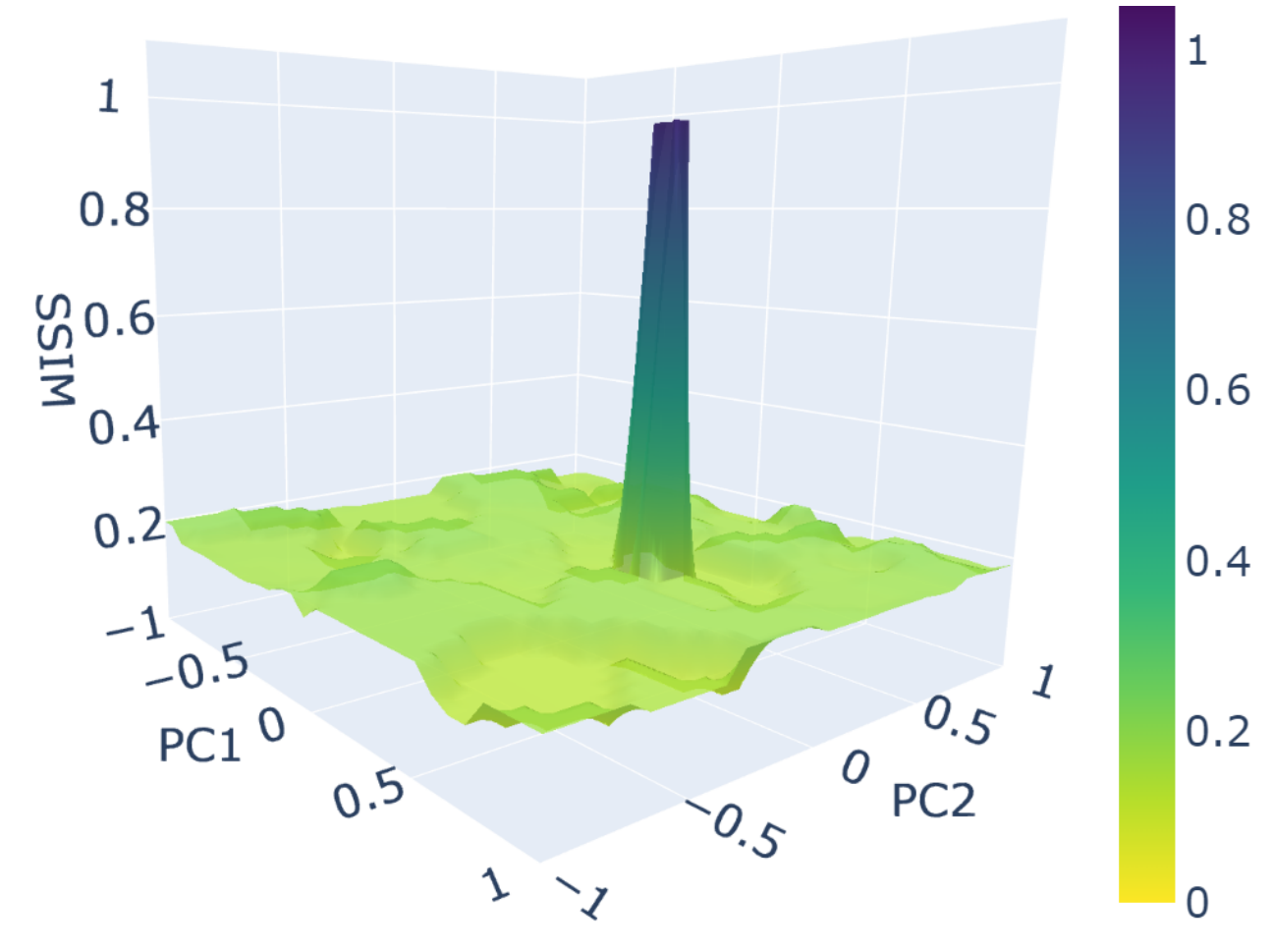}
        \caption{w/ Contrastive Loss}
        \label{fig:lc_b}
    \end{subfigure}
\caption{Effect of contrastive learning on the watermark similarity. Darker regions indicate higher similarity between the forged watermark and the genuine one.}
\label{fig:lc}
\end{figure}

We visualize the similarity distribution between $\mathbf{vm}_{atk}$ and $\mathbf{vm}$ under conditions with and without contrastive loss in \Cref{fig:lc}. The SSIM is used to quantify the discrepancy between forged and genuine watermarks, where an SSIM value of 1 indicates identical images, and lower values correspond to greater structural deviations. As shown in \Cref{fig:lc_a}, without the contrastive loss constraint, SSIM values remain high and the similarity landscape appears smooth, suggesting that forged watermarks $\mathbf{vm}_{atk}$ can easily replicate the genuine watermark $\mathbf{vm}$. In contrast, incorporating contrastive loss reshapes the landscape into a highly non-convex structure (\Cref{fig:lc_b}), where most forged vectors yield low SSIM values. A small cluster of watermarks derived from positive vectors exhibits higher similarity because these vectors are close to the genuine one. However, it remains challenging for an adversary to optimize its vector to align with these positive samples without the correct normalization statistics and the extraction vector.

\noindent \textbf{Q3: Why do watermarks allow visible verification?}
The visible verifiability of watermarks arises from the structured embedding space established by the transposed model architecture. The transposed model decodes embeddings into the pixel domain while preserving both spatial and semantic coherence. This design creates a mapping between the watermark extraction vector and its reconstructed output, ensuring that the generated watermark remains visually interpretable. Consequently, \texttt{FLClear} produces watermarks that are inherently human-recognizable, enabling direct and intuitive ownership verification. Even under model aggregation or minor perturbations, the semantic integrity of the reconstructed image remains visually discernible to human observers.

\section{Evaluation}
\label{sec:results}

\captionsetup{
  font={large},
  labelfont=bf,
  textfont={normalfont},
  justification=centering,
  skip=6pt
}

\begin{table*}[t]
\centering
\caption{Impact of various $y$ and $m$ values in the  contrastive loss function}
\label{tab:ym}
\renewcommand{\arraystretch}{1.15}
\setlength{\tabcolsep}{4pt}

\resizebox{\textwidth}{!}{%
\begin{tabular}{cccccccccccccccccccccc}
\toprule
\multirow{2}{*}{\textbf{Metric}} &
\multirow{2}{*}{\diagbox{$\boldsymbol{y}$}{$\boldsymbol{m}$}} &
\multicolumn{5}{c}{\textbf{MNIST}} &
\multicolumn{5}{c}{\textbf{Fashion-MNIST}} &
\multicolumn{5}{c}{\textbf{CIFAR-10}} &
\multicolumn{5}{c}{\textbf{CIFAR-100}} \\
\cmidrule(lr){3-7} \cmidrule(lr){8-12} \cmidrule(lr){13-17} \cmidrule(lr){18-22}
& & 0.1 & 0.3 & 0.5 & 0.7 & 0.9 & 0.1 & 0.3 & 0.5 & 0.7 & 0.9 & 0.1 & 0.3 & 0.5 & 0.7 & 0.9 & 0.1 & 0.3 & 0.5 & 0.7 & 0.9 \\

\midrule

\multirow{5}{*}{\textbf{Acc (\%)} $\uparrow$} 
& 0.1 & 99.51 & 99.49 & 99.47 & 99.48 & 99.49 & 89.97 & 90.47 & 90.31 & 90.09 & 90.25 & 91.60 & 89.23 & 91.34 & 90.48 & 89.74 & 69.25 & 68.32 & 69.00 & 68.35 & 68.77 \\
& 0.3 & 99.35 & 99.31 & 99.31 & 99.31 & 99.31 & 90.11 & 90.03 & 90.01 & 90.27 & 90.27 & 89.89 & 89.14 & 89.67 & 89.65 & 89.65 & 68.24 & 69.23 & 68.59 & 68.91 & 68.08 \\
& 0.5 & 99.50 & 99.32 & 99.49 & 99.52 & 99.51 & 89.91 & 89.87 & 90.18 & 89.99 & 90.34 & 91.82 & 89.22 & 89.55 & 89.66 & 89.99 & 68.75 & 68.83 & 69.22 & 68.83 & 68.75 \\
& 0.7 & 99.48 & 99.31 & 99.48 & 99.50 & 99.50 & 90.07 & 89.87 & 90.17 & 90.34 & 90.46 & 91.43 & 89.30 & 90.06 & 91.13 & 89.29 & 68.84 & 68.10 & 69.38 & 68.75 & 68.36 \\
& 0.9 & 99.46 & 99.32 & 99.50 & 99.51 & 99.45 & 89.64 & 90.14 & 90.21 & 89.85 & 90.17 & 90.49 & 89.57 & 90.37 & 89.34 & 90.36 & 69.15 & 68.77 & 68.35 & 69.09 & 68.55 \\
\midrule

\multirow{5}{*}{\textbf{SSIM} $\uparrow$} 
& 0.1 & 0.98 & 0.99 & 0.98 & 0.98 & 0.98 & 0.98 & 0.99 & 0.97 & 0.97 & 0.96 & 0.97 & 0.98 & 0.97 & 0.98 & 0.97 & 0.91 & 0.93 & 0.92 & 0.91 & 0.92 \\
& 0.3 & 0.97 & 0.98 & 0.98 & 0.98 & 0.98 & 0.93 & 0.93 & 0.89 & 0.87 & 0.87 & 0.97 & 0.98 & 0.99 & 0.99 & 0.99 & 0.93 & 0.93 & 0.93 & 0.93 & 0.93 \\
& 0.5 & 0.99 & 0.99 & 0.99 & 0.99 & 0.99 & 0.96 & 0.95 & 0.86 & 0.77 & 0.75 & 0.98 & 0.98 & 0.98 & 0.98 & 0.98 & 0.92 & 0.93 & 0.92 & 0.93 & 0.92 \\
& 0.7 & 0.99 & 0.99 & 0.99 & 0.99 & 0.99 & 0.85 & 0.95 & 0.80 & 0.67 & 0.71 & 0.98 & 0.99 & 0.99 & 0.99 & 0.98 & 0.92 & 0.93 & 0.93 & 0.93 & 0.93 \\
& 0.9 & 1.00 & 0.99 & 0.99 & 0.99 & 1.00 & 0.91 & 0.91 & 0.81 & 0.75 & 0.75 & 0.98 & 0.99 & 0.98 & 0.98 & 0.98 & 0.93 & 0.93 & 0.93 & 0.92 & 0.93 \\
\bottomrule
\end{tabular}
}
\end{table*}

In response to the research questions introduced in \Cref{sec:introduction} and elaborated in \Cref{sec:theory-discussion}, we present comprehensive experimental results in this section to address the following key questions.

\begin{itemize}
\item \textbf{Q1.} \emph{Can \texttt{FLClear} maintain collision-free watermark integrity after federated aggregation?} (\S \ref{sec:aggregation})
\item \textbf{Q2.} \emph{Can the embedded watermarks effectively withstand various adversarial attacks?} (\S \ref{sec:security})
\item \textbf{Q3.} \emph{Are the watermarks visually interpretable and verifiable through both qualitative and quantitative evaluations?} (\S \ref{sec:aggregation}, \S \ref{sec:security})
\end{itemize}

Besides these evaluations, we also assess the performance in terms of parameter sensitivity (\S\ref{sec:parameters}), the impact of contrastive learning (\S\ref{sec:ablation}), baseline comparison (\S\ref{sec:comparison}), watermark capacity (\S\ref{sec:capacity}), and overhead (\S\ref{sec:overhead}).

\subsection{Experimental Setup}
\noindent \textbf{Parameter Setting.}
We adopted a non-independent and identically distributed (Non-IID) data partitioning scheme described in~\cite{hsu2019measuring} with concentration parameter $\alpha = 0.8$. We used stochastic gradient descent (SGD) with a learning rate of 0.001 as the optimizer. The key parameters affecting watermark performance were varied as follows: the weight $y \in \{0.1,0.3,0.5,0.7,0.9\}$, the margin $m \in \{0.1,0.3,0.5,0.7,0.9\}$, the coefficient $\lambda \in \{0.1,0.5,1,5,10\}$, and the number of input vectors $num \in \{250, 500, 750\}$. We conducted FL experiments involving between 10 and 50 clients.

\noindent \textbf{Models and Datasets.}
We evaluated four DNN architectures of AlexNet~\cite{krizhevsky2012imagenet}, MobileNetV2~\cite{sandler2018mobilenetv2}, VGG13~\cite{simonyan2014very}, and ResNet18~\cite{he2016deep}.
Experiments were performed on four benchmark datasets: MNIST~\cite{LeCun:1998:MnistDatabaseHandwritten}, Fashion-MNIST~\cite{xiao2017fashion}, CIFAR-10~\cite{krizhevsky2009learning}, and CIFAR-100~\cite{krizhevsky2009learning}.   
Each model was paired with a corresponding dataset to ensure a balanced evaluation: 
AlexNet on MNIST, MobileNetV2 on Fashion-MNIST, VGG13 on CIFAR-10, and ResNet18 on CIFAR-100.

\noindent \textbf{Attack Methods.}
We implemented five watermark-based attacks including pruning, fine-tuning, quantization, overwriting, and forgery attacks. These attacks encompass a broad spectrum of model modification and watermark forgery scenarios, enabling a comprehensive evaluation of both the robustness and the security of the embedded watermarks.

\noindent \textbf{Baselines.}
Watermark performance after model aggregation was evaluated using five mainstream federated aggregation algorithms: FedAvg~\cite{mcmahan2017communication}, FedProx~\cite{li2020federated}, FedPAQ~\cite{reisizadeh2020fedpaq}, FedADAM~\cite{DBLP:conf/iclr/ReddiCZGRKKM21}, and SCAFFOLD~\cite{karimireddy2020scaffold}. 
To validate the effectiveness of \texttt{FLClear}, we compared it with state-of-the-art watermarking approaches, including WAFFLE~\cite{9603498}, \cite{10888286}, FedTracker~\cite{10504977}, FedIPR~\cite{9847383}, and ClearStamp~\cite{10.5555/3698900.3699195}. These baselines encompass both parameter-based and trigger-based federated watermarking schemes. ClearStamp was included for comparison, as it also employs a transposed model for watermark embedding.

\noindent \textbf{Evaluation Metrics.}
We employ multiple evaluation metrics to comprehensively assess the effectiveness of both the main task and the watermarking task. \emph{Model accuracy (Acc)} quantifies classification performance on the main task, whereas the \emph{SSIM} assesses the quality of the extracted visual watermark. Because SSIM serves as the primary optimization objective for watermark training, we further complement it with \emph{mean squared error (MSE)}, \emph{peak signal-to-noise ratio (PSNR)}, and \emph{learned perceptual image patch similarity (LPIPS)} \cite{hu2024learn} to provide a comprehensive assessment of visual fidelity. We also employ the \emph{attack success rate (ASR)} to evaluate the effectiveness of forgery attacks. ASR is defined as the ratio of successful forgeries whose extracted watermark satisfies $SSIM(\textbf{vm}, \textbf{vm}^{\prime}) \geq \tau$, relative to the total number of attack attempts. The number of attack attempts in the experiments was fixed at 300.

\subsection{Parameter Sensitivity}
\label{sec:parameters}

\begin{table}[t]
\centering
\caption{Impact of various $\lambda$ values}
\label{tab:lambda}
\renewcommand{\arraystretch}{1.1}
\setlength{\tabcolsep}{3pt}
\resizebox{\columnwidth}{!}{
\begin{tabular}{ccccccccc}
\toprule

\multirow{2}{*}{$\lambda$} & \multicolumn{2}{c}{\textbf{MNIST}} 
                            & \multicolumn{2}{c}{\textbf{Fashion-MNIST}} 
                            & \multicolumn{2}{c}{\textbf{CIFAR-10}} 
                            & \multicolumn{2}{c}{\textbf{CIFAR-100}} \\
\cmidrule(lr){2-3}\cmidrule(lr){4-5}\cmidrule(lr){6-7}\cmidrule(lr){8-9}
                            & Acc (\%) & SSIM & Acc (\%) & SSIM & Acc (\%) & SSIM & Acc (\%) & SSIM \\
\midrule
0.1               & 99.37 & 0.38 & 89.73 & 0.94 & 91.59 & 0.88 & 68.54 & 0.83 \\
0.5               & 99.40 & 0.93 & 90.57 & 0.96 & 91.25 & 0.92 & 68.87 & 0.91 \\
1                 & 99.36 & 0.98 & 89.91 & 0.96 & 91.18 & 0.96 & 68.96 & 0.92 \\
5                 & 99.35 & 1.00 & 90.14 & 0.93 & 91.52 & 0.98 & 68.20 & 0.93 \\
10                & 99.31 & 1.00 & 89.92 & 0.94 & 91.39 & 0.99 & 68.33 & 0.93 \\
\bottomrule
\end{tabular}
}
\end{table}

\begin{figure}[t]
    \centering
    \begin{subfigure}{0.48\columnwidth}
        \centering
        \includegraphics[width=\textwidth]{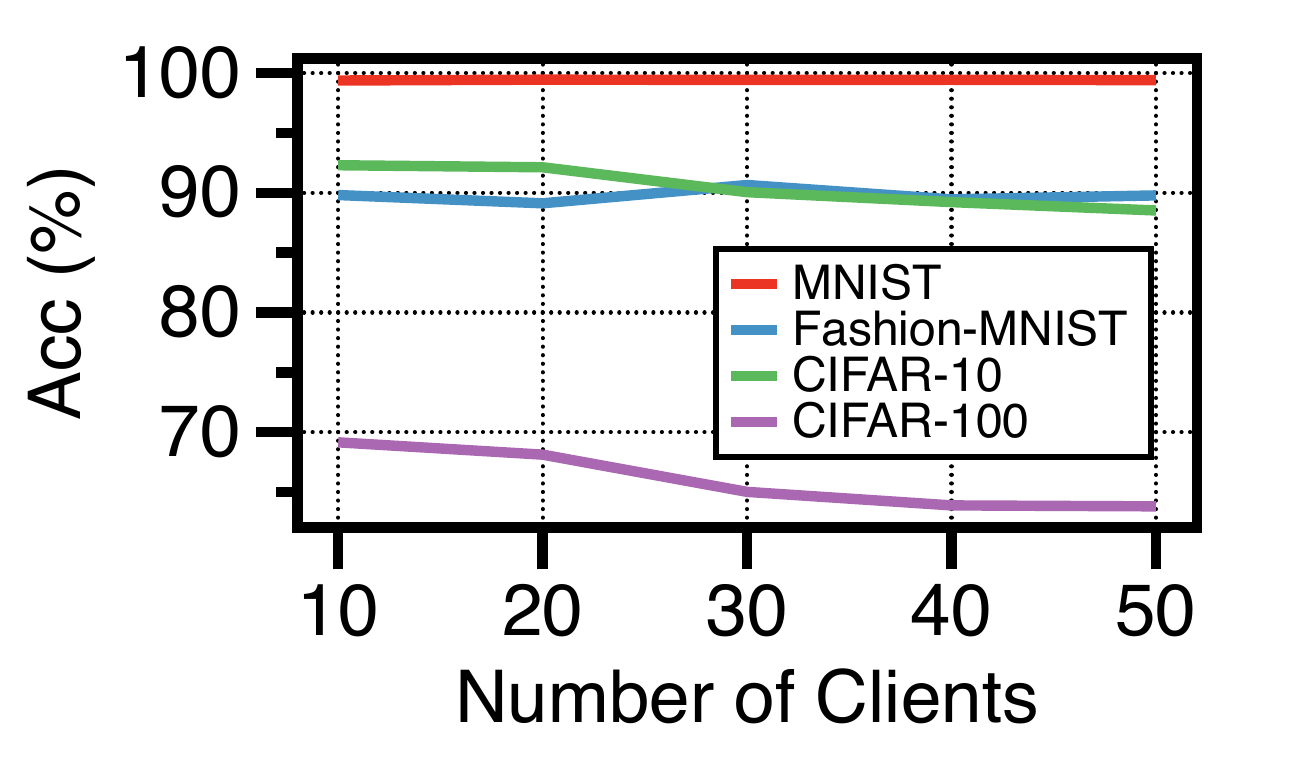}
        \caption{Model Accuracy}
        \label{fig:Acc-numClient}
    \end{subfigure}
    \begin{subfigure}{0.48\columnwidth}
        \centering
        \includegraphics[width=\textwidth]{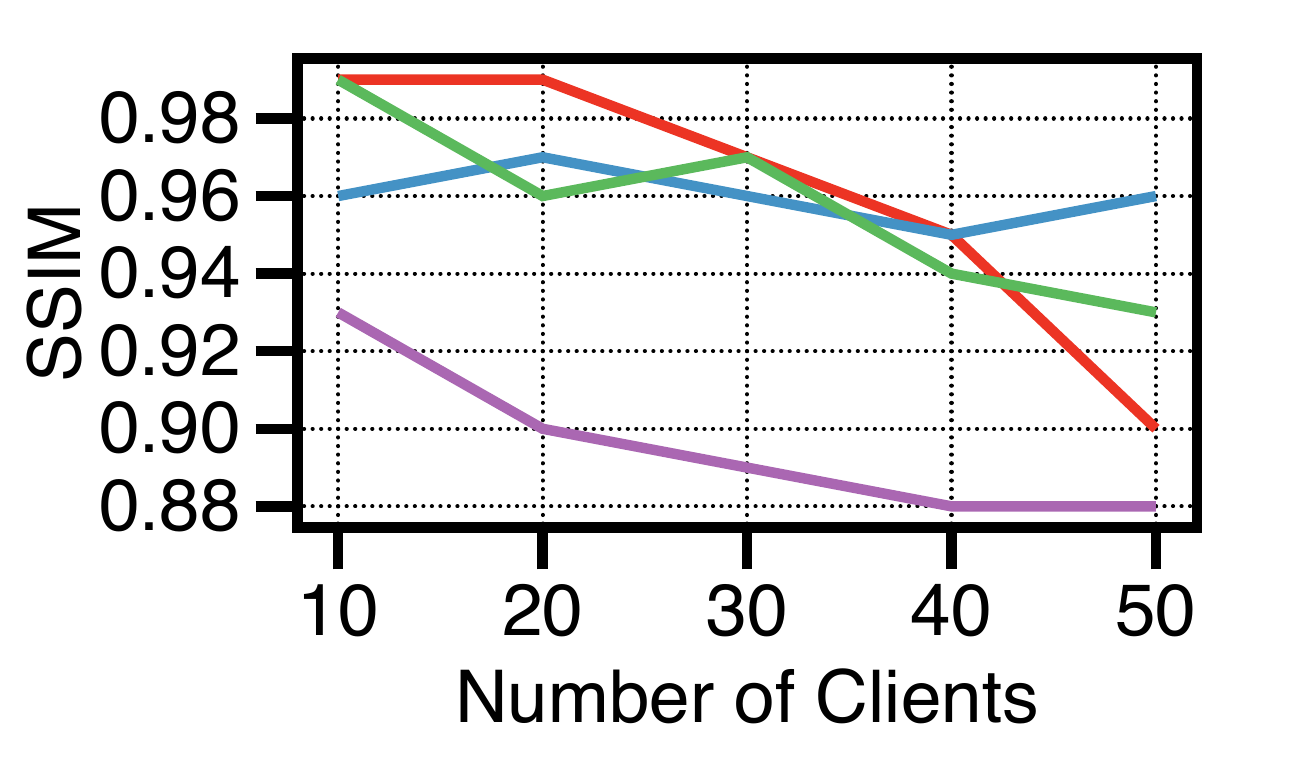}
        \caption{Watermark SSIM}
        \label{fig:SSIM-numClient}
    \end{subfigure}
    \caption{Impact of client numbers}
    \label{fig:numClient}
\end{figure}

\begin{figure}[t]
	\centering
     \begin{minipage}[c]{0.02\columnwidth}
     	\centering
     	\rotatebox{90}{\tiny{\textbf{MNIST}}}
    \end{minipage}%
    \begin{minipage}[c]{0.31\columnwidth} 
        \centering
        \caption*{\scriptsize{10 Clients}}\label{fig:MNIST-Agg-C10}
        \vspace{-3mm}
        \includegraphics[width=\columnwidth]{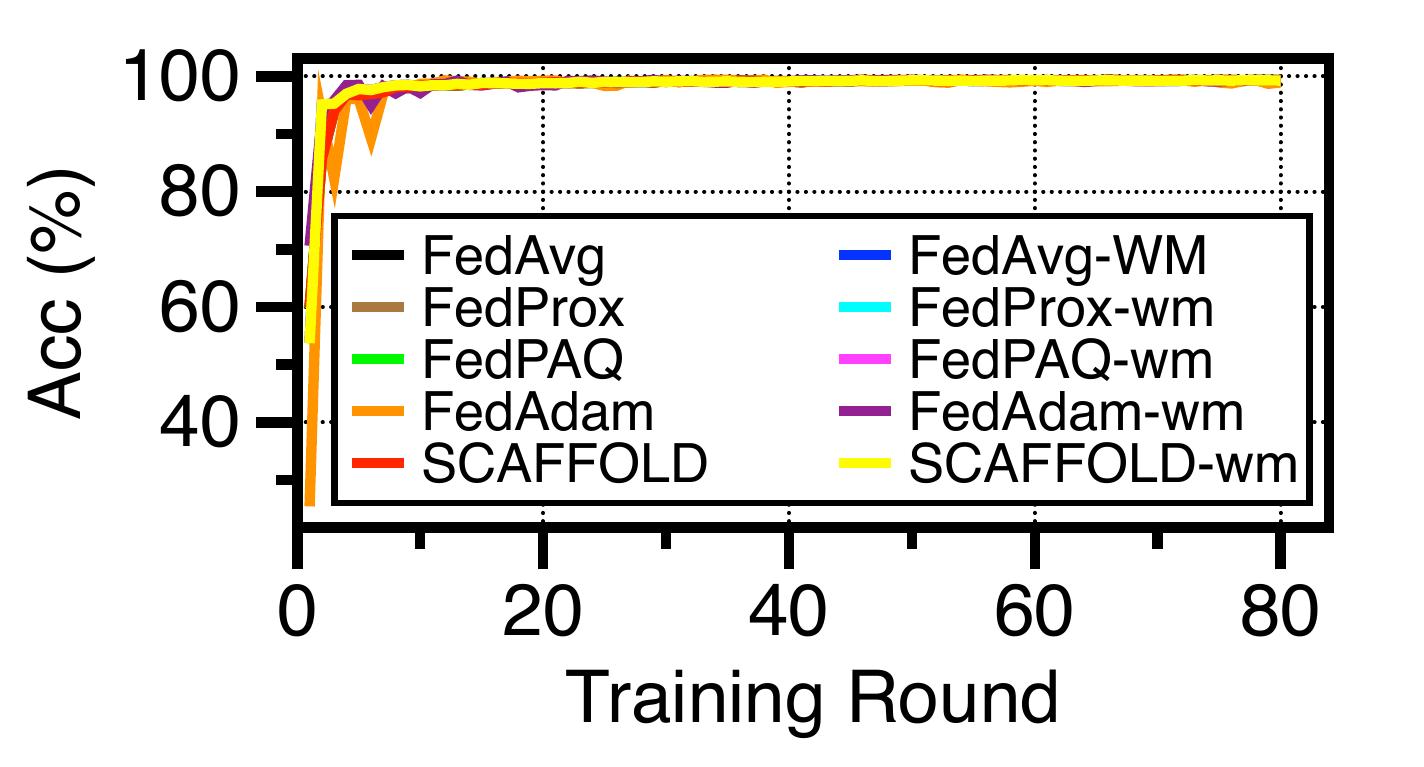} 
    \end{minipage}
    \begin{minipage}[c]{0.31\columnwidth} 
        \centering
        \caption*{\scriptsize{20 Clients}}
        \vspace{-3mm}
        \includegraphics[width=\columnwidth]{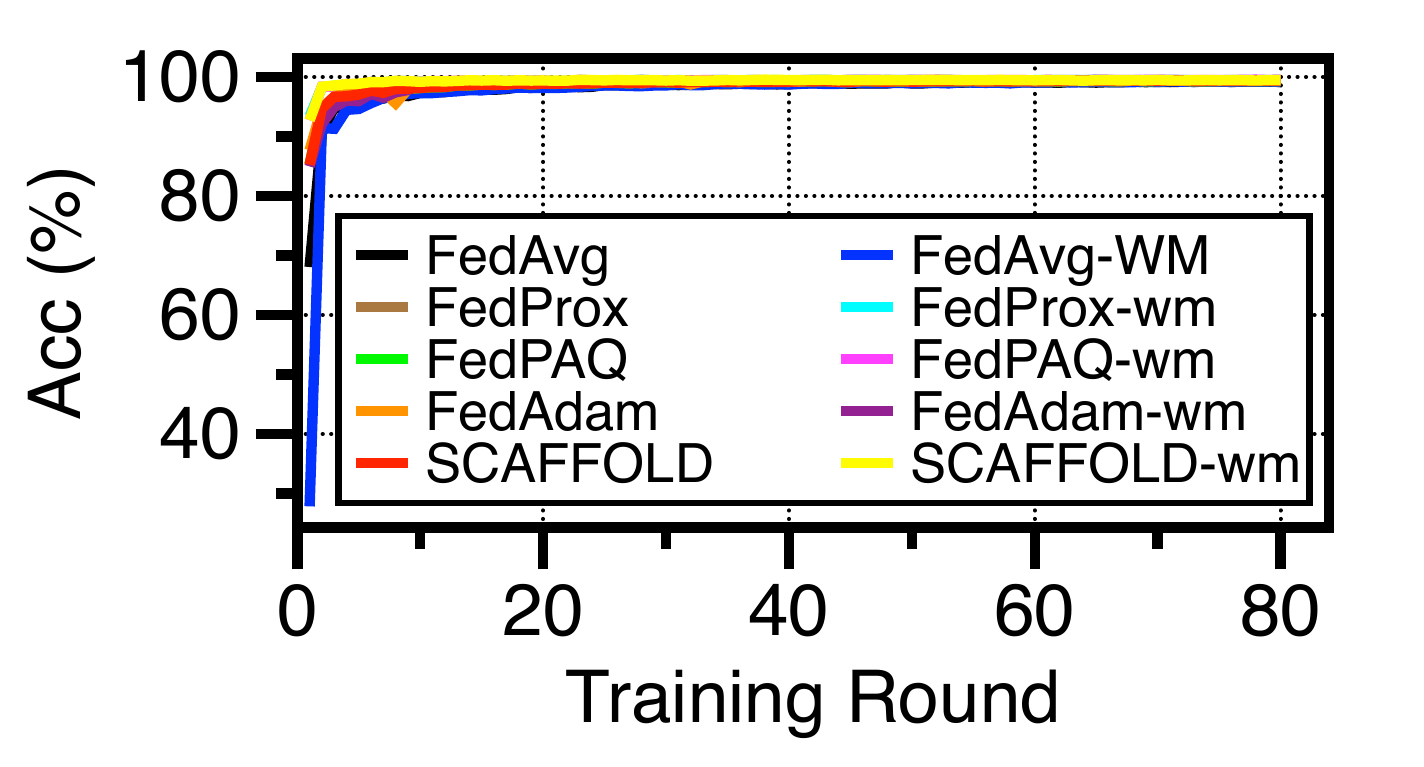} 
    \end{minipage}
    \begin{minipage}[c]{0.31\columnwidth} 
        \centering
        \caption*{\scriptsize{SSIM}}
        \vspace{-3mm}
        \includegraphics[width=\columnwidth]{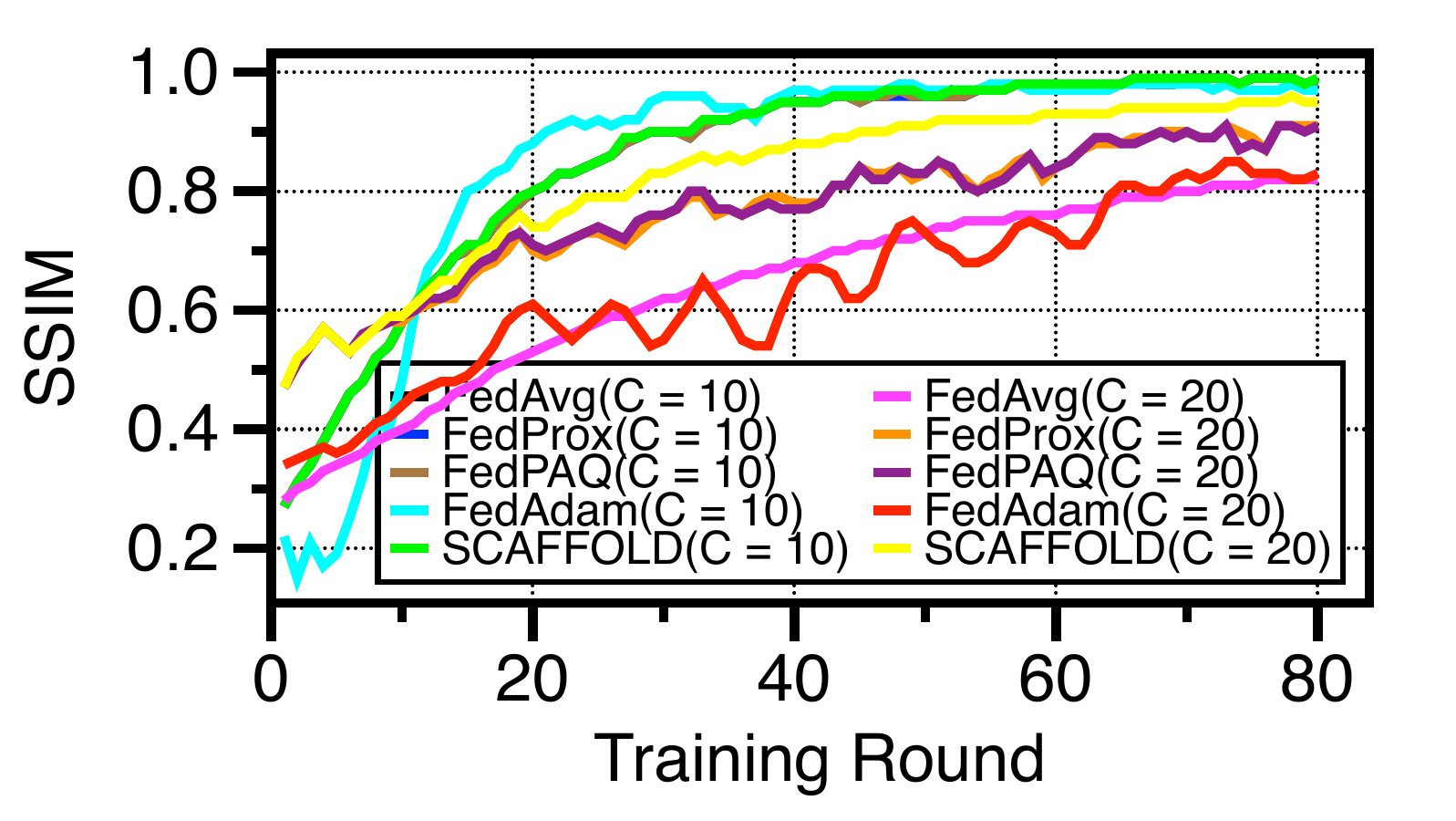} 
    \end{minipage}
    \begin{minipage}[c]{0.02\columnwidth}
     	\centering
     	\rotatebox{90}{\tiny{\textbf{Fashion-MNIST}}}
    \end{minipage}%
    \begin{minipage}[c]{0.31\columnwidth} 
        \centering
        \includegraphics[width=\columnwidth]{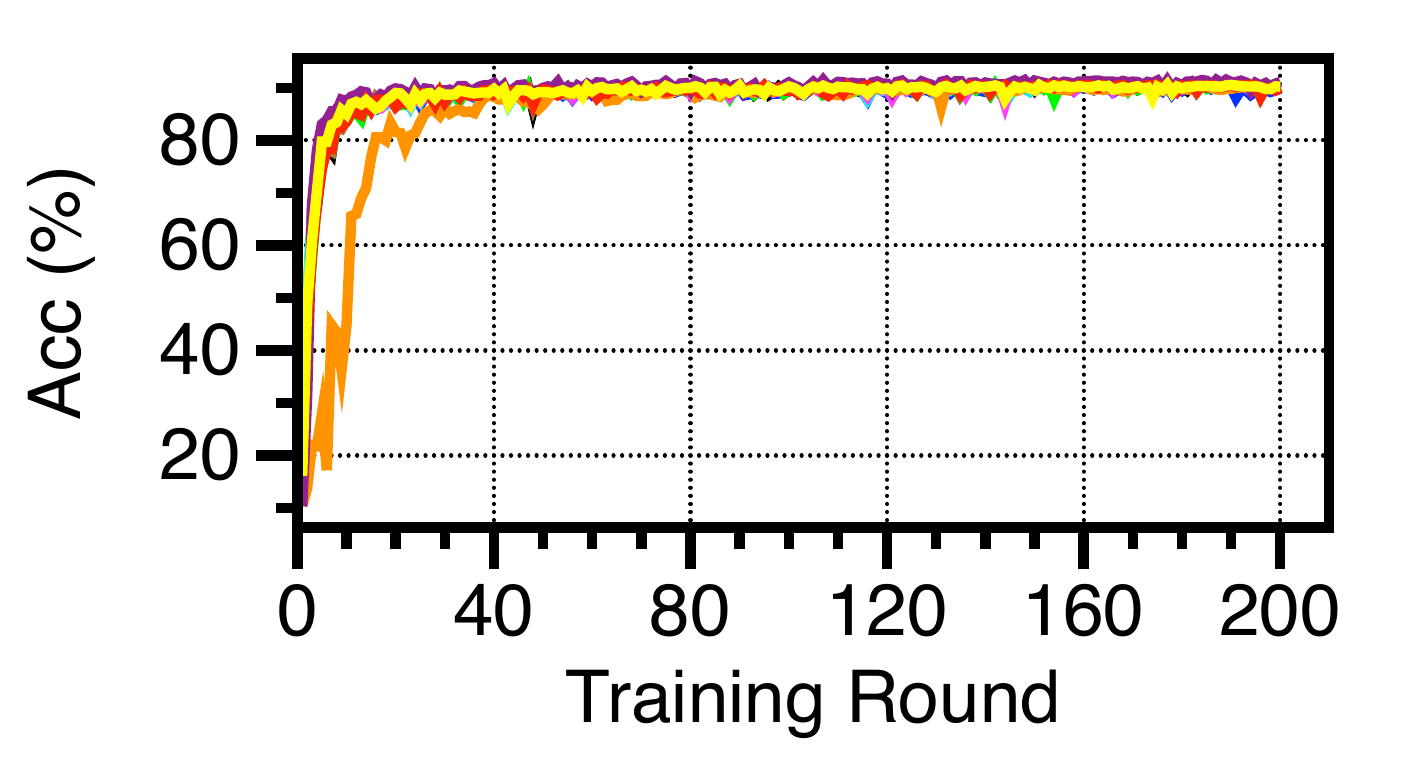} 
    \end{minipage}
    \begin{minipage}[c]{0.31\columnwidth} 
        \centering
        \includegraphics[width=\columnwidth]{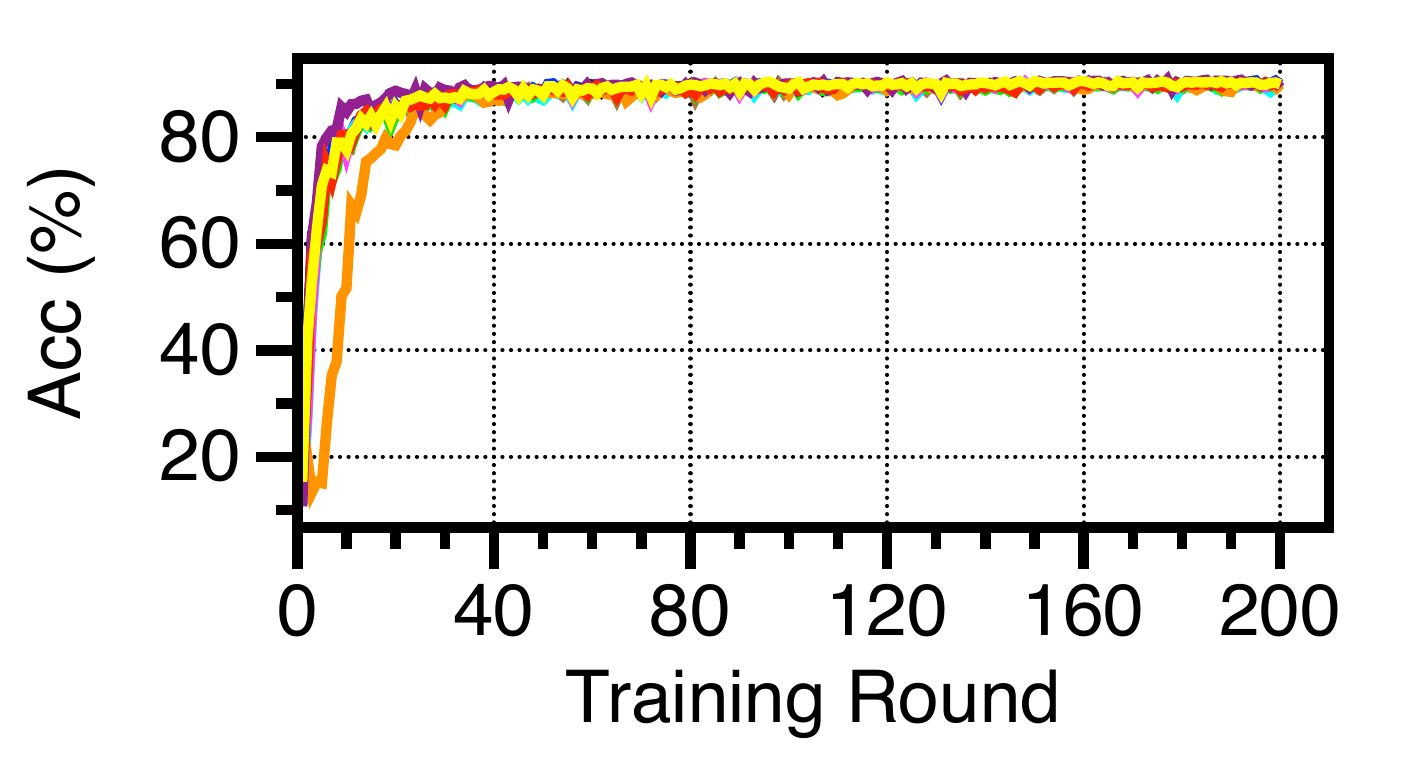} 
    \end{minipage}
    \begin{minipage}[c]{0.31\columnwidth} 
        \centering
        \includegraphics[width=\columnwidth]{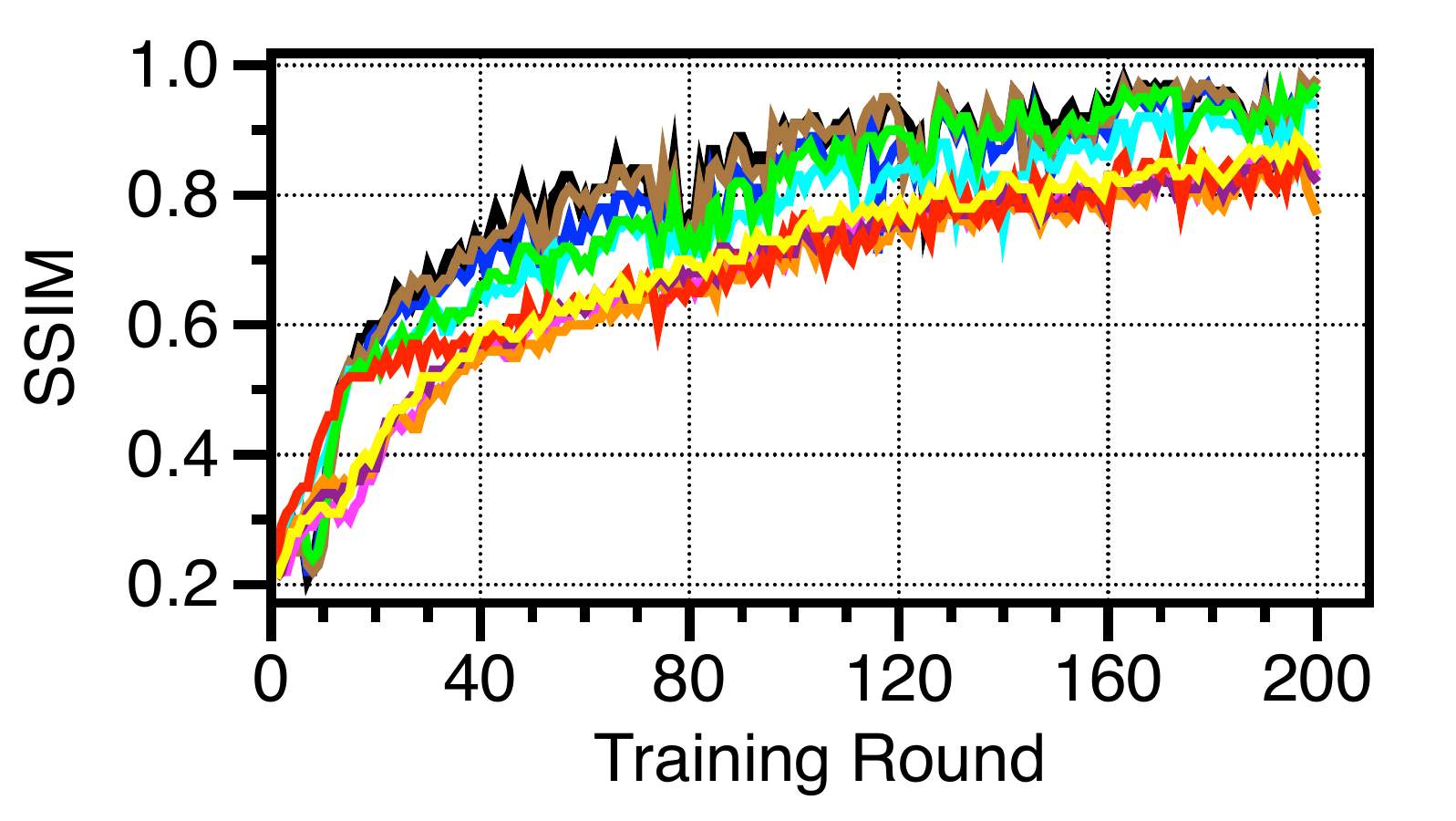} 
    \end{minipage}
    \begin{minipage}[c]{0.02\columnwidth}
     	\centering
     	\rotatebox{90}{\tiny{\textbf{CIFAR-10}}}
    \end{minipage}%
    \begin{minipage}[c]{0.31\columnwidth} 
        \centering
        \includegraphics[width=\columnwidth]{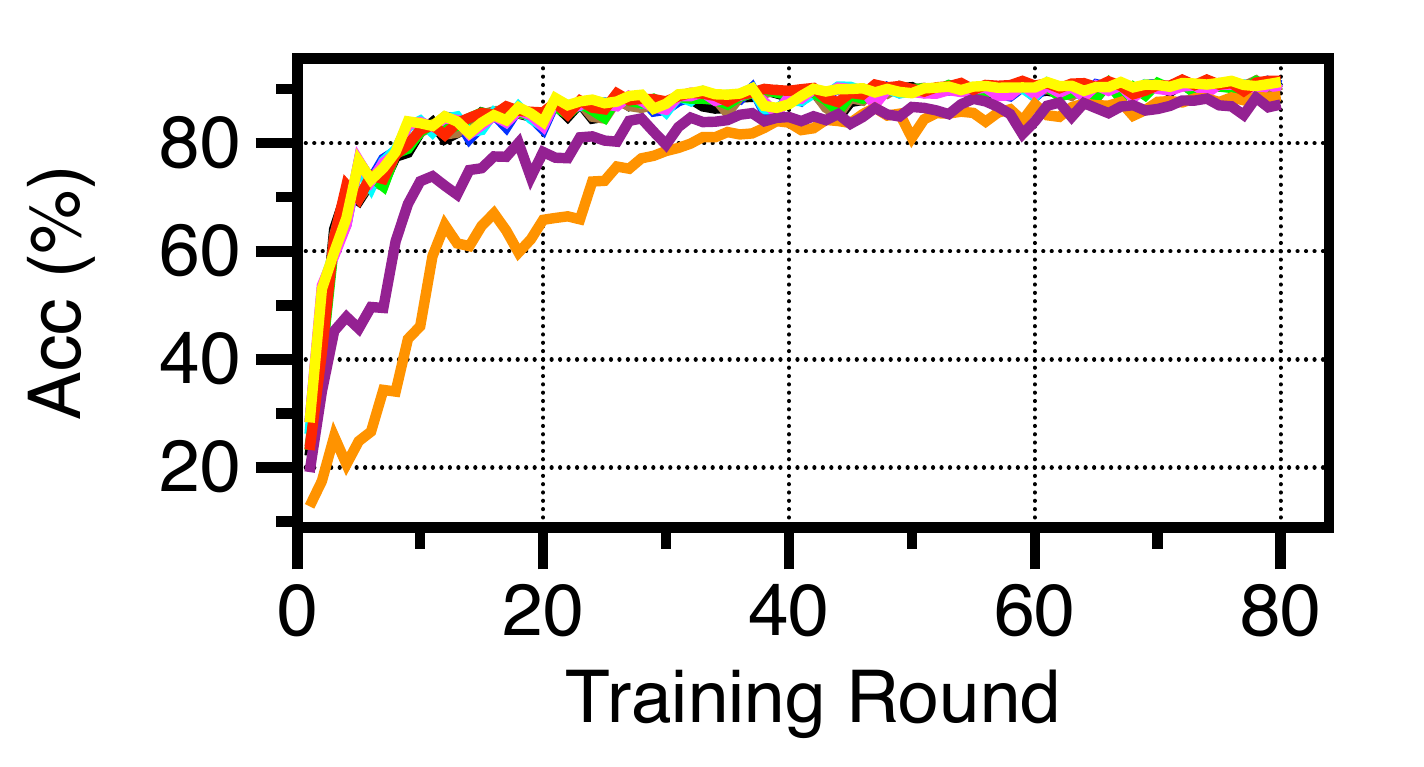} 
    \end{minipage}
    \begin{minipage}[c]{0.31\columnwidth} 
        \centering
        \includegraphics[width=\columnwidth]{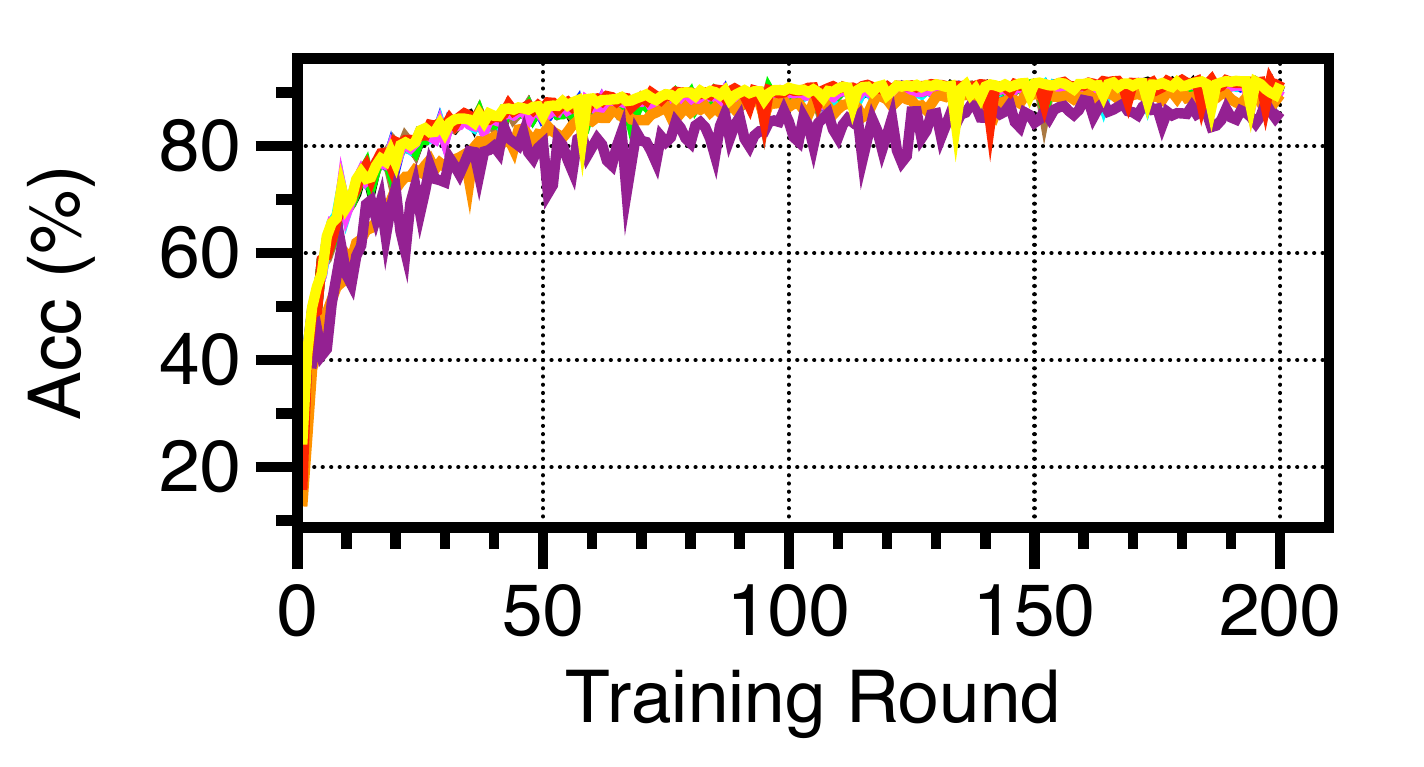} 
    \end{minipage}
    \begin{minipage}[c]{0.31\columnwidth} 
        \centering
        \includegraphics[width=\columnwidth]{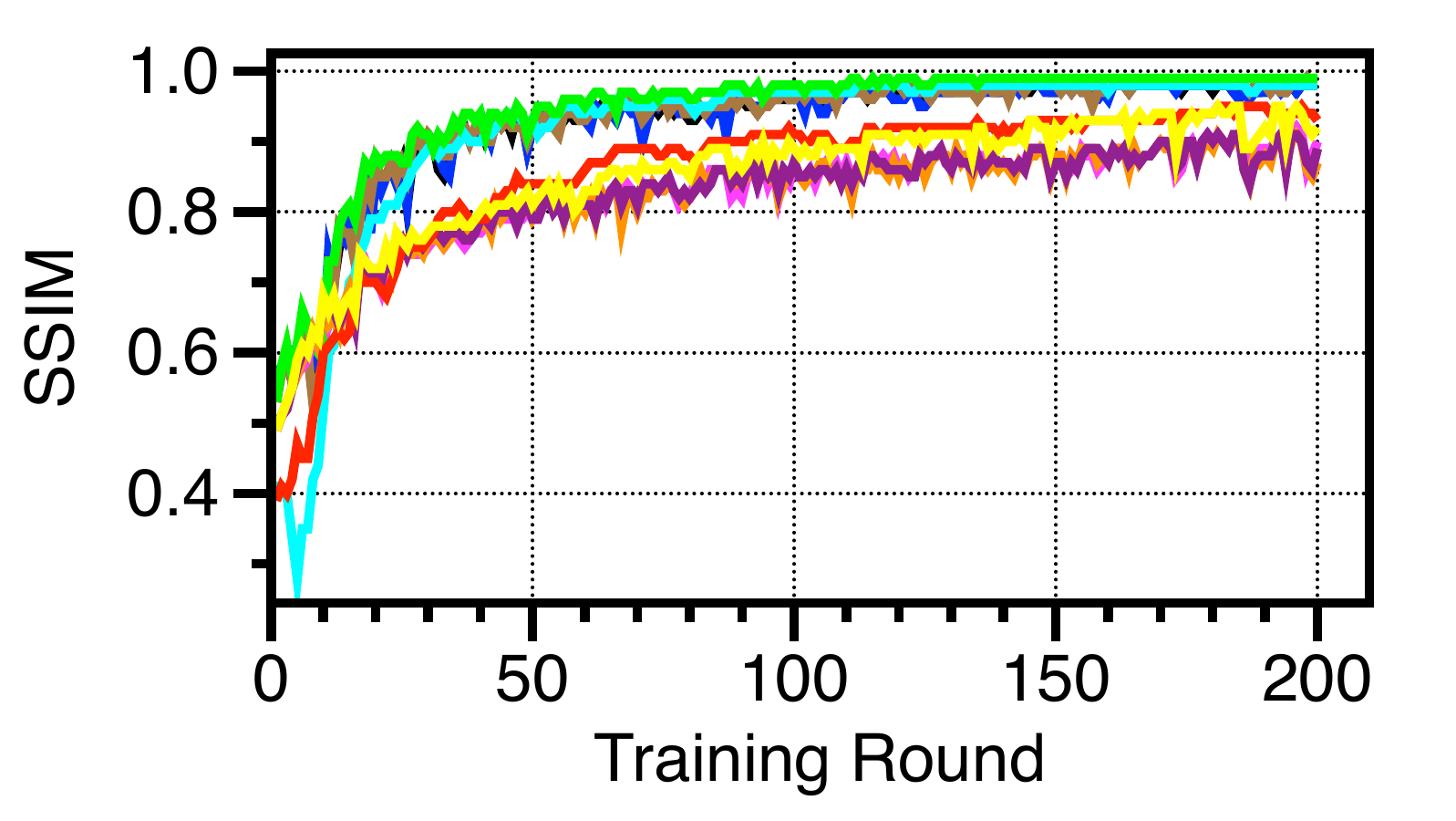} 
    \end{minipage}
    \begin{minipage}[c]{0.02\columnwidth}
     	\centering
     	\rotatebox{90}{\tiny{\textbf{CIFAR-100}}}
    \end{minipage}%
    \begin{minipage}[c]{0.31\columnwidth} 
        \centering
        \includegraphics[width=\columnwidth]{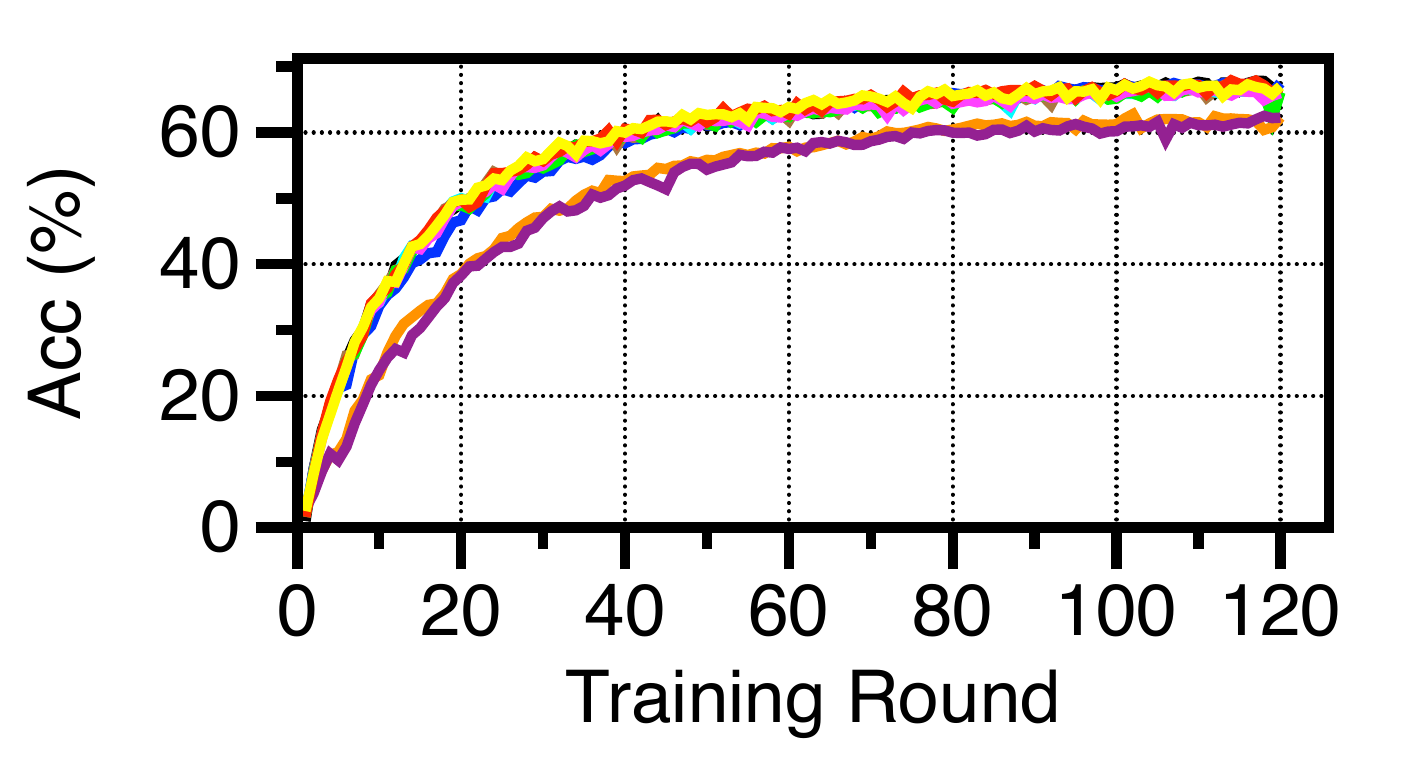} 
    \end{minipage}
    \begin{minipage}[c]{0.31\columnwidth} 
        \centering
        \includegraphics[width=\columnwidth]{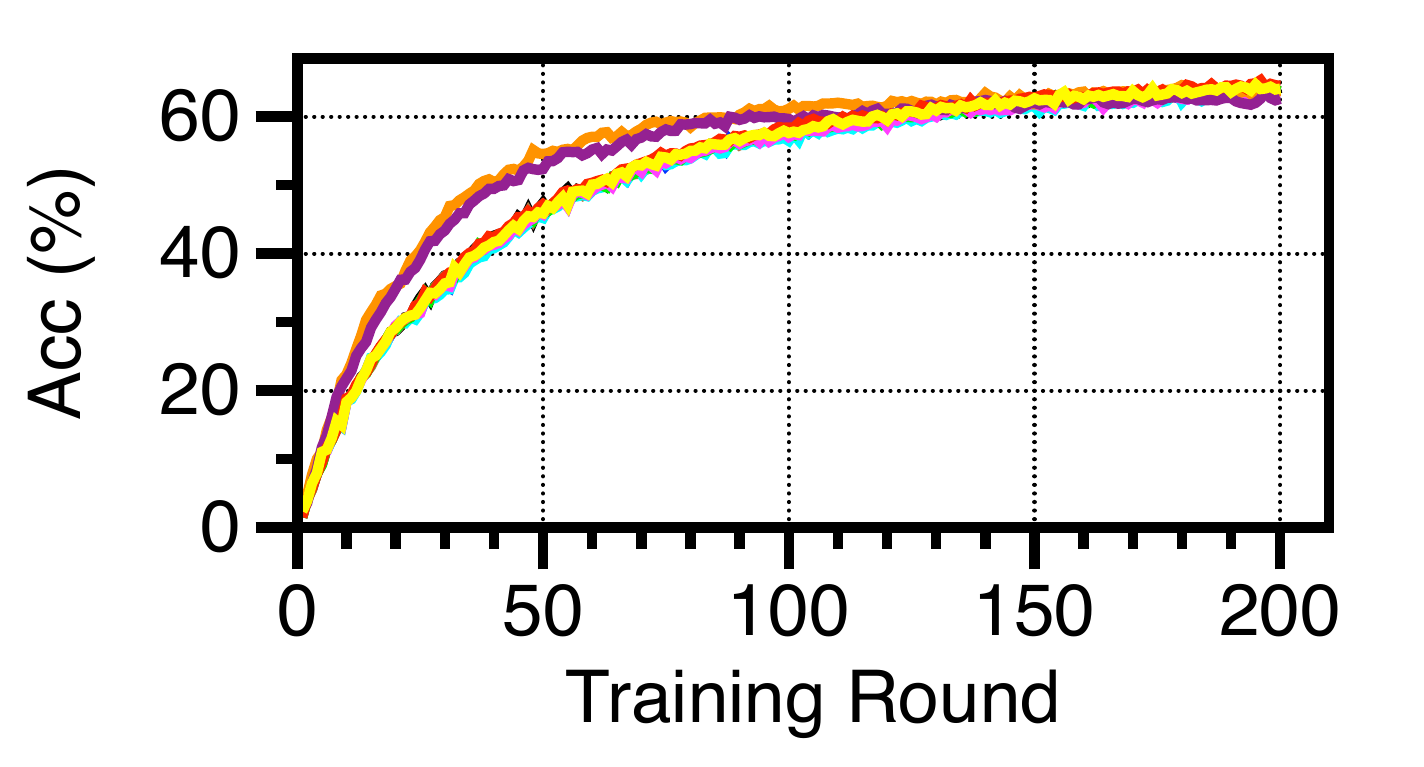} 
    \end{minipage}
    \begin{minipage}[c]{0.31\columnwidth} 
        \centering
        \includegraphics[width=\columnwidth]{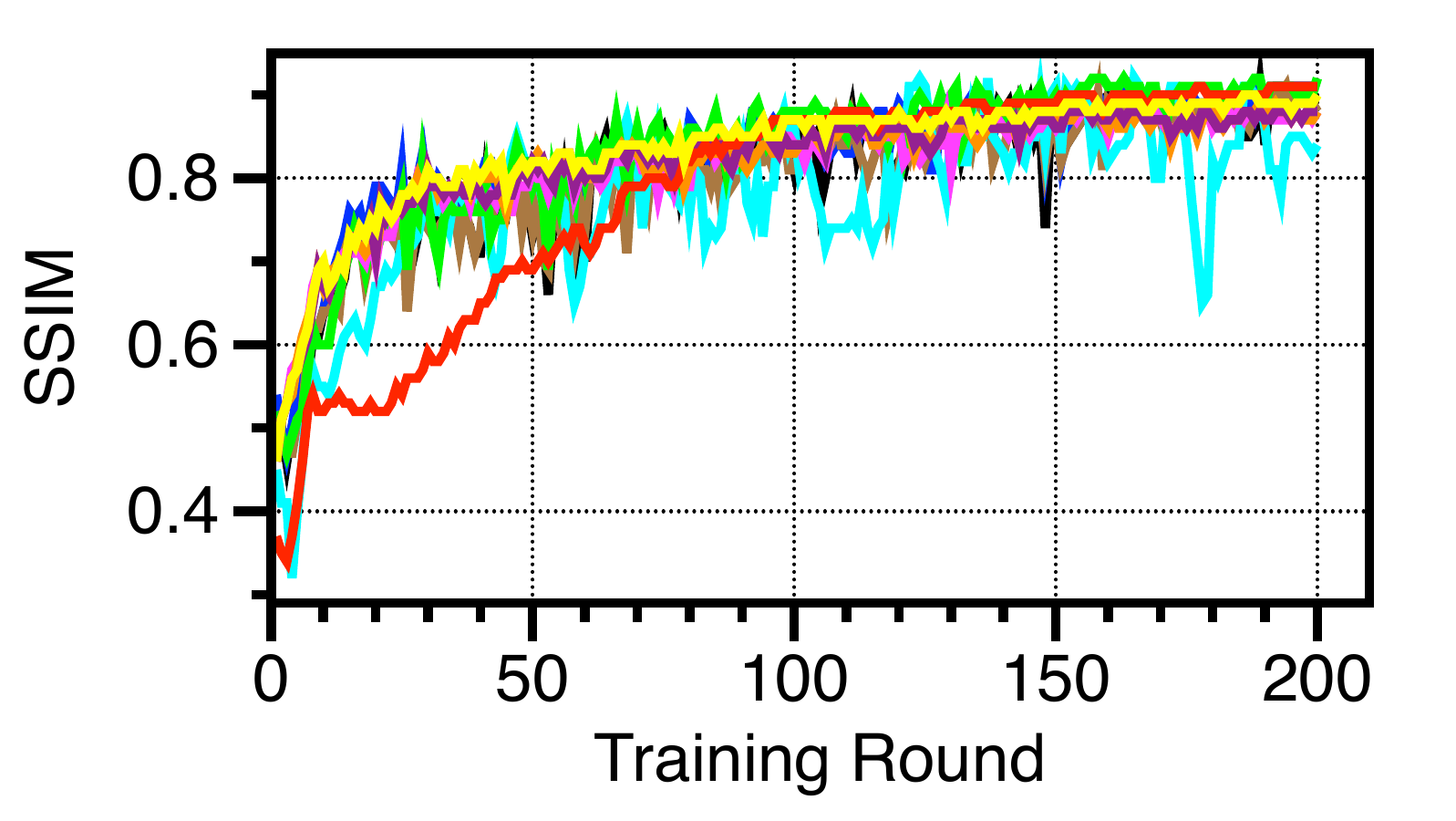} 
    \end{minipage}
    \caption{Performance with aggregation schemes}
	\label{fig:Agg}  
\end{figure}

\noindent \textbf{Impact of $y$ and $m$.}
\Cref{tab:ym} summarizes the effects of varying sample weight $y$ and margin $m$ in the contrastive loss function (in \Cref{eq:cl}) on model accuracy and watermark fidelity. The results indicate that $y$ and $m$ moderately influence learning stability. Smaller datasets (e.g., MNIST) exhibit greater robustness to parameter variations than more complex datasets (e.g., CIFAR-100). For MNIST and CIFAR-10, accuracy remains consistently high (above 99\% and 89\%, respectively) across all settings, indicating that the contrastive objective effectively preserves discriminative capability across diverse configurations. By contrast, CIFAR-100 shows more sensitivity, where intermediate $m$ (e.g., $m=0.5$) and $y$ (e.g., $y=0.3$) achieve a balanced trade-off between optimization stability and generalization, achieving 68.59\% accuracy and 0.93 SSIM. 

\noindent \textbf{Impact of $\lambda$.}
We analyze the influence of the weighting coefficient $\lambda$ in \Cref{tab:lambda}. $\lambda$ controls the trade-off between the main-task loss and the watermark-related loss. As observed, variations in $\lambda$ have negligible influence on model accuracy, which remains stable across datasets (e.g., 99\% for MNIST and 91\% for CIFAR-10). This indicates that incorporating watermark learning does not compromise main-task performance. In contrast,  SSIM increases with larger $\lambda$ values. For instance, on MNIST, SSIM rises from 0.38 at $\lambda=0.1$ to 1.0 when $\lambda \ge 5$, while CIFAR-100 demonstrates a similar increase from 0.83 to 0.93. These demonstrate that larger $\lambda$ values enhance watermark-embedding quality without destabilizing the global model. Empirically, a moderate $\lambda$ (e.g., 1) achieves a balanced trade-off between task performance and watermark fidelity, enabling reliable extraction with minimal accuracy loss.

\noindent \textbf{Impact of Client Numbers.}
We vary the number of clients from 10 to 50 to evaluate scalability. As shown in  \Cref{fig:numClient}, \texttt{FLClear} sustains high task performance and watermark fidelity even as the number of participating clients increases. Specifically, model accuracy remains stable across datasets, with MNIST and Fashion-MNIST exceeding 95\% and CIFAR-10 maintaining above 85\%, while CIFAR-100 exhibits a moderate decline owing to higher data heterogeneity. Regarding watermark fidelity, SSIM values stay above 0.9 for most datasets, indicating robust watermark reconstruction as client numbers grow. A slight SSIM reduction for CIFAR-100 (from 0.91 to 0.88) suggests that greater client diversity slightly affects embedding stability.

\subsection{Compatibility with Aggregation Schemes}
\label{sec:aggregation}

\begin{figure}[t]
	\centering
	\begin{minipage}[c]{0.02\columnwidth}
     	\centering
        \vspace{8mm}
     	\rotatebox{90}{\tiny{\textbf{MNIST}}}
    \end{minipage}%
    \begin{minipage}[c]{0.14\columnwidth} 
        \centering
        \caption*{\scriptsize{Original}}
        \vspace{-1mm}
        \includegraphics[width=\columnwidth]{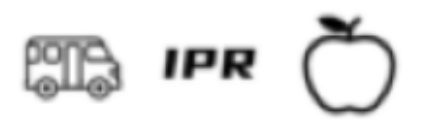}
        \scriptsize{SSIM $\uparrow$} \\ 
     	\scriptsize{MSE $\downarrow$} \\
     	\scriptsize{PSNR $\uparrow$} \\
     	\scriptsize{LPIPS $\downarrow$} \\ 
    \end{minipage}
    \begin{minipage}[c]{0.14\columnwidth} 
        \centering
        \caption*{\scriptsize{FedAvg}}
        \vspace{-1mm}
        \includegraphics[width=\columnwidth]{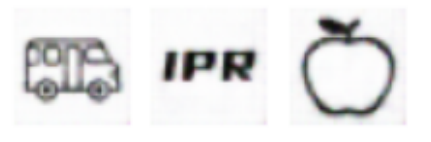}
        \scriptsize{0.984} \\
     	\scriptsize{0.002} \\
     	\scriptsize{26.697} \\
     	\scriptsize{0.018} \\ 
    \end{minipage}
    \begin{minipage}[c]{0.14\columnwidth} 
        \centering
        \caption*{\scriptsize{FedProx}}
        \vspace{-1mm}
        \includegraphics[width=\columnwidth]{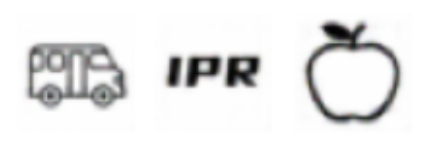}
        \scriptsize{0.994} \\
     	\scriptsize{0.001} \\
     	\scriptsize{31.133} \\
     	\scriptsize{0.004} \\  
    \end{minipage}
    \begin{minipage}[c]{0.14\columnwidth} 
        \centering
        \caption*{\scriptsize{FedPAQ}}
        \vspace{-1mm}
        \includegraphics[width=\columnwidth]{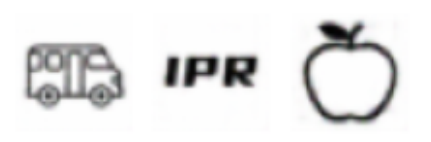}
        \scriptsize{0.994} \\
     	\scriptsize{0.001} \\
     	\scriptsize{30.276} \\
     	\scriptsize{0.047} \\  
    \end{minipage}
    \begin{minipage}[c]{0.14\columnwidth} 
        \centering
        \caption*{\scriptsize{FedADAM}}
        \vspace{-1mm}
        \includegraphics[width=\columnwidth]{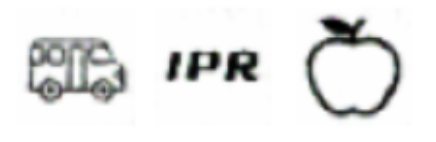}
        \scriptsize{0.973} \\
     	\scriptsize{0.004} \\
     	\scriptsize{23.790} \\
     	\scriptsize{0.026} \\  
    \end{minipage}
    \begin{minipage}[c]{0.14\columnwidth} 
        \centering
        \caption*{\scriptsize{SCAFFOLD}}
        \vspace{-1mm}
        \includegraphics[width=\columnwidth]{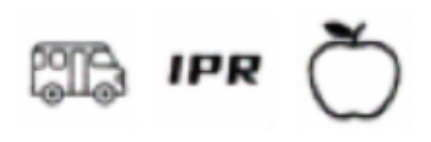}
        \scriptsize{0.994} \\
     	\scriptsize{0.001} \\
     	\scriptsize{30.307} \\
     	\scriptsize{0.005} \\  
    \end{minipage}
	\begin{minipage}[c]{0.02\columnwidth}
     	\centering
     	\rotatebox{90}{\tiny{\textbf{Fashion-MNIST}}}
    \end{minipage}%
    \begin{minipage}[c]{0.14\columnwidth} 
        \centering
        \vspace{1mm}
        \includegraphics[width=\columnwidth]{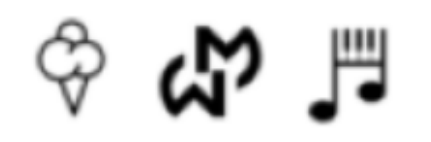}
        \scriptsize{SSIM} \\ 
     	\scriptsize{MSE} \\
     	\scriptsize{PSNR} \\
     	\scriptsize{LPIPS} \\ 
    \end{minipage}
    \begin{minipage}[c]{0.14\columnwidth} 
        \centering
        \vspace{1mm}
        \includegraphics[width=\columnwidth]{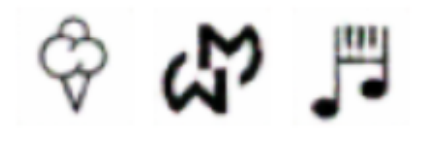}
        \scriptsize{0.990} \\
     	\scriptsize{0.002} \\
     	\scriptsize{27.810} \\
     	\scriptsize{0.011} \\ 
    \end{minipage}
    \begin{minipage}[c]{0.14\columnwidth} 
        \centering
        \vspace{1mm}
        \includegraphics[width=\columnwidth]{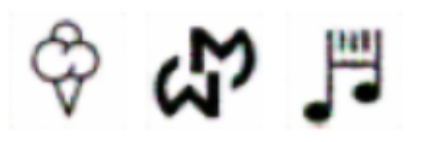}
        \scriptsize{0.988} \\
     	\scriptsize{0.002} \\
     	\scriptsize{27.016} \\
     	\scriptsize{0.016} \\  
    \end{minipage}
    \begin{minipage}[c]{0.14\columnwidth} 
        \centering
        \vspace{1mm}
        \includegraphics[width=\columnwidth]{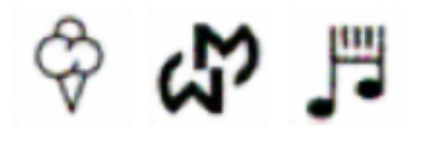}
        \scriptsize{0.991} \\
     	\scriptsize{0.002} \\
     	\scriptsize{27.840} \\
     	\scriptsize{0.013} \\  
    \end{minipage}
    \begin{minipage}[c]{0.14\columnwidth} 
        \centering
        \vspace{1mm}
        \includegraphics[width=\columnwidth]{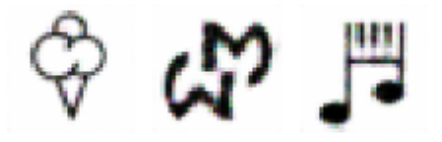}
        \scriptsize{0.956} \\
     	\scriptsize{0.007} \\
     	\scriptsize{24.663} \\
     	\scriptsize{0.035} \\  
    \end{minipage}
    \begin{minipage}[c]{0.14\columnwidth} 
        \centering
        \vspace{1mm}
        \includegraphics[width=\columnwidth]{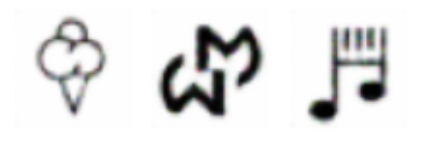}
        \scriptsize{0.981} \\
     	\scriptsize{0.004} \\
     	\scriptsize{24.613} \\
     	\scriptsize{0.015} \\  
    \end{minipage}
	\begin{minipage}[c]{0.02\columnwidth}
     	\centering
     	\rotatebox{90}{\tiny{\textbf{CIFAR-10 (grayscale)}}}
    \end{minipage}%
    \begin{minipage}[c]{0.14\columnwidth} 
        \centering
        \vspace{1mm}
        \includegraphics[width=\columnwidth]{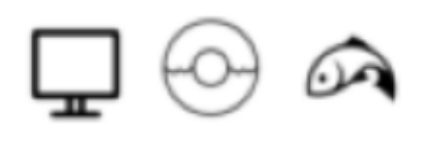}
        \scriptsize{SSIM} \\ 
     	\scriptsize{MSE} \\
     	\scriptsize{PSNR} \\
     	\scriptsize{LPIPS} \\ 
    \end{minipage}
    \begin{minipage}[c]{0.14\columnwidth} 
        \centering
        \vspace{1mm}
        \includegraphics[width=\columnwidth]{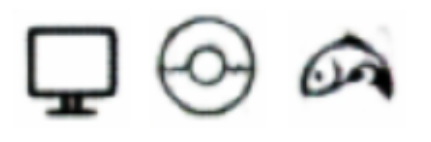}
        \scriptsize{0.952} \\
     	\scriptsize{0.009} \\
     	\scriptsize{22.367} \\
     	\scriptsize{0.037} \\ 
    \end{minipage}
    \begin{minipage}[c]{0.14\columnwidth} 
        \centering
        \vspace{1mm}
        \includegraphics[width=\columnwidth]{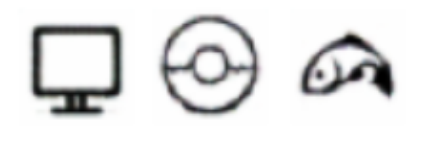}
        \scriptsize{0.967} \\
     	\scriptsize{0.009} \\
     	\scriptsize{26.747} \\
     	\scriptsize{0.033} \\ 
    \end{minipage}
    \begin{minipage}[c]{0.14\columnwidth} 
        \centering
        \vspace{1mm}
        \includegraphics[width=\columnwidth]{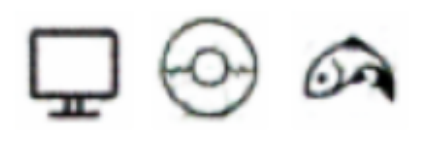}
        \scriptsize{0.965} \\
     	\scriptsize{0.005} \\
     	\scriptsize{23.653} \\
     	\scriptsize{0.035} \\ 
    \end{minipage}
    \begin{minipage}[c]{0.14\columnwidth} 
        \centering
        \vspace{1mm}
        \includegraphics[width=\columnwidth]{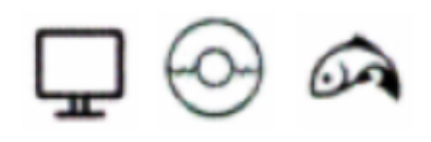}
        \scriptsize{0.977} \\
     	\scriptsize{0.003} \\
     	\scriptsize{25.480} \\
     	\scriptsize{0.016} \\ 
    \end{minipage}
    \begin{minipage}[c]{0.14\columnwidth} 
        \centering
        \vspace{1mm}
        \includegraphics[width=\columnwidth]{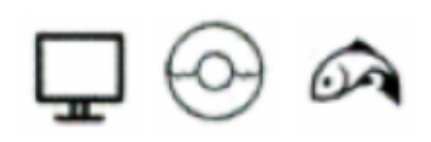}
        \scriptsize{0.983} \\
     	\scriptsize{0.002} \\
     	\scriptsize{26.417} \\
     	\scriptsize{0.009} \\  
    \end{minipage}
	\begin{minipage}[c]{0.02\columnwidth}
     	\centering
     	\rotatebox{90}{\tiny{\textbf{CIFAR-10 (color)}}}
    \end{minipage}%
    \begin{minipage}[c]{0.14\columnwidth} 
        \centering
        \vspace{1mm}
        \includegraphics[width=\columnwidth]{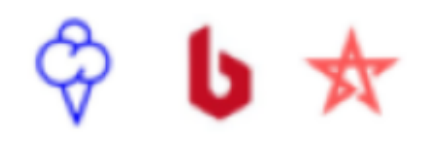}
        \scriptsize{SSIM} \\ 
     	\scriptsize{MSE} \\
     	\scriptsize{PSNR} \\
     	\scriptsize{LPIPS} \\ 
    \end{minipage}
    \begin{minipage}[c]{0.14\columnwidth} 
        \centering
        \vspace{1mm}
        \includegraphics[width=\columnwidth]{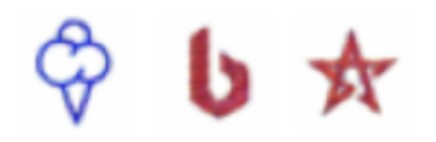}
        \scriptsize{0.902} \\
     	\scriptsize{0.006} \\
     	\scriptsize{23.170} \\
     	\scriptsize{0.057} \\ 
    \end{minipage}
    \begin{minipage}[c]{0.14\columnwidth} 
        \centering
        \vspace{1mm}
        \includegraphics[width=\columnwidth]{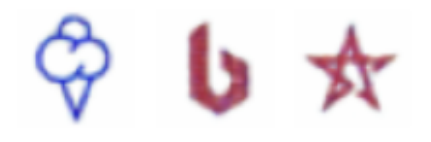}
        \scriptsize{0.893} \\
     	\scriptsize{0.007} \\
     	\scriptsize{22.303} \\
     	\scriptsize{0.076} \\ 
    \end{minipage}
    \begin{minipage}[c]{0.14\columnwidth} 
        \centering
        \vspace{1mm}
        \includegraphics[width=\columnwidth]{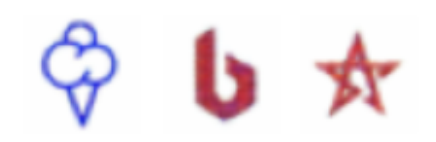}
        \scriptsize{0.919} \\
     	\scriptsize{0.004} \\
     	\scriptsize{24.950} \\
     	\scriptsize{0.051} \\ 
    \end{minipage}
    \begin{minipage}[c]{0.14\columnwidth} 
        \centering
        \vspace{1mm}
        \includegraphics[width=\columnwidth]{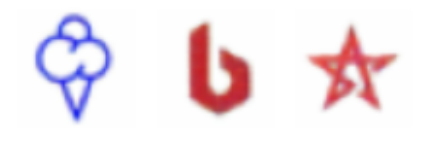}
        \scriptsize{0.915} \\
     	\scriptsize{0.009} \\
     	\scriptsize{23.617} \\
     	\scriptsize{0.076} \\ 
    \end{minipage}
    \begin{minipage}[c]{0.14\columnwidth} 
        \centering
        \vspace{1mm}
        \includegraphics[width=\columnwidth]{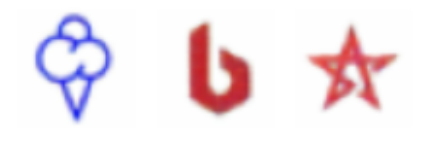}
        \scriptsize{0.945} \\
     	\scriptsize{0.007} \\
     	\scriptsize{27.750} \\
     	\scriptsize{0.017} \\  
    \end{minipage}
	\begin{minipage}[c]{0.02\columnwidth}
     	\centering
     	\rotatebox{90}{\tiny{\textbf{CIFAR-100 (grayscale)}}}
    \end{minipage}%
    \begin{minipage}[c]{0.14\columnwidth} 
        \centering
        \vspace{1mm}
        \includegraphics[width=\columnwidth]{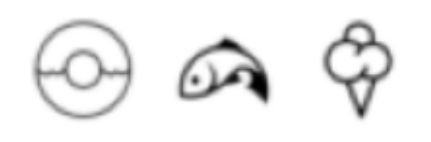}
        \scriptsize{SSIM} \\ 
     	\scriptsize{MSE} \\
     	\scriptsize{PSNR} \\
     	\scriptsize{LPIPS} \\ 
    \end{minipage}
    \begin{minipage}[c]{0.14\columnwidth} 
        \centering
        \vspace{1mm}
        \includegraphics[width=\columnwidth]{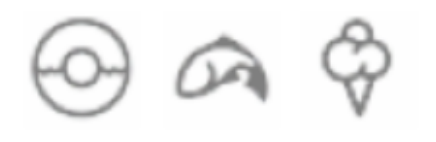}
        \scriptsize{0.965} \\
     	\scriptsize{0.005} \\
     	\scriptsize{23.653} \\
     	\scriptsize{0.035} \\ 
    \end{minipage}
    \begin{minipage}[c]{0.14\columnwidth} 
        \centering
        \vspace{1mm}
        \includegraphics[width=\columnwidth]{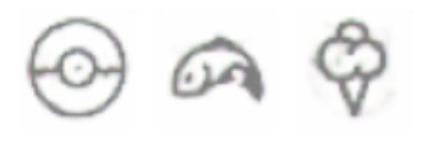}
        \scriptsize{0.936} \\
     	\scriptsize{0.007} \\
     	\scriptsize{22.417} \\
     	\scriptsize{0.033} \\ 
    \end{minipage}
    \begin{minipage}[c]{0.14\columnwidth} 
        \centering
        \vspace{1mm}
        \includegraphics[width=\columnwidth]{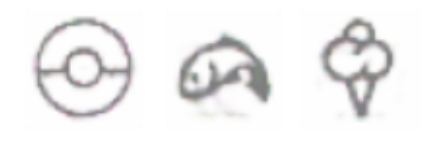}
        \scriptsize{0.953} \\
     	\scriptsize{0.007} \\
     	\scriptsize{22.767} \\
     	\scriptsize{0.042} \\ 
    \end{minipage}
    \begin{minipage}[c]{0.14\columnwidth} 
        \centering
        \vspace{1mm}
        \includegraphics[width=\columnwidth]{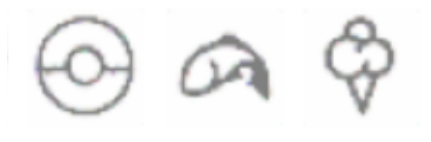}
        \scriptsize{0.939} \\
     	\scriptsize{0.007} \\
     	\scriptsize{21.943} \\
     	\scriptsize{0.039} \\ 
    \end{minipage}
    \begin{minipage}[c]{0.14\columnwidth} 
        \centering
        \vspace{1mm}
        \includegraphics[width=\columnwidth]{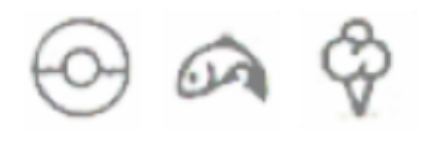}
        \scriptsize{0.953} \\
     	\scriptsize{0.006} \\
     	\scriptsize{22.983} \\
     	\scriptsize{0.030} \\  
    \end{minipage}
	\begin{minipage}[c]{0.02\columnwidth}
     	\centering
     	\rotatebox{90}{\tiny{\textbf{CIFAR-100 (color)}}} 	
    \end{minipage}%
    \begin{minipage}[c]{0.14\columnwidth} 
        \centering
        \vspace{1mm}
        \includegraphics[width=\columnwidth]{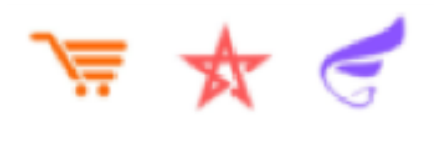}
        \scriptsize{SSIM} \\ 
     	\scriptsize{MSE} \\
     	\scriptsize{PSNR} \\
     	\scriptsize{LPIPS} \\ 
    \end{minipage}
    \begin{minipage}[c]{0.14\columnwidth} 
        \centering
        \vspace{1mm}
        \includegraphics[width=\columnwidth]{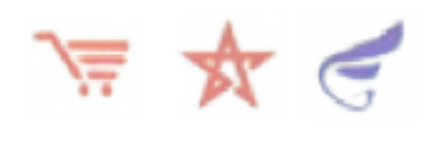}
        \scriptsize{0.885} \\
     	\scriptsize{0.006} \\
     	\scriptsize{22.550} \\
     	\scriptsize{0.080} \\ 
    \end{minipage}
    \begin{minipage}[c]{0.14\columnwidth} 
        \centering
        \vspace{1mm}
        \includegraphics[width=\columnwidth]{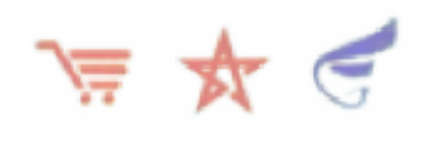}
        \scriptsize{0.888} \\
     	\scriptsize{0.006} \\
     	\scriptsize{22.630} \\
     	\scriptsize{0.046} \\ 
    \end{minipage}
    \begin{minipage}[c]{0.14\columnwidth} 
        \centering
        \vspace{1mm}
        \includegraphics[width=\columnwidth]{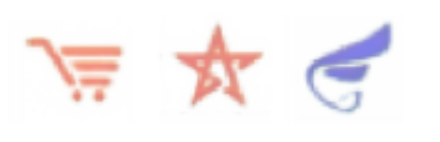}
        \scriptsize{0.909} \\
     	\scriptsize{0.005} \\
     	\scriptsize{23.183} \\
     	\scriptsize{0.034} \\ 
    \end{minipage}
    \begin{minipage}[c]{0.14\columnwidth} 
        \centering
        \vspace{1mm}
        \includegraphics[width=\columnwidth]{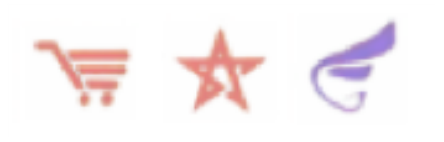}
        \scriptsize{0.898} \\
     	\scriptsize{0.005} \\
     	\scriptsize{23.413} \\
     	\scriptsize{0.032} \\ 
    \end{minipage}
    \begin{minipage}[c]{0.14\columnwidth} 
        \centering
        \vspace{1mm}
        \includegraphics[width=\columnwidth]{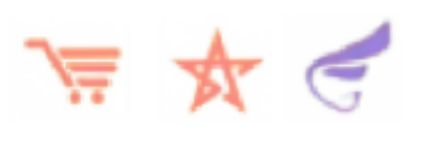}
        \scriptsize{0.931} \\
     	\scriptsize{0.006} \\
     	\scriptsize{22.410} \\
     	\scriptsize{0.043} \\  
    \end{minipage}
\caption{Watermark visualization with aggregation schemes}
\label{fig:w-ar}
\end{figure}

\begin{figure}[t]
	\centering
	\begin{minipage}[c]{0.02\columnwidth}
     	\centering
     	\rotatebox{90}{\tiny{\textbf{MNIST}}}
    \end{minipage}%
    \begin{minipage}[c]{0.22\columnwidth} 
        \centering
        \caption*{\scriptsize{Pruning}}
        \vspace{-3mm}
        \includegraphics[width=\columnwidth]{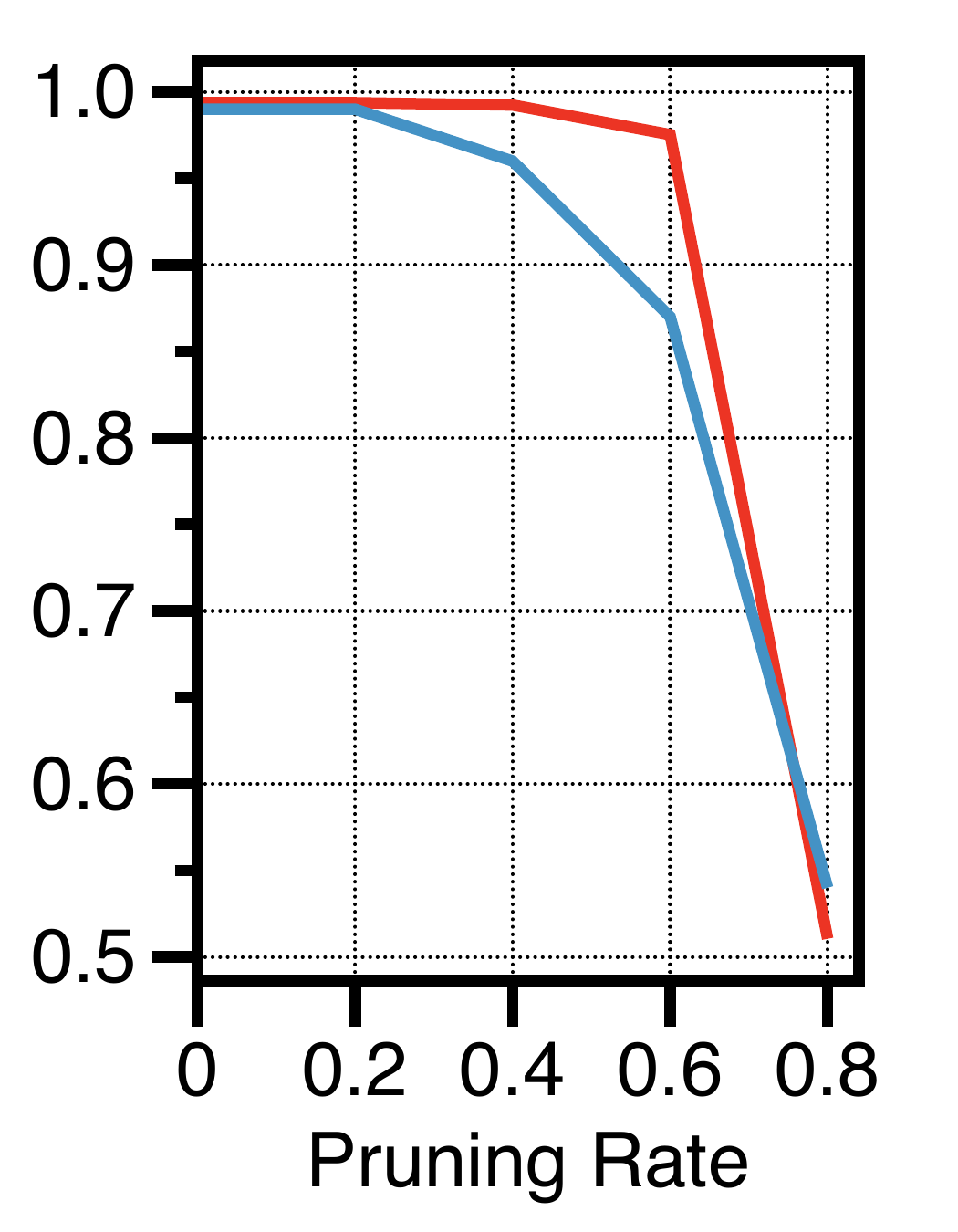} 
    \end{minipage}
    \begin{minipage}[c]{0.25\columnwidth} 
        \centering
        \caption*{\scriptsize{Fine-tuning}}
        \vspace{-3mm}
        \includegraphics[width=\columnwidth]{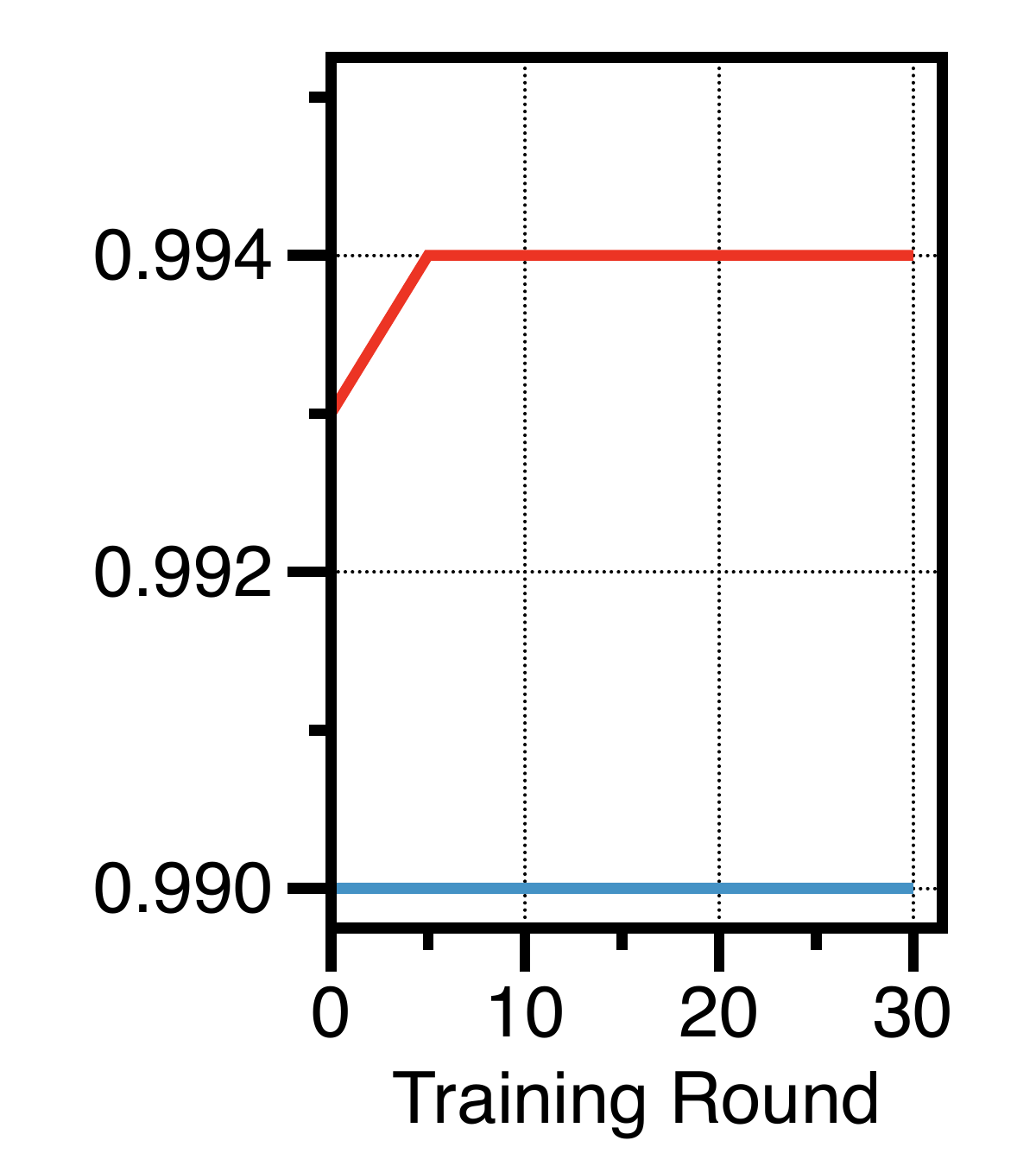} 
    \end{minipage}
    \begin{minipage}[c]{0.22\columnwidth} 
        \centering
        \caption*{\scriptsize{Quantization}}
        \vspace{-3mm}
        \includegraphics[width=\columnwidth]{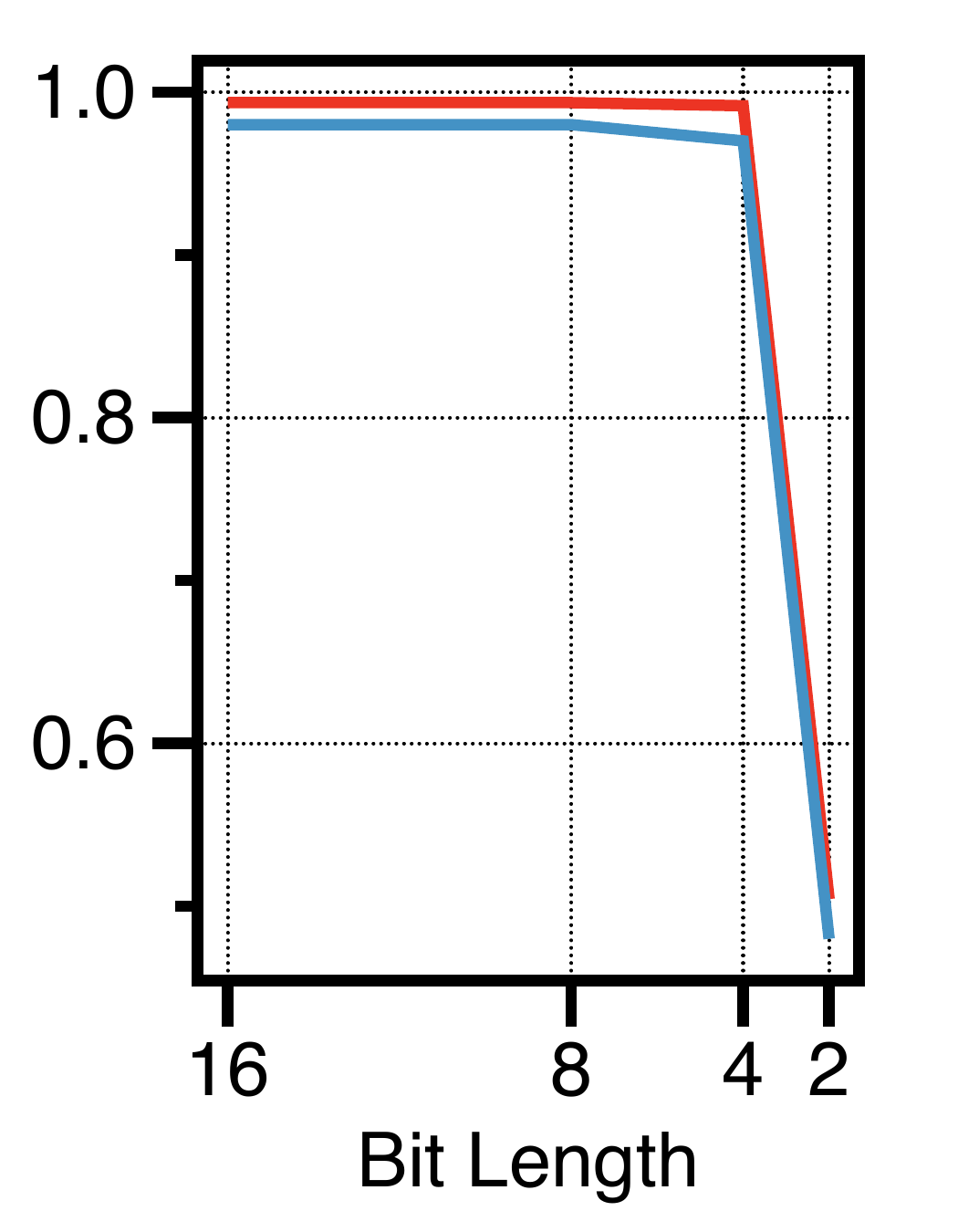} 
    \end{minipage}
    \begin{minipage}[c]{0.22\columnwidth} 
        \centering
        \caption*{\scriptsize{Overwriting}}
        \vspace{-3mm}
        \includegraphics[width=\columnwidth]{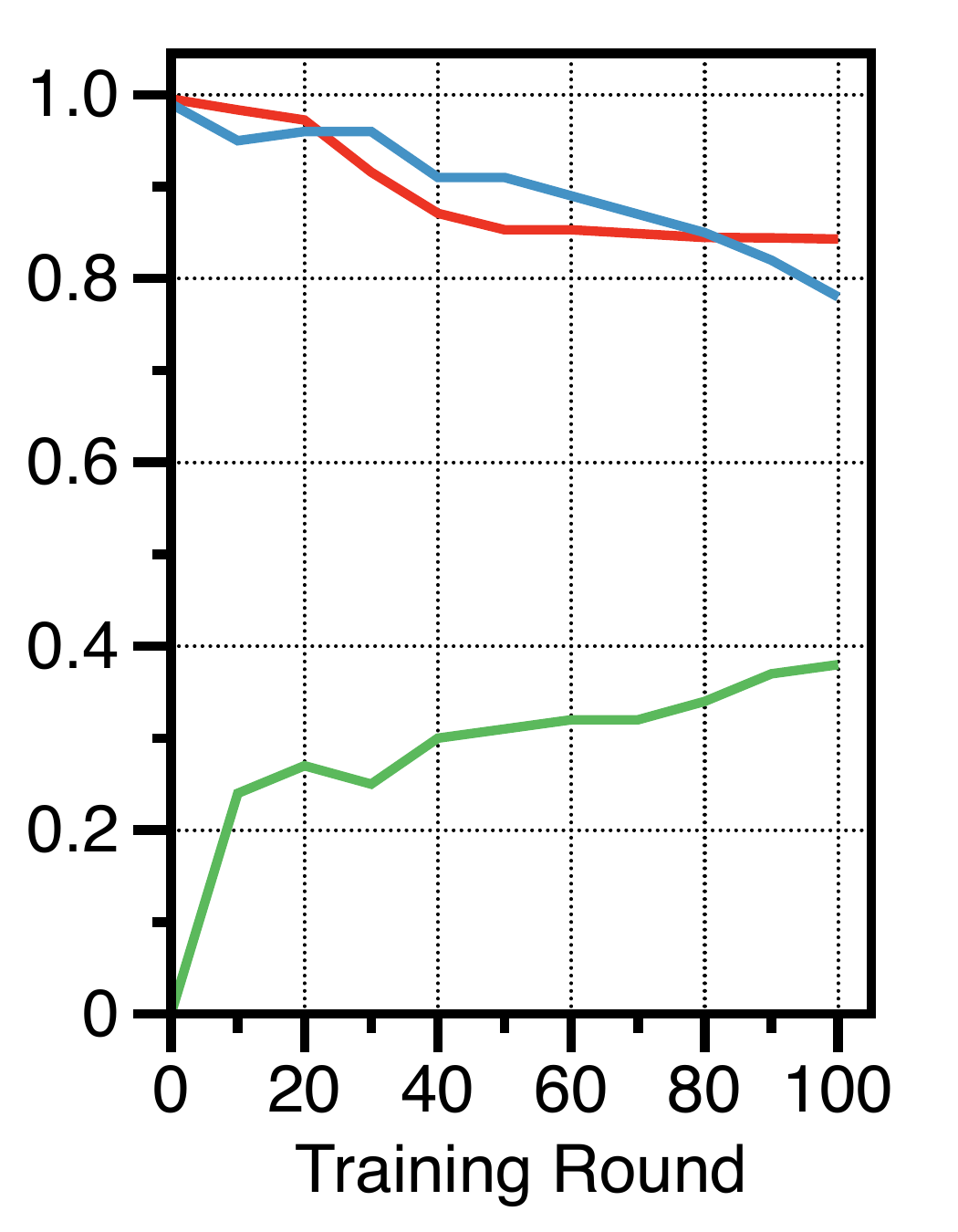} 
    \end{minipage}
    \begin{minipage}[c]{0.02\columnwidth}
     	\centering
     	\rotatebox{90}{\tiny{\textbf{Fashion-MNIST}}}
    \end{minipage}%
    \begin{minipage}[c]{0.22\columnwidth} 
        \centering
        \includegraphics[width=\columnwidth]{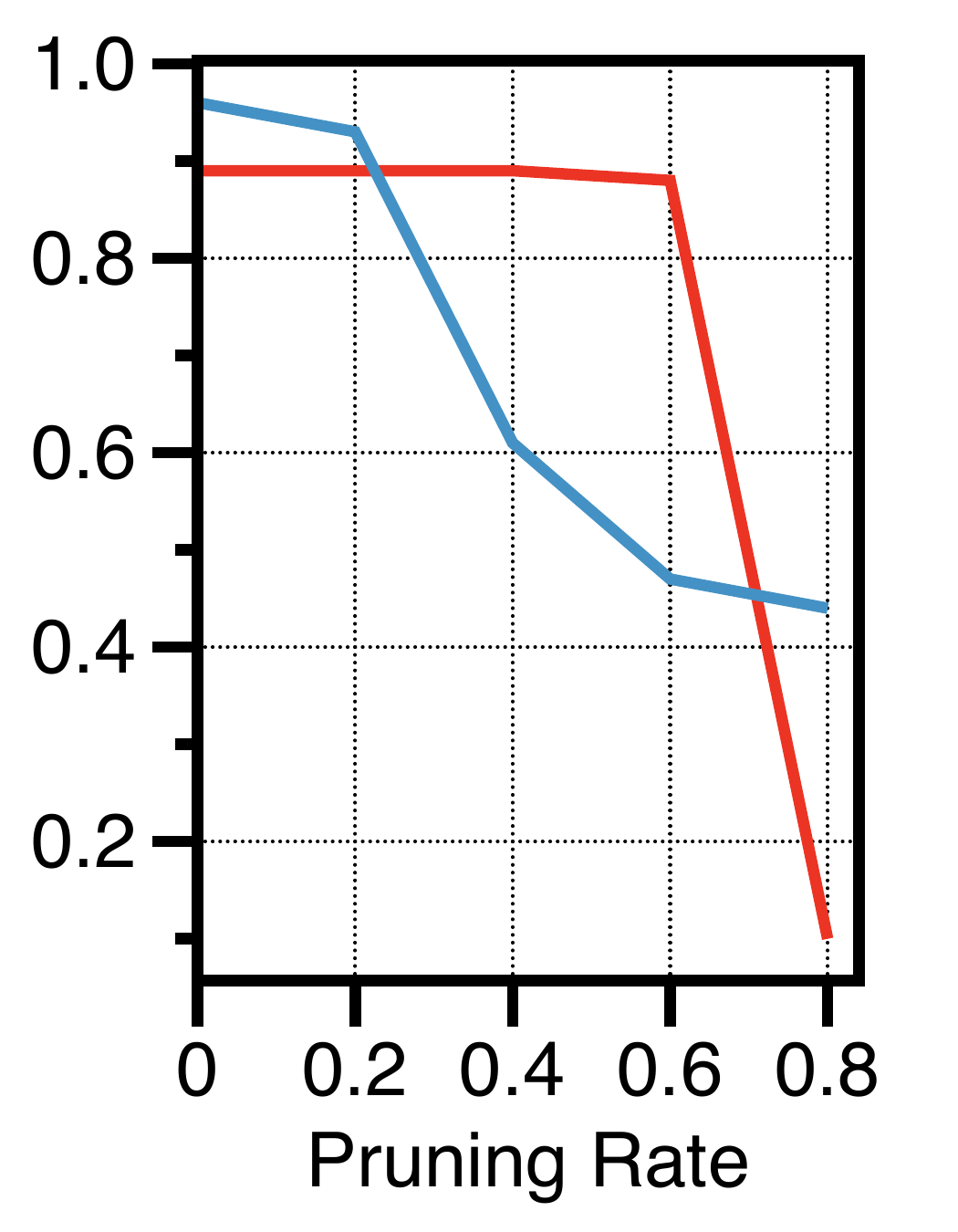} 
    \end{minipage}
    \begin{minipage}[c]{0.25\columnwidth} 
        \centering
        \includegraphics[width=\columnwidth]{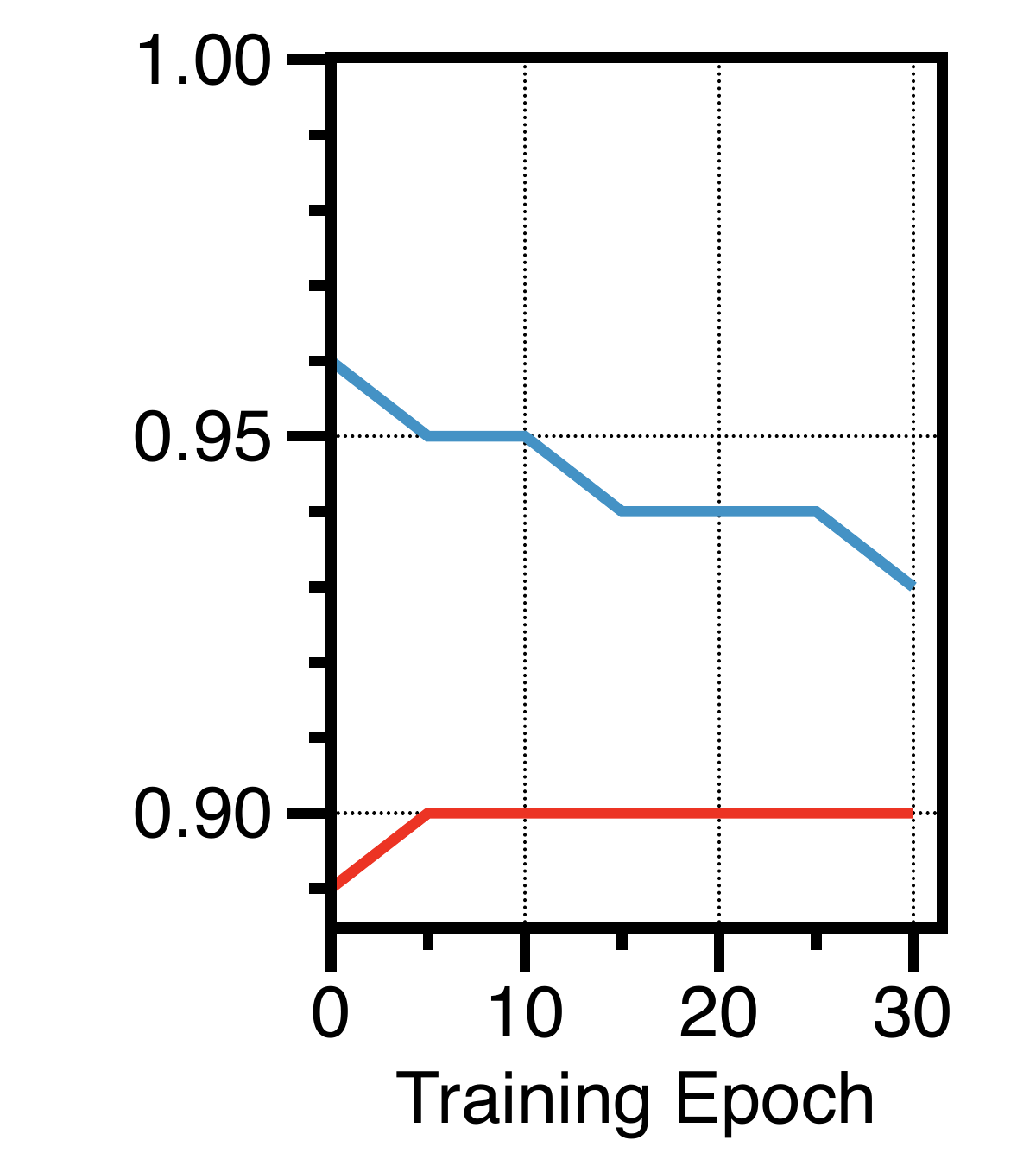} 
    \end{minipage}
    \begin{minipage}[c]{0.22\columnwidth} 
        \centering
        \includegraphics[width=\columnwidth]{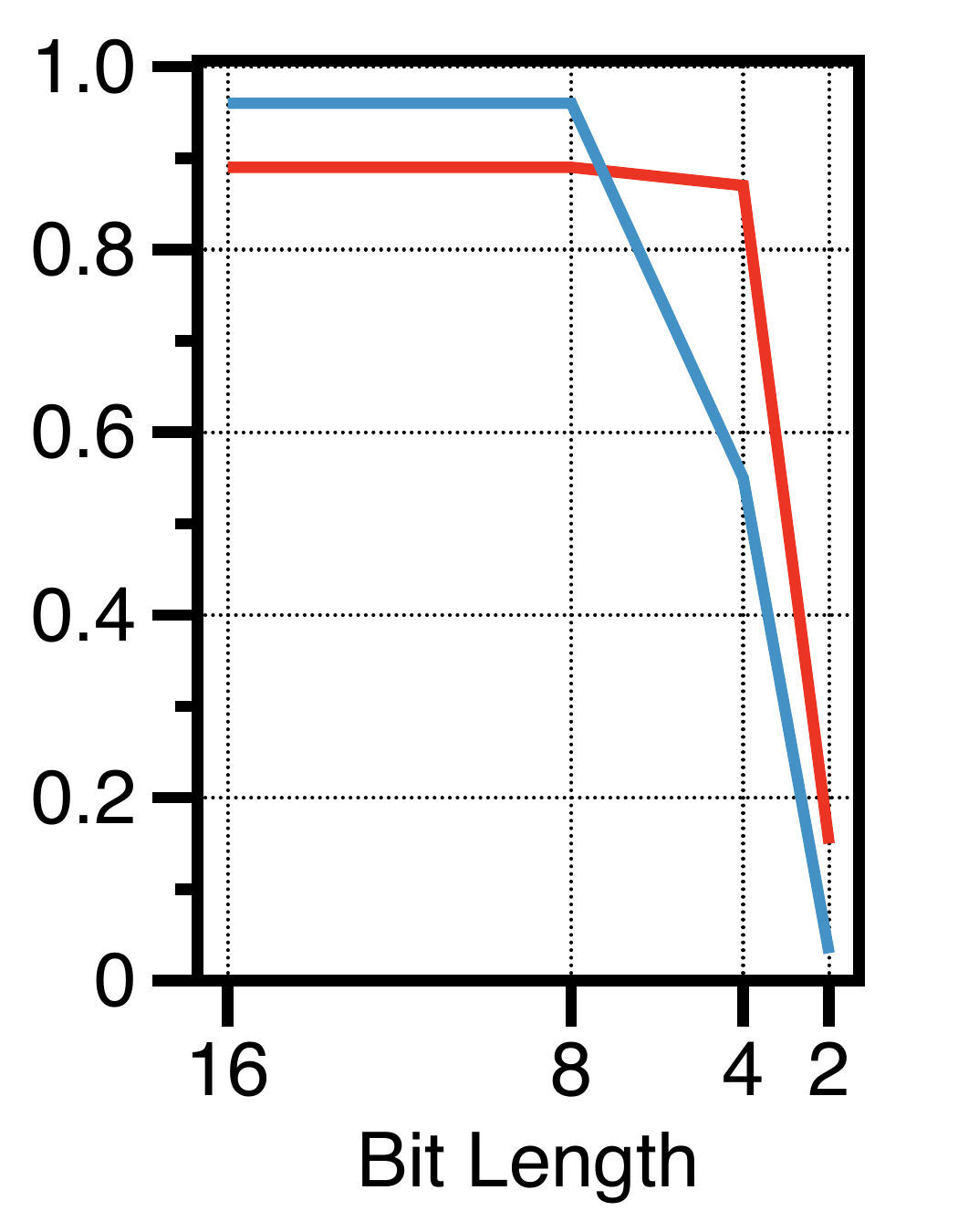} 
    \end{minipage}
    \begin{minipage}[c]{0.22\columnwidth} 
        \centering
        \includegraphics[width=\columnwidth]{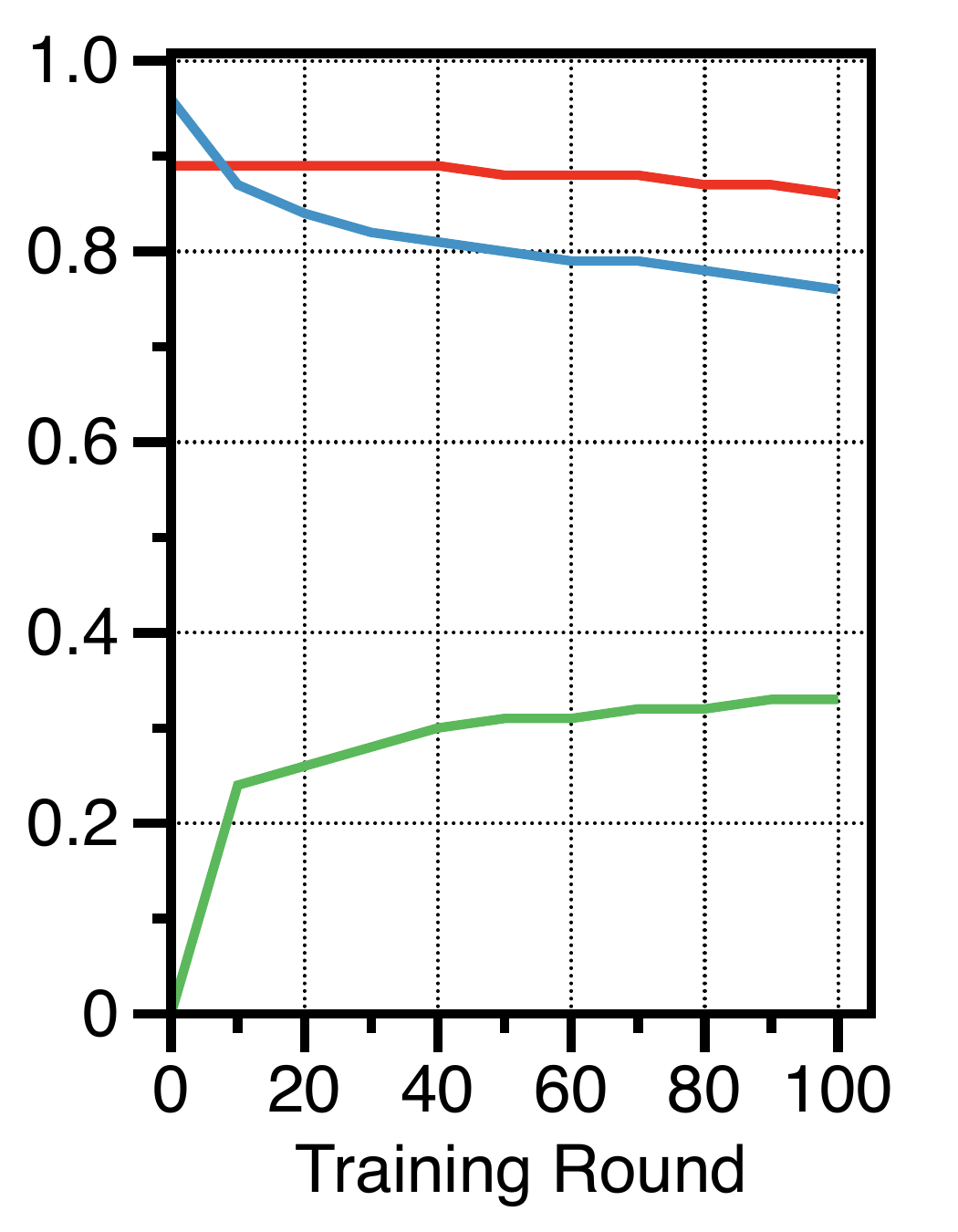} 
    \end{minipage}
    \begin{minipage}[c]{0.02\columnwidth}
     	\centering
     	\rotatebox{90}{\tiny{\textbf{CIFAR-10}}}
    \end{minipage}%
    \begin{minipage}[c]{0.22\columnwidth} 
        \centering
        \includegraphics[width=\columnwidth]{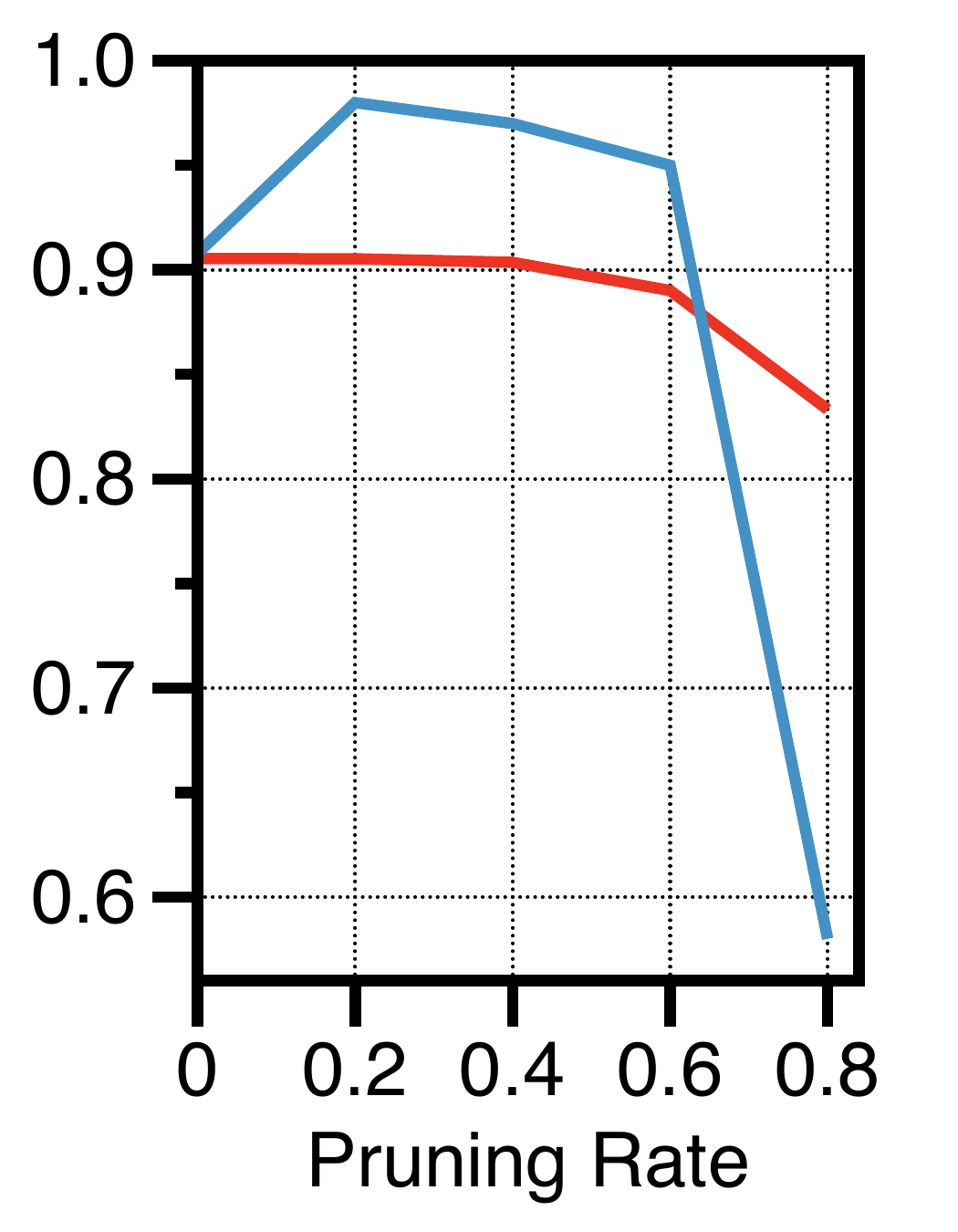} 
    \end{minipage}
    \begin{minipage}[c]{0.25\columnwidth} 
        \centering
        \includegraphics[width=\columnwidth]{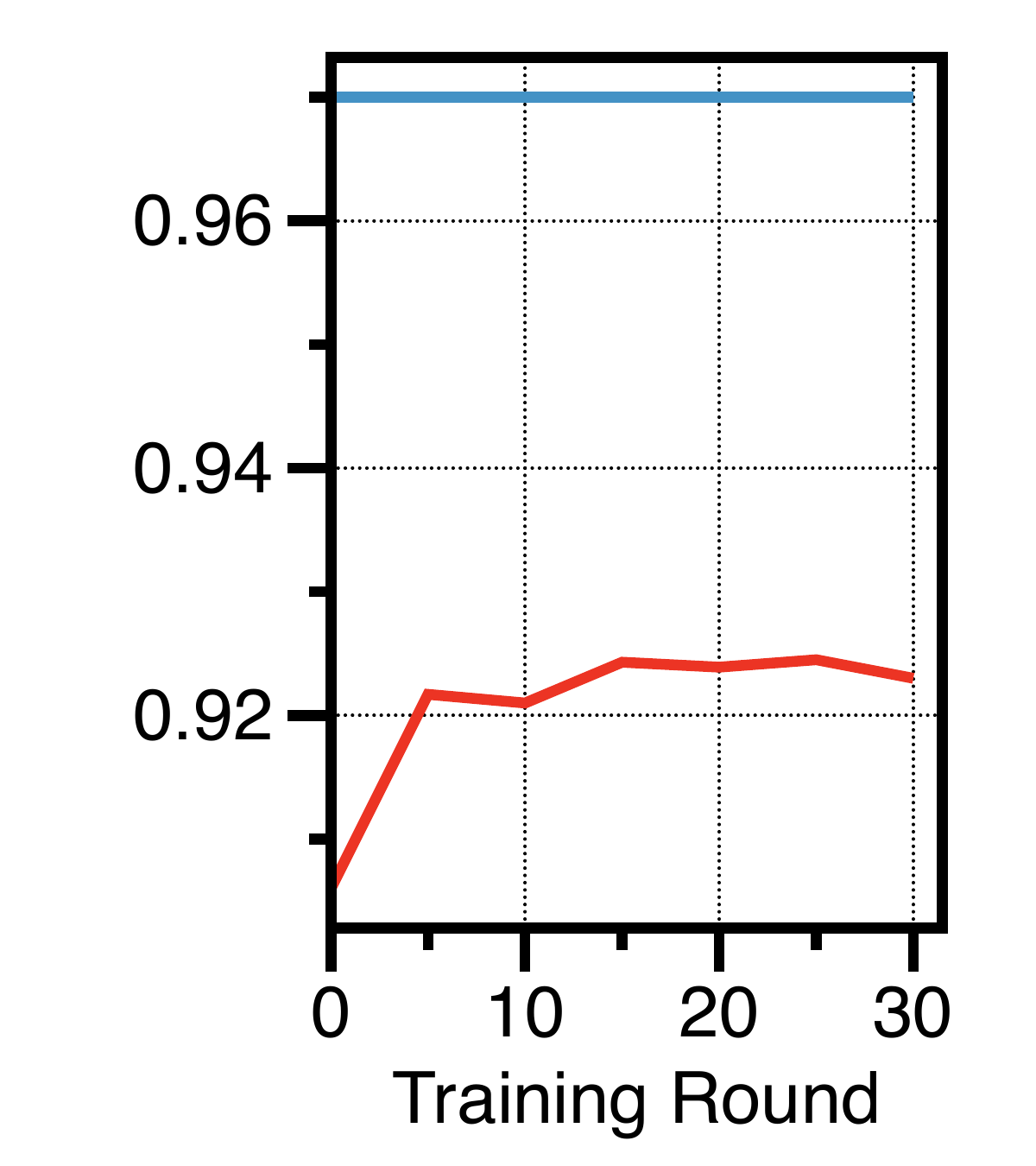} 
    \end{minipage}
    \begin{minipage}[c]{0.22\columnwidth} 
        \centering
        \includegraphics[width=\columnwidth]{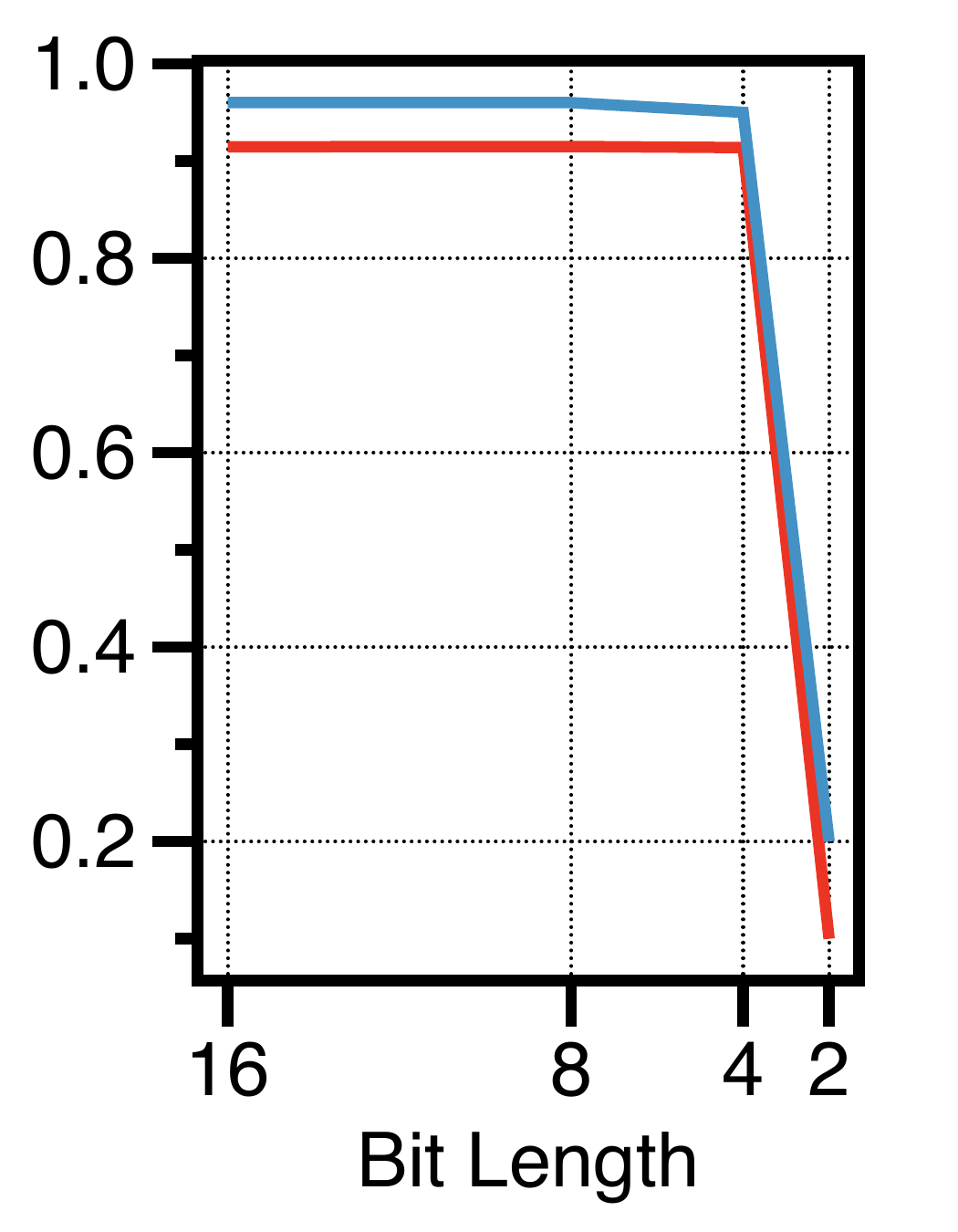} 
    \end{minipage}
    \begin{minipage}[c]{0.22\columnwidth} 
        \centering
        \includegraphics[width=\columnwidth]{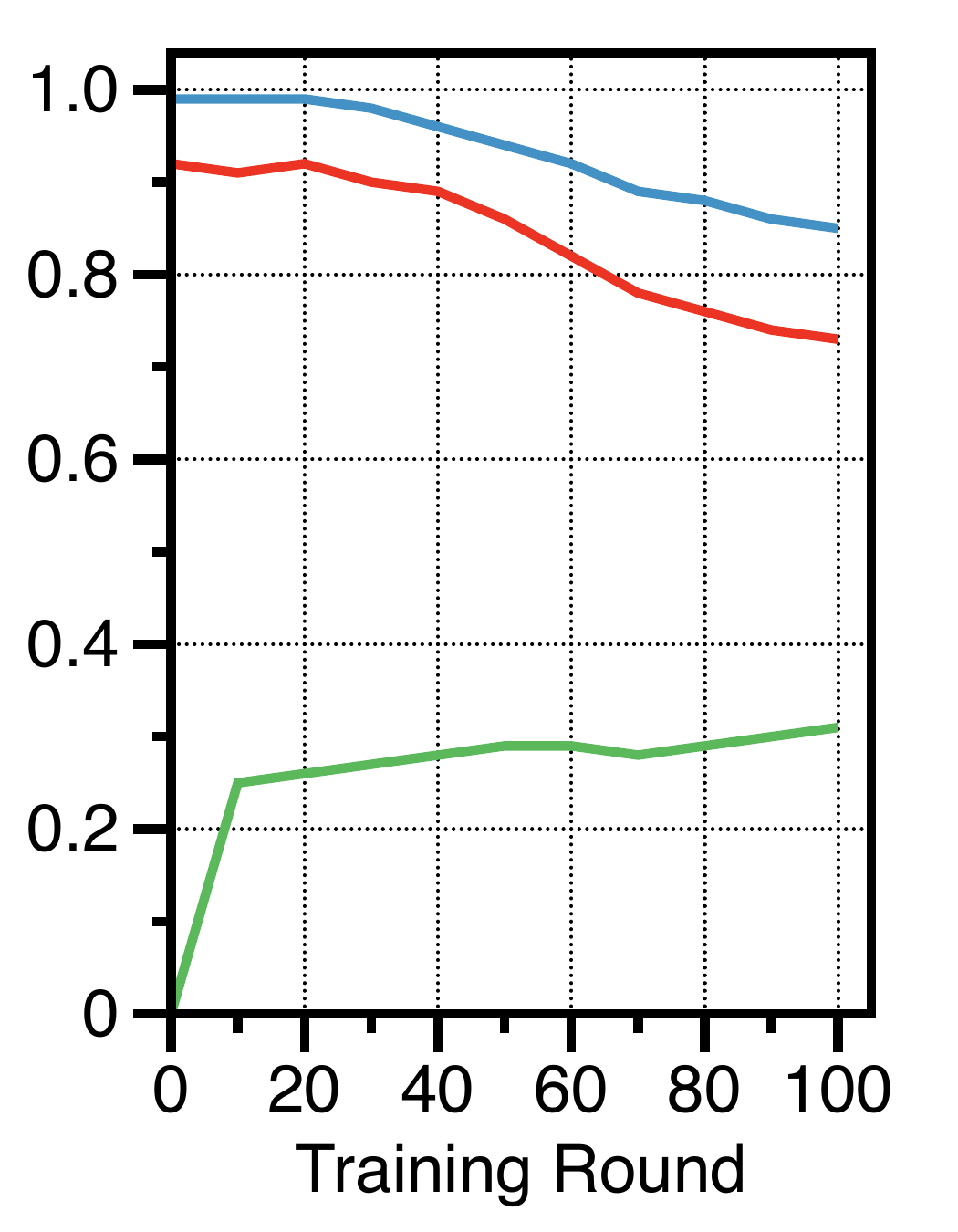} 
    \end{minipage}
    \begin{minipage}[c]{0.02\columnwidth}
     	\centering
     	\rotatebox{90}{\tiny{\textbf{CIFAR-100}}}
    \end{minipage}%
    \begin{minipage}[c]{0.22\columnwidth} 
        \centering
        \includegraphics[width=\columnwidth]{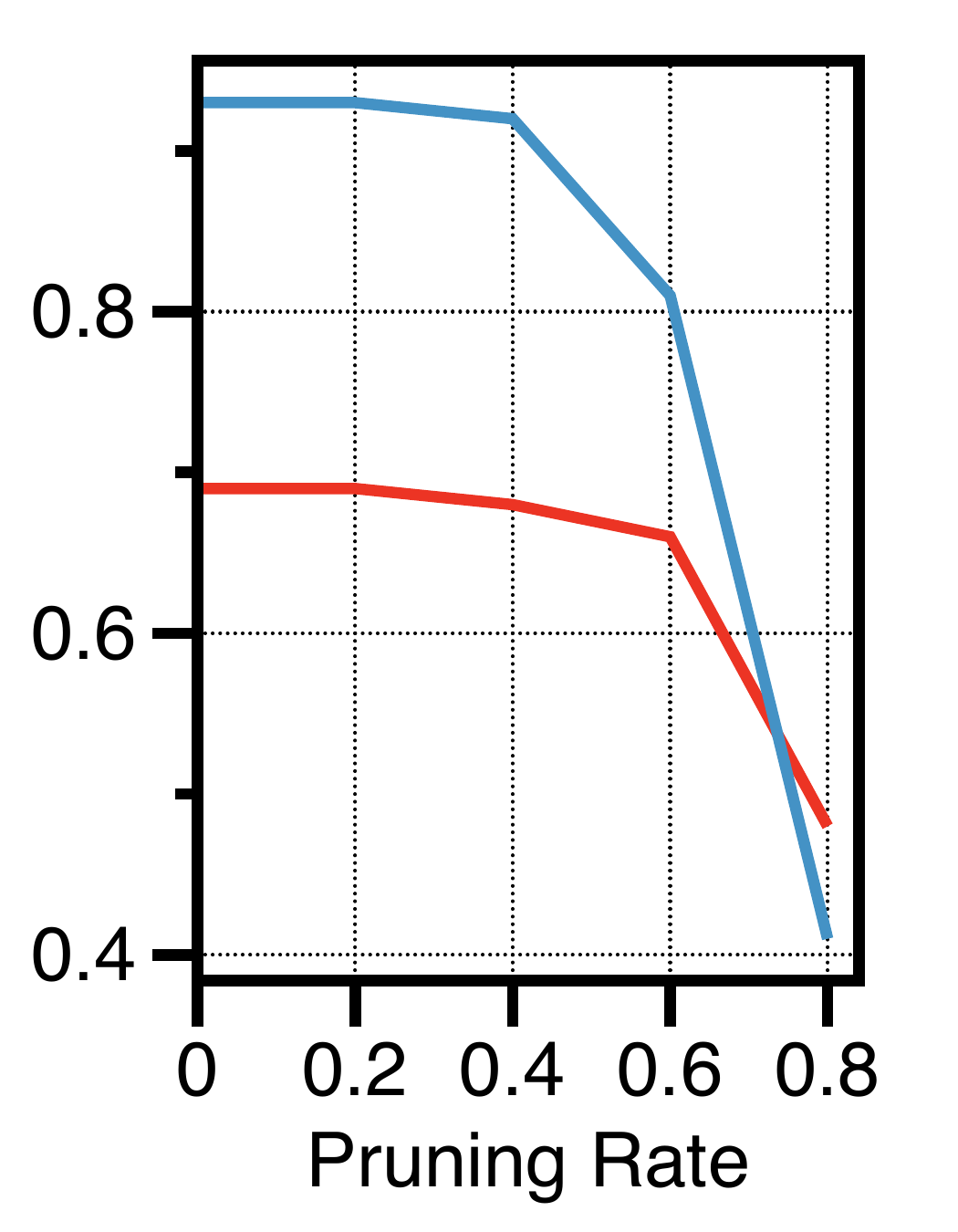} 
    \end{minipage}
    \begin{minipage}[c]{0.25\columnwidth} 
        \centering
        \includegraphics[width=\columnwidth]{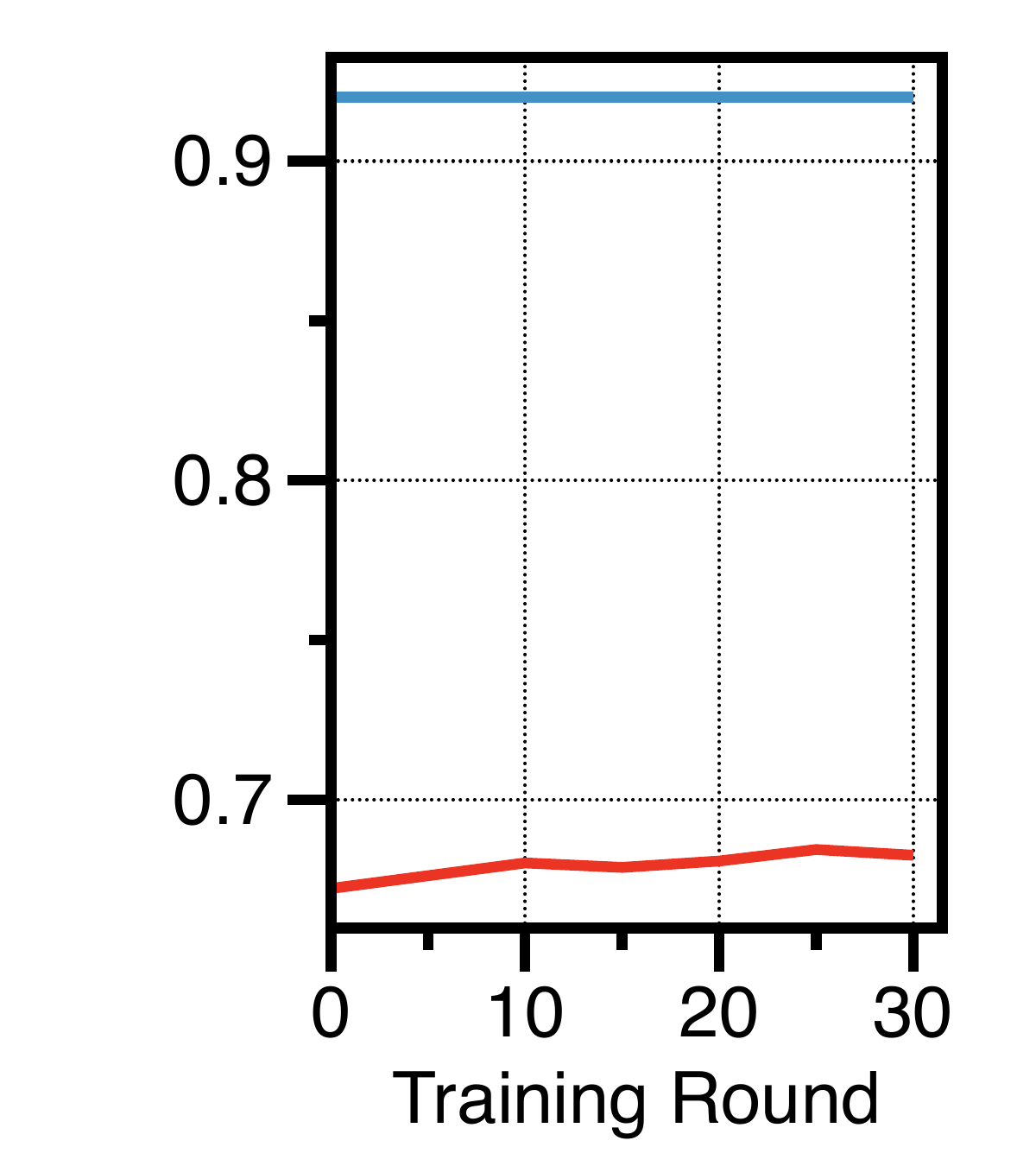} 
    \end{minipage}
    \begin{minipage}[c]{0.22\columnwidth} 
        \centering
        \includegraphics[width=\columnwidth]{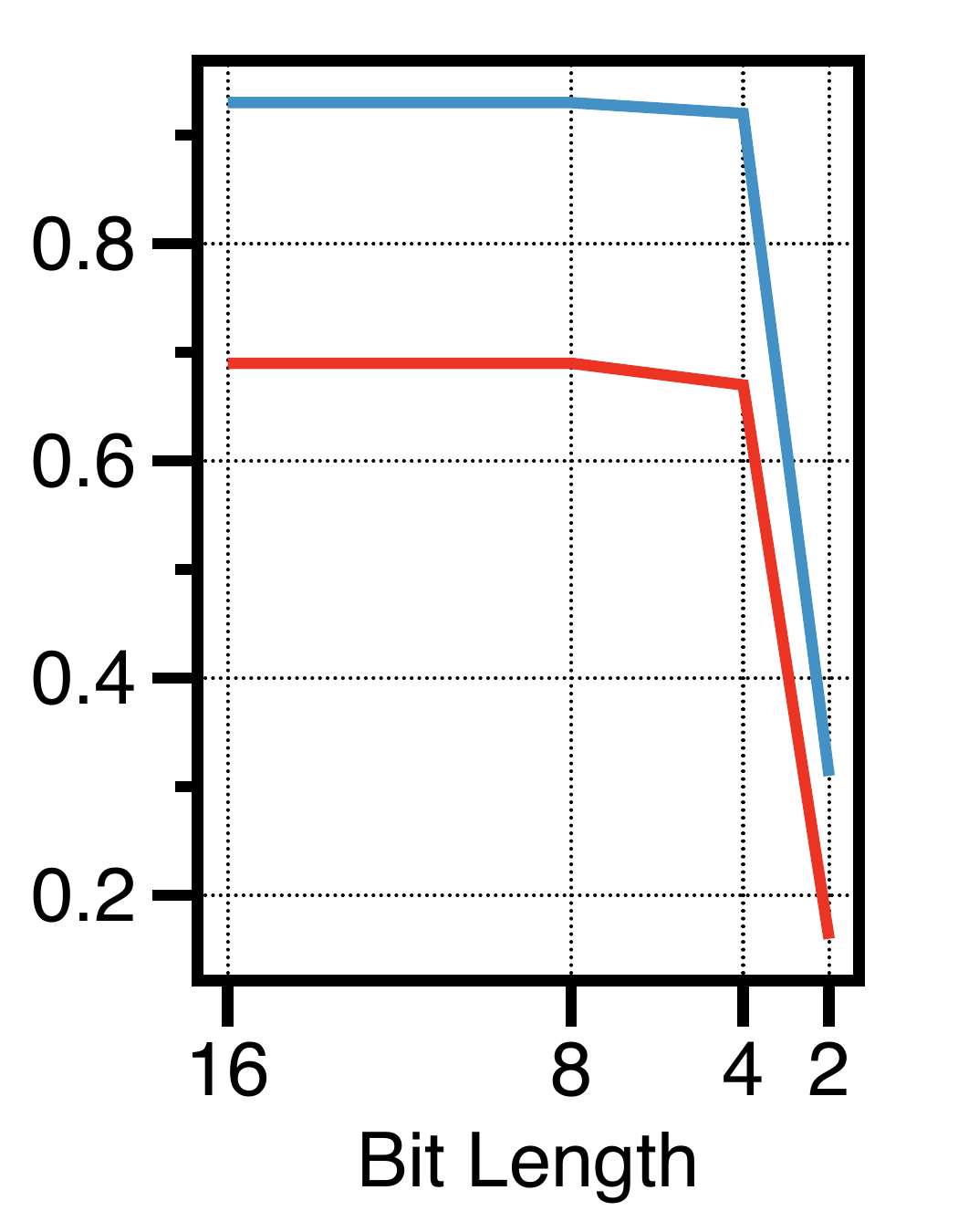} 
    \end{minipage}
    \begin{minipage}[c]{0.22\columnwidth} 
        \centering
        \includegraphics[width=\columnwidth]{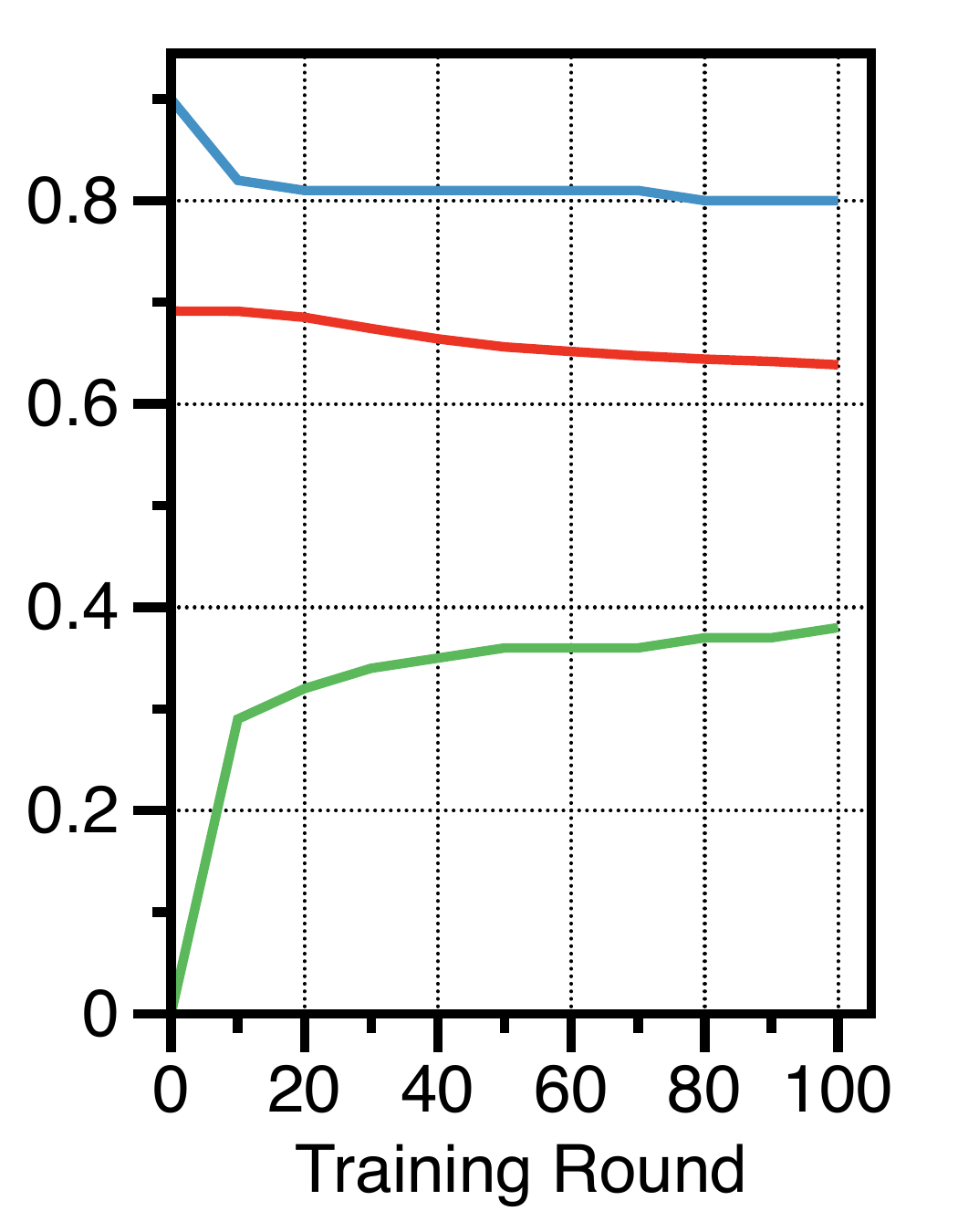} 
    \end{minipage}
    \begin{subfigure}{0.6\columnwidth}
        \centering
        \includegraphics[width=\textwidth]{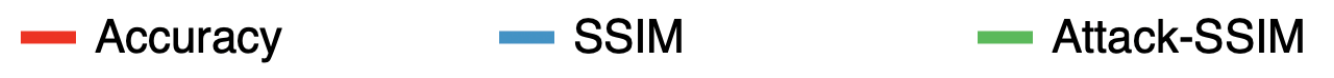}
    \end{subfigure}
\caption{Performance under modification attacks}
\label{fig:modification-attack}    
\end{figure}

The applicability of \texttt{FLClear} relies on ensuring that the aggregation process does not degrade either its watermark fidelity or main-task performance. \Cref{fig:Agg} illustrates the impact of different aggregation schemes on model accuracy and watermark fidelity across datasets under varying client numbers. Each row corresponds to results on MNIST, Fashion-MNIST, CIFAR-10, and CIFAR-100 datasets, respectively. All aggregation schemes integrated with \texttt{FLClear} (denoted “-WM”) exhibit convergence behavior nearly identical to their non-watermarked counterparts. This observation demonstrates that watermark embedding introduces negligible performance degradation, thereby preserving training fidelity. Similar trends hold for more complex datasets (CIFAR-10 and CIFAR-100), where watermark-integrated models achieve comparable convergence and final accuracy to baseline models.

Moreover, SSIM results in \Cref{fig:Agg} indicate that the watermark structure remains highly consistent throughout training. For simpler datasets such as MNIST, SSIM rapidly converges above 0.9, while for more complex CIFAR-based and Fashion-MNIST datasets, it stabilizes between 0.8 and 0.9 after approximately 100 rounds due to the increased difficulty of preserving structural details. Furthermore, watermark fidelity has a slight decrease as the number of clients increases from 10 to 20, likely due to enhanced aggregation diversity. These results confirm that \texttt{FLClear} maintains a favorable balance between model accuracy and watermark fidelity across diverse aggregation strategies.

To assess whether watermark images exhibit collisions, we select three watermark images (including both grayscale and color) from each dataset across all aggregation schemes for visual inspection. Because SSIM serves as the optimization objective for enhancing watermark fidelity during training, we consistently adopt SSIM as a primary metric to quantify structural preservation across experiments. In addition, we provide a comprehensive evaluation using multiple image-quality metrics, including MSE, PSNR, and LPIPS, to further validate the robustness and perceptual integrity of the reconstructed watermarks.
The results in \Cref{fig:w-ar} reveal that all aggregation strategies maintain high watermark fidelity. For instance, on MNIST, all methods achieve SSIM $\geq$ 0.973 and LPIPS $\leq$ 0.047, indicating minimal perceptual distortion. For more complex datasets such as CIFAR-100, FedPAQ and FEDADAM achieve low MSE (e.g., both 0.005 for color images on  CIFAR-100), and SCAFFOLD preserves clearer watermarks with PSNR of 22.41 dB. 

\subsection{Watermark Robustness and Security}
\label{sec:security}

\begin{figure}[t]
    \centering
	\begin{subfigure}[c]{0.02\columnwidth}
    \centering
    \vspace{8mm}
    \rotatebox{90}{\tiny{\textbf{MNIST}}}
    \end{subfigure}%
    \begin{minipage}[c]{0.155\columnwidth}
        \centering
        \caption*{\scriptsize{Original}}
        \vspace{2mm}
        \includegraphics[width=\columnwidth]{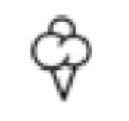}
    \end{minipage}%
    \begin{minipage}[c]{0.08\columnwidth}
        \centering
        \vspace{8mm}
        \tiny{SSIM $\uparrow$} \\ 
     	\tiny{MSE $\downarrow$} \\
     	\tiny{PSNR $\uparrow$} \\
     	\tiny{LPIPS $\downarrow$} \\ 
    \end{minipage}%
    \begin{minipage}[c]{0.15\columnwidth}
        \centering
        \caption*{\scriptsize{Pruning (40\%)}}
        \includegraphics[width=\columnwidth]{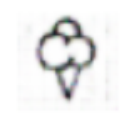}
    \end{minipage}%
    \begin{minipage}[c]{0.05\columnwidth}
        \centering
        \vspace{8mm}
        \tiny{0.954} \\ 
     	\tiny{0.007} \\
     	\tiny{20.06} \\
     	\tiny{0.022} \\ 
    \end{minipage}%
    \begin{minipage}[c]{0.15\columnwidth}
        \centering
        \caption*{\scriptsize{Fine-tuning (15 rounds)}}
        \includegraphics[width=\columnwidth]{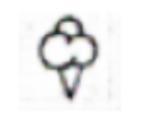}
    \end{minipage}%
    \begin{minipage}[c]{0.05\columnwidth}
        \centering
        \vspace{8mm}
        \tiny{0.998} \\ 
     	\tiny{0.001} \\
     	\tiny{34.35} \\
     	\tiny{0.001} \\ 
    \end{minipage}%
    \begin{minipage}[c]{0.15\columnwidth}
        \centering
        \caption*{\scriptsize{Quantization (8-bit)}}
        \includegraphics[width=\columnwidth]{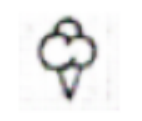}
    \end{minipage}%
    \begin{minipage}[c]{0.05\columnwidth}
        \centering
        \vspace{8mm}
        \tiny{0.988} \\ 
     	\tiny{0.001} \\
     	\tiny{28.47} \\
     	\tiny{0.013} \\ 
    \end{minipage}%
    
    \begin{subfigure}[c]{0.02\columnwidth}
        \centering
        
        \rotatebox{90}{\tiny{\textbf{Fashion-MNIST}}}
    \end{subfigure}%
    \begin{minipage}[c]{0.14\columnwidth}
        \centering
        \includegraphics[width=\columnwidth]{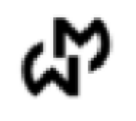}
    \end{minipage}%
    \begin{minipage}[c]{0.08\columnwidth}
        \centering
        \tiny{SSIM} \\ 
     	\tiny{MSE} \\
     	\tiny{PSNR} \\
     	\tiny{LPIPS} \\ 
    \end{minipage}%
    \begin{minipage}[c]{0.15\columnwidth}
        \centering
        \includegraphics[width=\columnwidth]{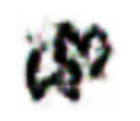}
    \end{minipage}%
    \begin{minipage}[c]{0.05\columnwidth}
        \centering
        \tiny{0.802} \\ 
     	\tiny{0.048} \\
     	\tiny{13.18} \\
     	\tiny{0.334} \\ 
    \end{minipage}%
    \begin{minipage}[c]{0.15\columnwidth}
        \centering
        \includegraphics[width=\columnwidth]{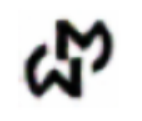}
    \end{minipage}%
    \begin{minipage}[c]{0.05\columnwidth}
        \centering
        \tiny{0.973} \\ 
     	\tiny{0.006} \\
     	\tiny{21.77} \\
     	\tiny{0.028} \\ 
    \end{minipage}%
    \begin{minipage}[c]{0.15\columnwidth}
        \centering
        \includegraphics[width=\columnwidth]{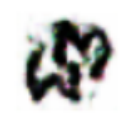}
    \end{minipage}
    \begin{minipage}[c]{0.05\columnwidth}
        \centering
        \tiny{0.795} \\ 
     	\tiny{0.048} \\
     	\tiny{13.15} \\
     	\tiny{0.327} \\ 
    \end{minipage}%
    
    \begin{subfigure}[c]{0.02\columnwidth}
        \centering
        \rotatebox{90}{\tiny{\textbf{CIFAR-10}}}
    \end{subfigure}%
    \begin{minipage}[c]{0.141\columnwidth}
        \centering
        \includegraphics[width=\columnwidth]{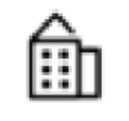}
    \end{minipage}%
    \begin{minipage}[c]{0.08\columnwidth}
        \centering
        \tiny{SSIM} \\ 
     	\tiny{MSE} \\
     	\tiny{PSNR} \\
     	\tiny{LPIPS} \\ 
    \end{minipage}%
    \begin{minipage}[c]{0.153\columnwidth}
        \centering
        \includegraphics[width=\columnwidth]{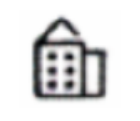}
    \end{minipage}%
    \begin{minipage}[c]{0.05\columnwidth}
        \centering
        \tiny{0.989} \\ 
     	\tiny{0.002} \\
     	\tiny{25.91} \\
     	\tiny{0.015} \\ 
    \end{minipage}%
    \begin{minipage}[c]{0.153\columnwidth}
        \centering
        \includegraphics[width=\columnwidth]{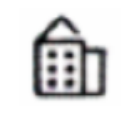}
    \end{minipage}%
    \begin{minipage}[c]{0.05\columnwidth}
        \centering
        \tiny{0.978} \\ 
     	\tiny{0.005} \\
     	\tiny{22.27} \\
     	\tiny{0.022} \\ 
    \end{minipage}%
    \begin{minipage}[c]{0.153\columnwidth}
        \centering
        \includegraphics[width=\columnwidth]{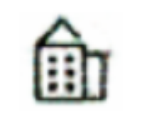}
    \end{minipage}%
    \begin{minipage}[c]{0.05\columnwidth}
        \centering
        \tiny{0.966} \\ 
     	\tiny{0.007} \\
     	\tiny{21.53} \\
     	\tiny{0.021} \\ 
    \end{minipage}%
    
    \begin{subfigure}[c]{0.02\columnwidth}
        \centering
        \rotatebox{90}{\tiny{\textbf{CIFAR-100}}}
    \end{subfigure}%
    \begin{minipage}[c]{0.14\columnwidth}
        \centering
        \includegraphics[width=\columnwidth]{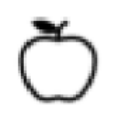}
    \end{minipage}%
    \begin{minipage}[c]{0.08\columnwidth}
        \centering
        \tiny{SSIM} \\ 
     	\tiny{MSE} \\
     	\tiny{PSNR} \\
     	\tiny{LPIPS} \\ 
    \end{minipage}%
    \begin{minipage}[c]{0.15\columnwidth}
        \centering
        \includegraphics[width=\columnwidth]{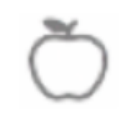}
    \end{minipage}%
    \begin{minipage}[c]{0.05\columnwidth}
        \centering
        \tiny{0.910} \\ 
     	\tiny{0.017} \\
     	\tiny{17.66} \\
     	\tiny{0.062} \\ 
    \end{minipage}%
    \begin{minipage}[c]{0.15\columnwidth}
        \centering
        \includegraphics[width=\columnwidth]{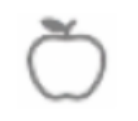}
    \end{minipage}%
    \begin{minipage}[c]{0.05\columnwidth}
        \centering
        \tiny{0.921} \\ 
     	\tiny{0.016} \\
     	\tiny{17.78} \\
     	\tiny{0.060} \\ 
    \end{minipage}%
    \begin{minipage}[c]{0.15\columnwidth}
        \centering
        \includegraphics[width=\columnwidth]{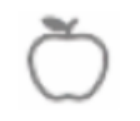}
    \end{minipage}
    \begin{minipage}[c]{0.05\columnwidth}
        \centering
        \tiny{0.918} \\ 
     	\tiny{0.015} \\
     	\tiny{17.99} \\
     	\tiny{0.066} \\ 
    \end{minipage}%
\caption{Watermark images under modification attacks}
\label{fig:modification}
\end{figure}

\begin{figure}[t]
	\centering
    \begin{minipage}[c]{0.02\columnwidth}
     	\centering
     	\rotatebox{90}{\tiny{\textbf{MNIST}}}
    \end{minipage}%
    \begin{minipage}[c]{0.2\columnwidth} 
        \centering
        \caption*{\tiny{Adversary's Original and Overwriting Watermarks}}
        \vspace{-1mm}
        \includegraphics[width=\columnwidth]{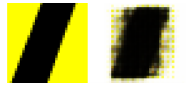}
    \end{minipage}
    \begin{minipage}[c]{0.7\columnwidth} 
        \centering
        \caption*{\scriptsize{Watermarks of Different Clients}}
        \vspace{-1mm}
        \includegraphics[width=\columnwidth]{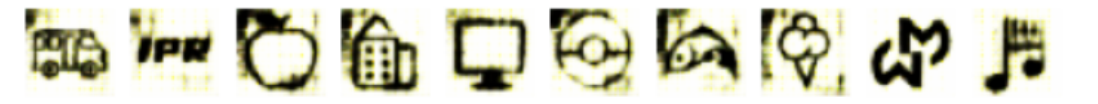}
    \end{minipage}
    \begin{minipage}[c]{0.02\columnwidth}
     	\centering
     	\rotatebox{90}{\tiny{\textbf{Fashion-MNIST}}}
    \end{minipage}%
    \begin{minipage}[c]{0.2\columnwidth} 
        \centering
        \vspace{1.2mm}
        \includegraphics[width=\columnwidth]{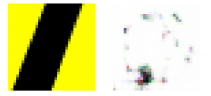}
    \end{minipage}
    \begin{minipage}[c]{0.7\columnwidth} 
        \centering
        \vspace{1.2mm}
        \includegraphics[width=\columnwidth]{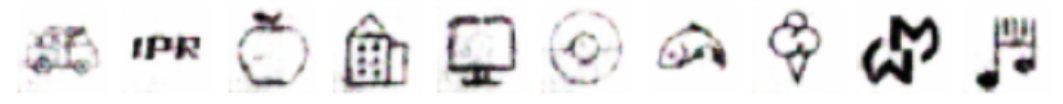}
    
    \end{minipage}
	\begin{minipage}[c]{0.02\columnwidth}
     	\centering
     	\rotatebox{90}{\tiny{\textbf{CIFAR-10}}}
    \end{minipage}%
    \begin{minipage}[c]{0.2\columnwidth} 
        \centering
        \vspace{1.2mm}
        \includegraphics[width=\columnwidth]{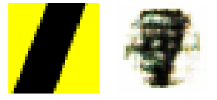}
    \end{minipage}
    \begin{minipage}[c]{0.7\columnwidth} 
        \centering
        \vspace{1.2mm}
        \includegraphics[width=\columnwidth]{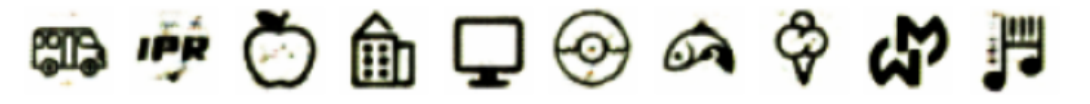}
        
    \end{minipage}
	\begin{minipage}[c]{0.02\columnwidth}
     	\centering
     	\rotatebox{90}{\tiny{\textbf{CIFAR-100}}}
    \end{minipage}%
    \begin{minipage}[c]{0.2\columnwidth} 
        \centering
        \vspace{1.2mm}
        \includegraphics[width=\columnwidth]{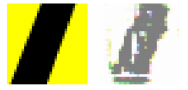}
    \end{minipage}
    \begin{minipage}[c]{0.7\columnwidth} 
        \centering
        \vspace{1.2mm}
        \includegraphics[width=\columnwidth]{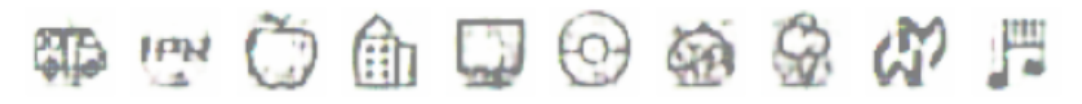}
    \end{minipage}
\caption{Watermark images under overwriting attacks}
\label{fig:overwrite}
\end{figure} 

\begin{table*}[htbp]
\centering
\caption{ASR results under forgery attacks with parameter combinations. The green regions indicate the best defense performance.}
\label{tab:asr}
\resizebox{\textwidth}{!}{
\begin{tabular}{cccccccccccccccc}
\toprule
\multirow{2}{*}{\shortstack{\textbf{Attack}\\ \\\textbf{Type}}} &
\multirow{2}{*}{\textbf{Threshold $\tau$}} &
\multicolumn{4}{c}{$\boldsymbol{m=0.1, y=0.1, num=250}$} & 
\multicolumn{4}{c}{$\boldsymbol{m=0.1, y=0.9, num=500}$} & 
\multicolumn{4}{c}{$\boldsymbol{m=0.5, y=0.5, num=750}$} \\

\cmidrule(lr){3-6} \cmidrule(lr){7-10} \cmidrule(lr){11-14} & 
& \textbf{MNIST} & \textbf{Fashion-MNIST} & \textbf{CIFAR-10} & \textbf{CIFAR-100} 
& \textbf{MNIST} & \textbf{Fashion-MNIST} & \textbf{CIFAR-10} & \textbf{CIFAR-100}
& \textbf{MNIST} & \textbf{Fashion-MNIST} & \textbf{CIFAR-10} & \textbf{CIFAR-100} \\
\midrule

\multirow{5}{*}{\textbf{Targeted (\%)}} 
& 0.1 & 33.70 & 100.00 & 17.70 & 78.30 & 27.30 & 100.00 & 81.00 & 89.00 & 100.00 & 100.00 & 100.00 & 100.00 \\
& 0.3 & 25.00 & 9.30 & 0.70 & 0.30 & 23.00 & \cellcolor{green!20}0.00 & \cellcolor{green!20}0.00 & \cellcolor{green!20}0.00 & 78.70 & 81.30 & 99.00 & 3.30 \\
& 0.5 & 22.00 & \cellcolor{green!20}0.00 & \cellcolor{green!20}0.00 & \cellcolor{green!20}0.00 & 18.30 & \cellcolor{green!20}0.00 & \cellcolor{green!20}0.00 & \cellcolor{green!20}0.00 & 47.70 & \cellcolor{green!20}\cellcolor{green!20}0.00 & 3.70 & \cellcolor{green!20}0.00 \\
& 0.7 & 16.30 & \cellcolor{green!20}0.00 & \cellcolor{green!20}0.00 & \cellcolor{green!20}0.00 & 17.00 & \cellcolor{green!20}0.00 & \cellcolor{green!20}0.00 & \cellcolor{green!20}0.00 & 33.30 & \cellcolor{green!20}0.00 & \cellcolor{green!20}0.00 & \cellcolor{green!20}0.00 \\
& 0.9 & \cellcolor{green!20}0.00 & \cellcolor{green!20}0.00 & \cellcolor{green!20}0.00 & \cellcolor{green!20}0.00 & \cellcolor{green!20}0.00 & \cellcolor{green!20}0.00 & \cellcolor{green!20}0.00 & \cellcolor{green!20}0.00 & \cellcolor{green!20}0.00 & \cellcolor{green!20}0.00 & \cellcolor{green!20}0.00 & \cellcolor{green!20}0.00 &  \\
\midrule

\multirow{5}{*}{\textbf{Untargeted (\%)}} 
& 0.1 & 5.30 & 99.30 & 6.70 & 100.00 & 14.00 & 6.70 & 46.70 & 100.00 & 100.00 & 100.00 & 100.00 & 100.00 \\
& 0.3 & \cellcolor{green!20}0.00 & \cellcolor{green!20}0.00 & \cellcolor{green!20}0.00 & \cellcolor{green!20}0.00 & \cellcolor{green!20}0.00 & \cellcolor{green!20}0.00 & \cellcolor{green!20}0.00 & \cellcolor{green!20}0.00 & \cellcolor{green!20}0.00 & \cellcolor{green!20}0.00 & \cellcolor{green!20}0.00 & \cellcolor{green!20}0.00 \\
& 0.5 & \cellcolor{green!20}0.00 & \cellcolor{green!20}0.00 & \cellcolor{green!20}0.00 & \cellcolor{green!20}0.00 & \cellcolor{green!20}0.00 & \cellcolor{green!20}0.00 & \cellcolor{green!20}0.00 & \cellcolor{green!20}0.00 & \cellcolor{green!20}0.00 & \cellcolor{green!20}0.00 & \cellcolor{green!20}0.00 & \cellcolor{green!20}0.00  \\
& 0.7 & \cellcolor{green!20}0.00 & \cellcolor{green!20}0.00 & \cellcolor{green!20}0.00 & \cellcolor{green!20}0.00 & \cellcolor{green!20}0.00 & \cellcolor{green!20}0.00 & \cellcolor{green!20}0.00 & \cellcolor{green!20}0.00 & \cellcolor{green!20}0.00 & \cellcolor{green!20}0.00 & \cellcolor{green!20}0.00 & \cellcolor{green!20}0.00  \\
& 0.9 & \cellcolor{green!20}0.00 & \cellcolor{green!20}0.00 & \cellcolor{green!20}0.00 & \cellcolor{green!20}0.00 & \cellcolor{green!20}0.00 & \cellcolor{green!20}0.00 & \cellcolor{green!20}0.00 & \cellcolor{green!20}0.00 & \cellcolor{green!20}0.00 & \cellcolor{green!20}0.00 & \cellcolor{green!20}0.00 & \cellcolor{green!20}0.00  \\
\bottomrule
\end{tabular}}
\end{table*}

\begin{figure}[t]
    \centering
    \begin{subfigure}{0.48\columnwidth}
        \centering
        \includegraphics[width=\textwidth]{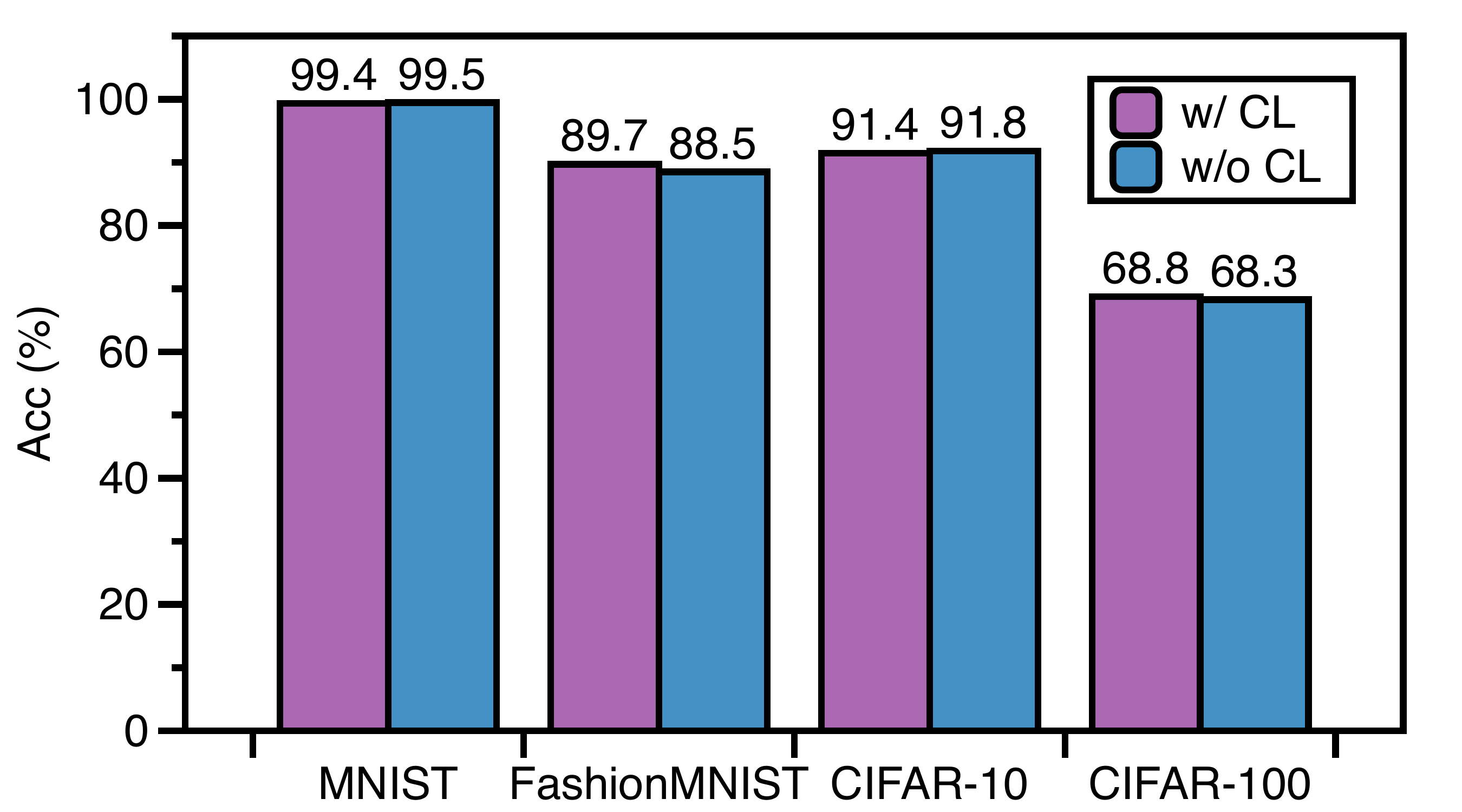}
        \caption{Accuracy}
        \label{fig:acc-abl}
    \end{subfigure}
    \begin{subfigure}{0.48\columnwidth}
        \centering
        \includegraphics[width=\textwidth]{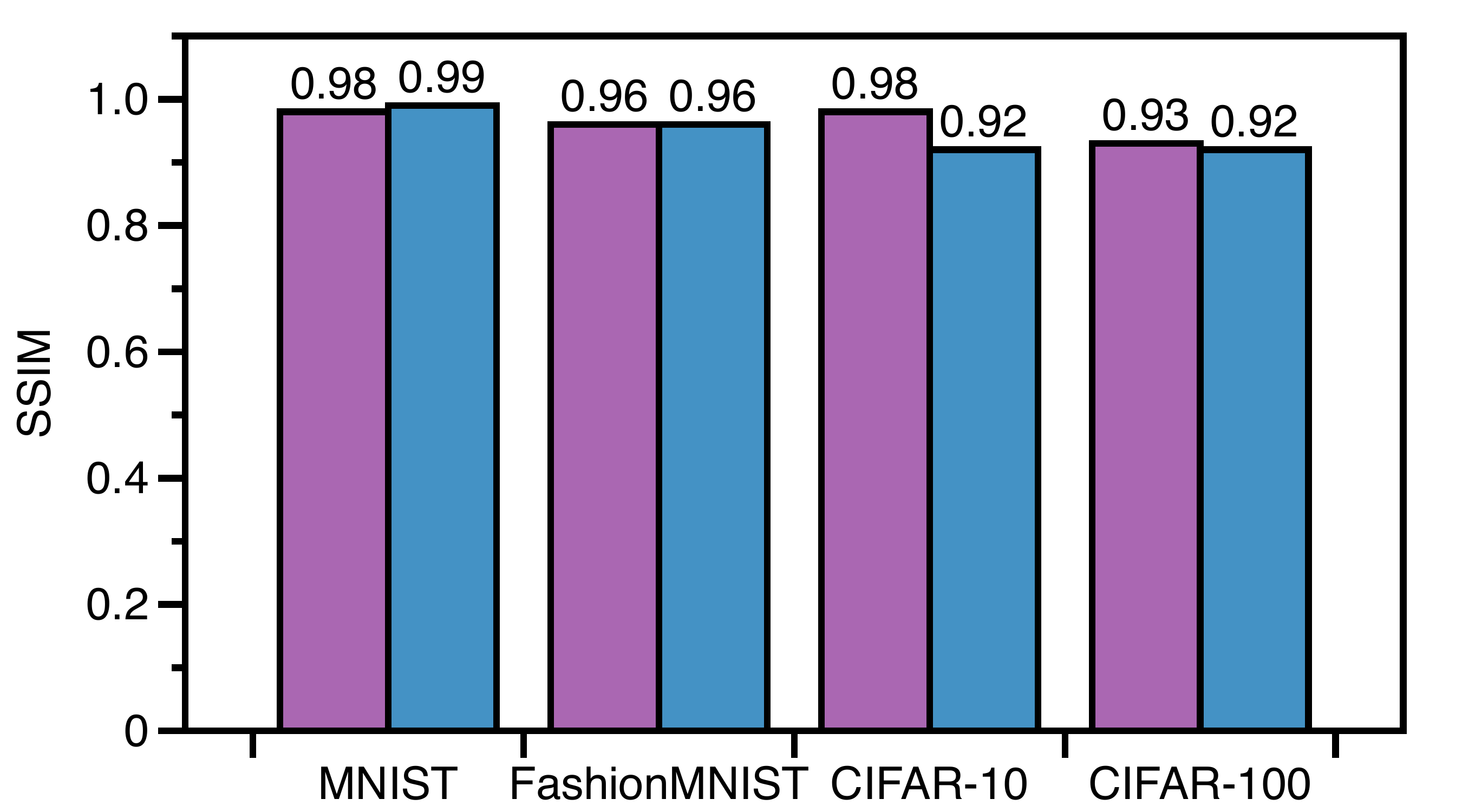}
        \caption{SSIM}
        \label{fig:ssim-abl}
    \end{subfigure}
    \caption{Impact of contrastive learning}
    \label{fig:acc-ssim-abl}
\end{figure}
 
\noindent \textbf{Watermark Robustness.}
In \Cref{fig:modification-attack}, we evaluate the watermark robustness and model performance under various model modification attacks, including pruning, fine-tuning, quantization, and overwriting. Results demonstrate that \texttt{FLClear} preserves both task fidelity and perceptual integrity under diverse model modification attacks.
For pruning attacks, model accuracy and SSIM remain nearly unaffected below a 0.6 pruning rate but drop sharply at 0.8. This indicates that the embedded watermark is resilient to moderate parameter removal but sensitive to extreme sparsification.
During fine-tuning, accuracy and SSIM fluctuate only slightly across 30 training rounds, demonstrating that the watermark remains stable during re-optimization without hindering convergence.
Under quantization attacks, the watermark remains robust to low-precision deployment. Both metrics stay high once the bit width exceeds 4.  
For overwriting attacks, model accuracy and watermark SSIM show a slight decrease, whereas the attacker’s watermark SSIM rises. However, the attacker’s SSIM remains consistently below that of the genuine watermark, which indicates that these attacks can be mitigated by choosing an appropriate verification threshold.

We further present a visual and quantitative analysis of the embedded watermarks under modification attacks in \Cref{fig:modification}. Despite the structural perturbations introduced by these attacks, the reconstructed watermarks exhibit strong fidelity, as evidenced by consistently high SSIM values (e.g., 0.954–0.998 on MNIST and 0.910–0.921 on CIFAR-100) and low MSE ($\leq$ 0.017 across all datasets). Among all modification types, fine-tuning results in the least distortion, achieving both high SSIM (up to 0.998) and high PSNR (up to 34.35 dB), indicating minimal deviation from the originally designated watermark. Pruning and quantization introduce more noticeable artifacts, particularly on Fashion-MNIST (e.g., SSIM decreasing to 0.802 with a corresponding LPIPS of 0.334). Nonetheless, the combination of high PSNR and low LPIPS across datasets confirms that watermark information is embedded within stable feature subspaces that remain resilient to common model perturbations, thereby ensuring reliable verification even after substantial model modification.

We illustrate the watermarks against overwriting attacks in \Cref{fig:overwrite}. Even when an adversary injects its own watermark images to overwrite existing ownership signals, the reconstructed watermarks display noticeable blurring and structural degradation. In contrast, legitimate clients’ watermarks (right columns) retain clear semantic features and recognizable structures. The adversary cannot successfully overwrite the watermark because it only has access to the global model parameters but lacks the corresponding BN statistics required for the watermarking task. Without these statistics, accurately reconstructing the transposed model becomes infeasible, making it difficult for the adversary to overwrite legitimate  watermarks.

\noindent \textbf{Watermark Security.} 
To evaluate the forgery attacks, in \Cref{tab:asr}, we presents the ASR under various parameter configurations $(m, y, num)$ across datasets. In targeted forgery attacks, ASR strongly depends on the margin $m$ and verification threshold $\tau$. When $m = 0.5$, ASR approaches 100\% at $\tau = 0.1$, indicating that low thresholds enable forged triggers closely mimic genuine watermarks. As $\tau$ increases beyond 0.5, ASR declines sharply across all datasets. This demonstrates the security of the proposed verification mechanism under stricter criteria. Conversely, untargeted forgery attacks yield near-zero ASR under all settings, confirming that random inputs cannot trigger valid watermark responses. Moreover, increasing input-vector count $num$ or sample weight $y$ provides negligible ASR improvement. Comprehensive ASR results and analysis are provided in Appendix~\ref{app:asr}.

\subsection{Ablation Study}
\label{sec:ablation}

\begin{table}[t]
\centering
\caption{Performance under modification attacks with and without contrastive learning}
\label{tab:ma-abl}
\renewcommand{\arraystretch}{1.11}
\setlength{\tabcolsep}{3pt}

\resizebox{\columnwidth}{!}{
\begin{tabular}{cccccccccc}
\toprule
\multicolumn{2}{c}{\multirow{2}{*}{\textbf{Modification Attack}}} & 
\multicolumn{2}{c}{\textbf{MNIST}} & 
\multicolumn{2}{c}{\textbf{Fashion-MNIST}} & 
\multicolumn{2}{c}{\textbf{CIFAR-10}} & 
\multicolumn{2}{c}{\textbf{CIFAR-100}} \\
\cmidrule(lr){3-4}\cmidrule(lr){5-6}\cmidrule(lr){7-8}\cmidrule(lr){9-10}
& & \textbf{Acc(\%)} & \textbf{SSIM} & \textbf{Acc(\%)} & \textbf{SSIM} & \textbf{Acc(\%)} & \textbf{SSIM} & \textbf{Acc(\%)} & \textbf{SSIM} \\
\midrule

\multirow{2}{*}{Pruning (40\%)} 
& w/o CL & 99.42 & 0.98 & 88.56 & 0.46 & 91.07 & 0.91 & 67.77 & 0.70 \\
& w/ CL   & 99.24 & 0.96 & 89.64 & 0.61 & 90.36 & 0.97 & 68.47 & 0.92 \\
\midrule

\multirow{2}{*}{Fine-tuning (20 rounds)} 
& w/o CL & 99.57 & 0.99 & 89.71 & 0.77 & 92.63 & 0.92 & 70.64 & 0.72 \\
& w/ CL   & 99.45 & 0.99 & 90.61 & 0.94 & 92.39 & 0.97 & 68.08 & 0.92 \\
\midrule

\multirow{2}{*}{Quantization (4-bit)} 
& w/o CL & 99.44 & 0.99 & 85.14 & 0.44 & 90.68 & 0.89 & 66.20 & 0.76 \\
& w/ CL   & 99.16 & 0.97 & 87.73 & 0.55 & 91.38 & 0.95 & 67.22 & 0.92 \\
\midrule

\multirow{2}{*}{Overwriting (50 rounds)} 
& w/o CL & 92.63 & 0.65 & 87.12 & 0.39 & 90.98 & 0.86 & 68.17 & 0.76 \\
& w/ CL   & 85.31 & 0.91 & 88.84 & 0.80 & 86.70 & 0.94 & 65.61 & 0.81 \\
\bottomrule
\end{tabular}
}
\end{table}

\begin{table}[t]
\centering
\caption{ASR results with/without contrastive learning}
\label{tab:asr-nocl}
\renewcommand{\arraystretch}{1.1}
\setlength{\tabcolsep}{4pt}
\resizebox{\columnwidth}{!}{
\begin{tabular}{cccccc}
\toprule
\multirow{2}{*}{\shortstack{\textbf{Attack}\\\textbf{Type}}} &
\multirow{2}{*}{\textbf{Threshold $\tau$}} &
\multicolumn{4}{c}{\textbf{w/o Contrastive Learning}} \\
\cmidrule(lr){3-6}
& & \textbf{MNIST} & \textbf{Fashion-MNIST} & \textbf{CIFAR-10} & \textbf{CIFAR-100} \\
\midrule

\multirow{5}{*}{\textbf{Target (\%)}} 
& 0.1 & 100.00 ({\color{red} $\uparrow$ 66.30}) & 100.00  & 100.00 ({\color{red} $\uparrow$ 82.30}) & 100.00 ({\color{red} $\uparrow$ 21.70}) \\
& 0.3 & 99.70 ({\color{red} $\uparrow$ 74.70}) & 100.00 ({\color{red} $\uparrow$ 90.70}) & 100.00 ({\color{red} $\uparrow$ 99.30}) & 100.00 ({\color{red} $\uparrow$ 99.70}) \\
& 0.5 & 68.00 ({\color{red} $\uparrow$ 46.00}) & 37.30 ({\color{red} $\uparrow$ 37.30}) & 48.30 ({\color{red} $\uparrow$ 48.30}) & 65.70 ({\color{red} $\uparrow$ 65.70}) \\
& 0.7 & 67.00 ({\color{red} $\uparrow$ 50.70}) & \cellcolor{green!20}0.00 & 36.00 ({\color{red} $\uparrow$ 36.00}) & 0.70 ({\color{red} $\uparrow$0.70}) \\
& 0.9 & 55.30 ({\color{red} $\uparrow$ 55.30})& \cellcolor{green!20}0.00 & 21.00 ({\color{red} $\uparrow$ 21.00}) & \cellcolor{green!20}0.00 \\
\midrule

\multirow{5}{*}{\textbf{Untarget (\%)}} 
& 0.1 & 100.00 ({\color{red} $\uparrow$ 94.70}) & 100.00 ({\color{red} $\uparrow$ 0.70}) & 100.00 ({\color{red} $\uparrow$93.30}) & 100.00  \\
& 0.3 & \cellcolor{green!20}0.00 & \cellcolor{green!20}0.00 & \cellcolor{green!20}0.00 & \cellcolor{green!20}0.00 \\
& 0.5 & \cellcolor{green!20}0.00 & \cellcolor{green!20}0.00 & \cellcolor{green!20}0.00 & \cellcolor{green!20}0.00 \\
& 0.7 & \cellcolor{green!20}0.00 & \cellcolor{green!20}0.00 & \cellcolor{green!20}0.00 & \cellcolor{green!20}0.00 \\
& 0.9 & \cellcolor{green!20}0.00 & \cellcolor{green!20}0.00 & \cellcolor{green!20}0.00 & \cellcolor{green!20}0.00 \\
\bottomrule
\end{tabular}
}
\end{table}

\begin{figure*}[t]
	\centering
	\begin{minipage}[c]{0.02\columnwidth}
     	\centering
     	\rotatebox{90}{\tiny{\textbf{MNIST}}}
    \end{minipage}%
    \begin{minipage}[c]{0.32\columnwidth} 
        \centering
        \caption*{\scriptsize{Pruning: Acc}}
        \vspace{-3mm}
        \includegraphics[width=\columnwidth]{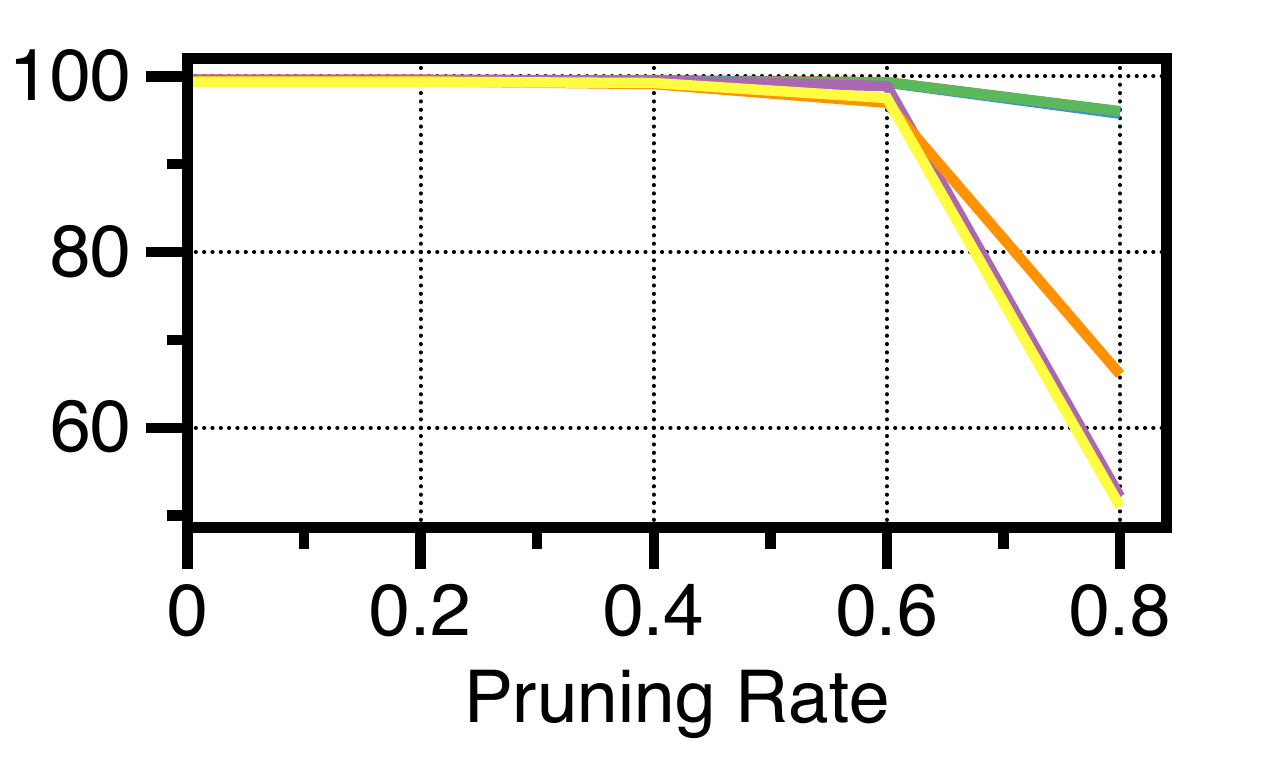} 
    \end{minipage}
    \begin{minipage}[c]{0.32\columnwidth} 
        \centering
        \caption*{\scriptsize{Pruning: WM-Acc}}
        \vspace{-3mm}
        \includegraphics[width=\columnwidth]{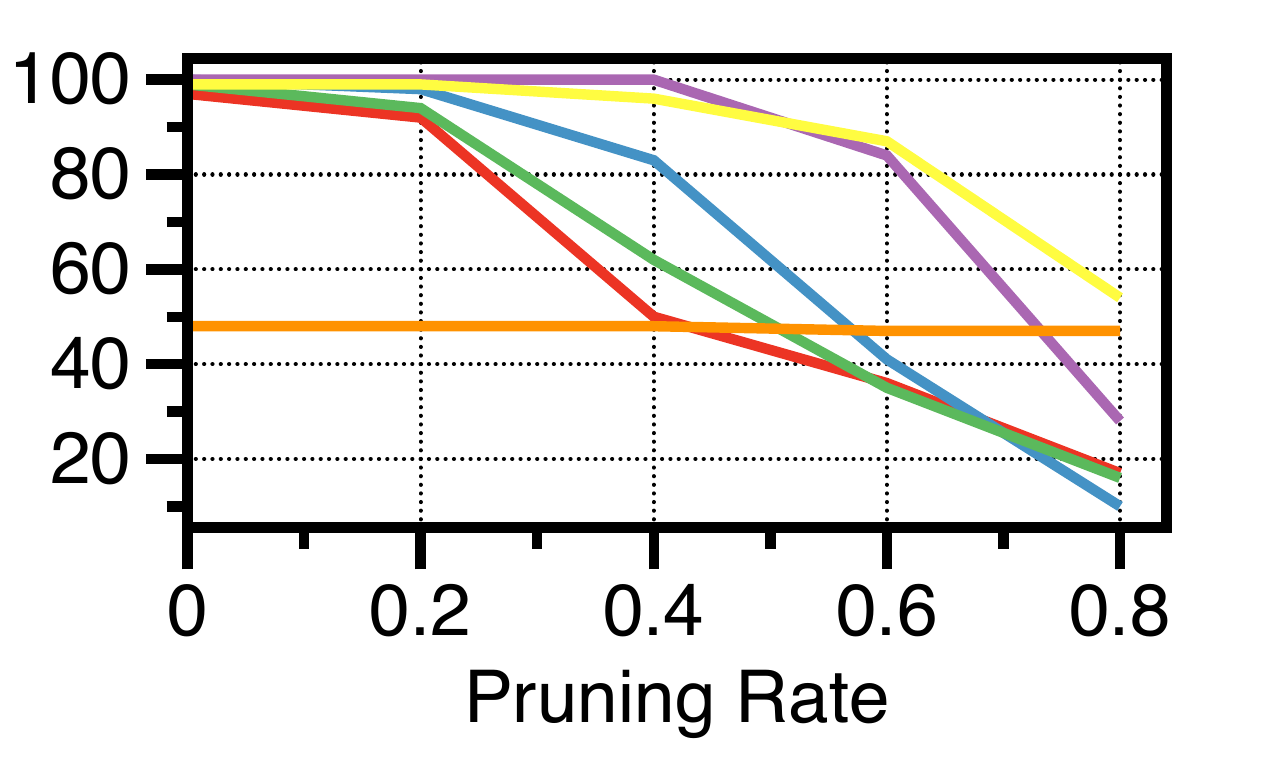} 
    \end{minipage}
    \begin{minipage}[c]{0.32\columnwidth} 
        \centering
        \caption*{\scriptsize{Fine-tuning: Acc}}
        \vspace{-3mm}
        \includegraphics[width=\columnwidth]{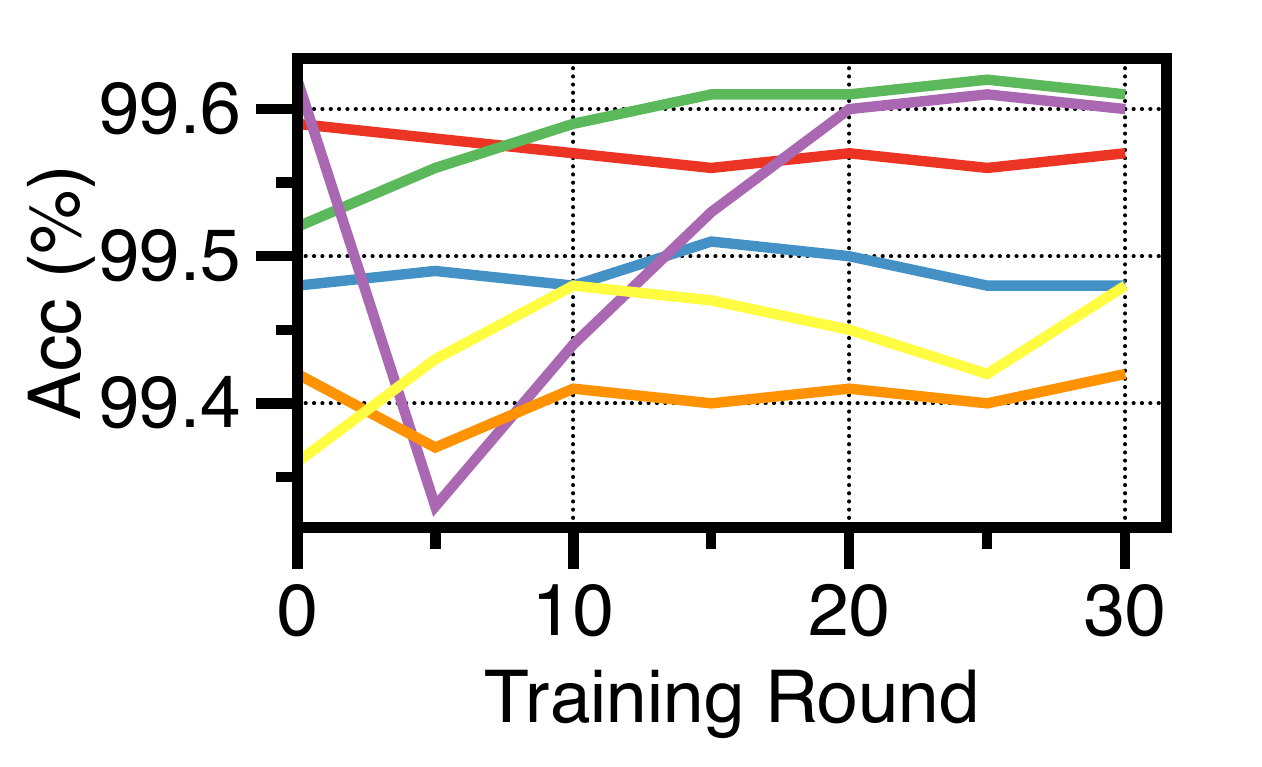} 
    \end{minipage}
    \begin{minipage}[c]{0.32\columnwidth} 
        \centering
        \caption*{\scriptsize{Fine-tuning: WM-Acc}}
        \vspace{-3mm}
        \includegraphics[width=\columnwidth]{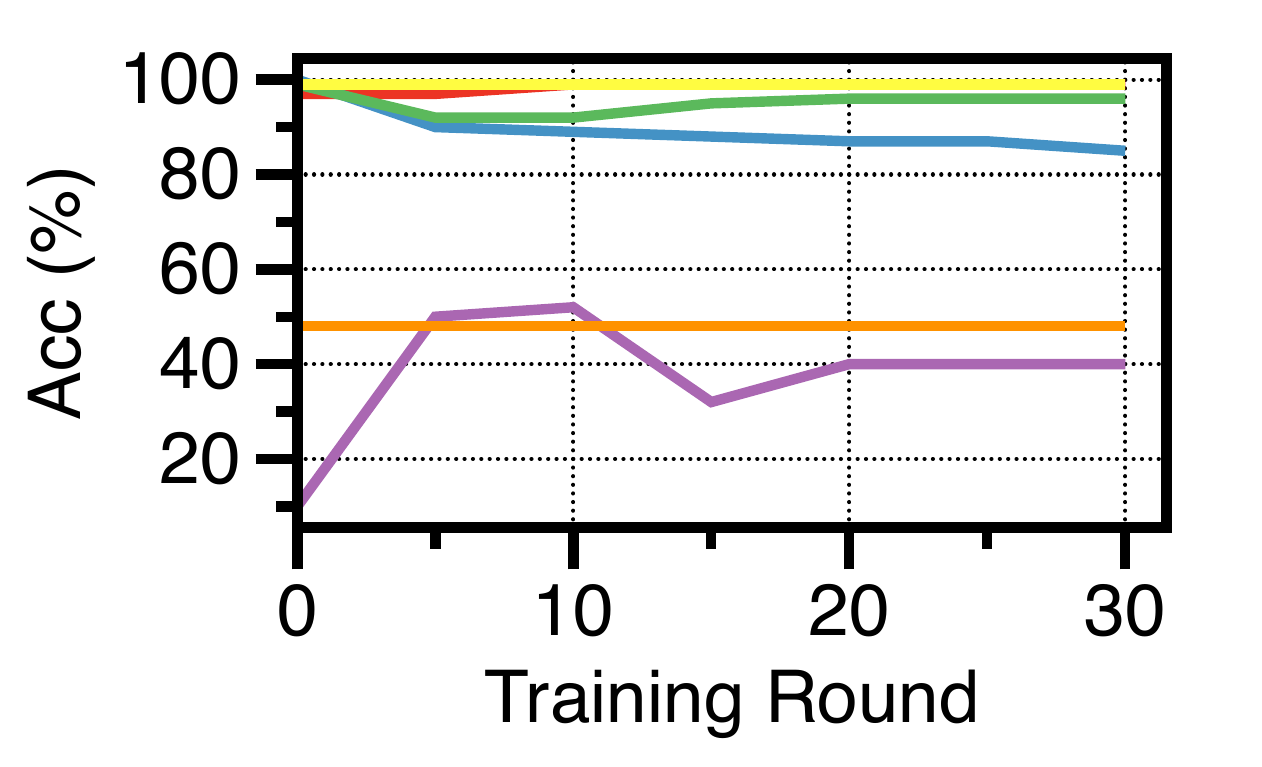} 
    \end{minipage}
    \begin{minipage}[c]{0.32\columnwidth} 
        \centering
        \caption*{\scriptsize{Quantization: Acc}}
        \vspace{-3mm}
        \includegraphics[width=\columnwidth]{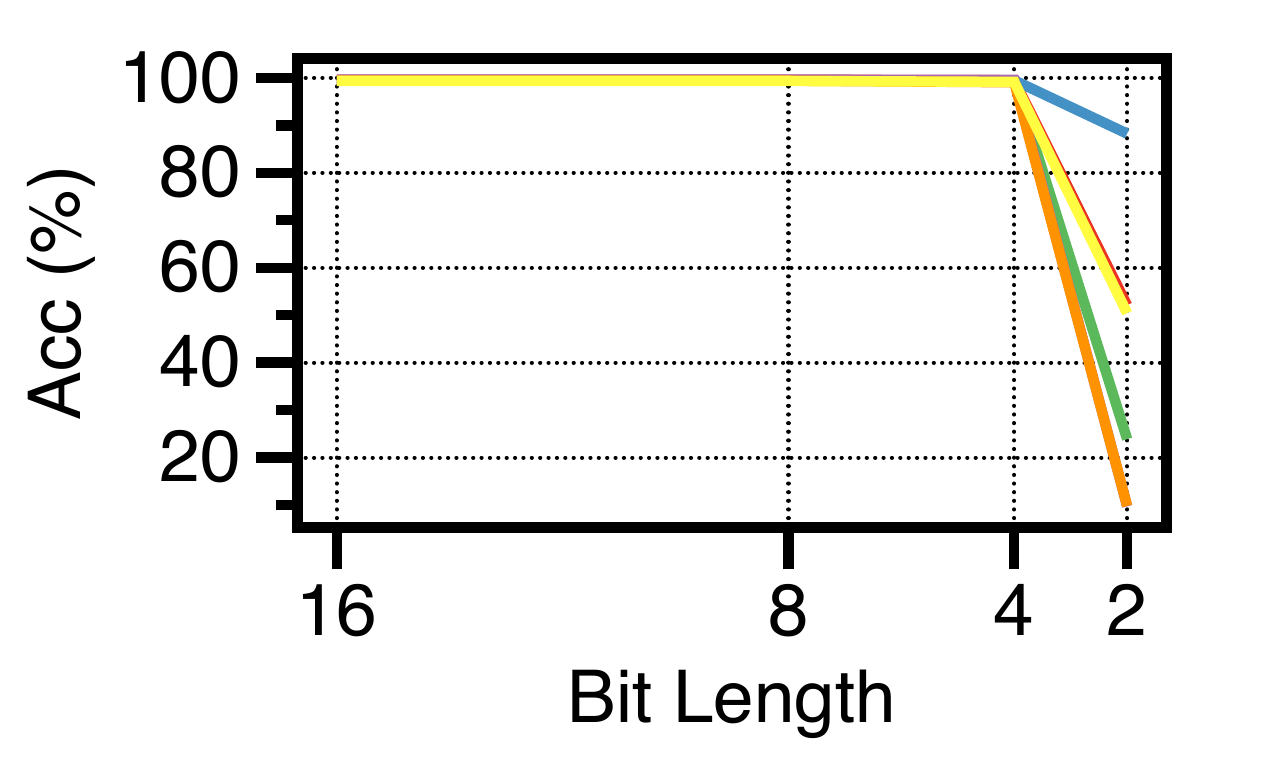} 
    \end{minipage}
    \begin{minipage}[c]{0.32\columnwidth} 
        \centering
        \caption*{\scriptsize{Quantization: WM-Acc}}
        \vspace{-3mm}
        \includegraphics[width=\columnwidth]{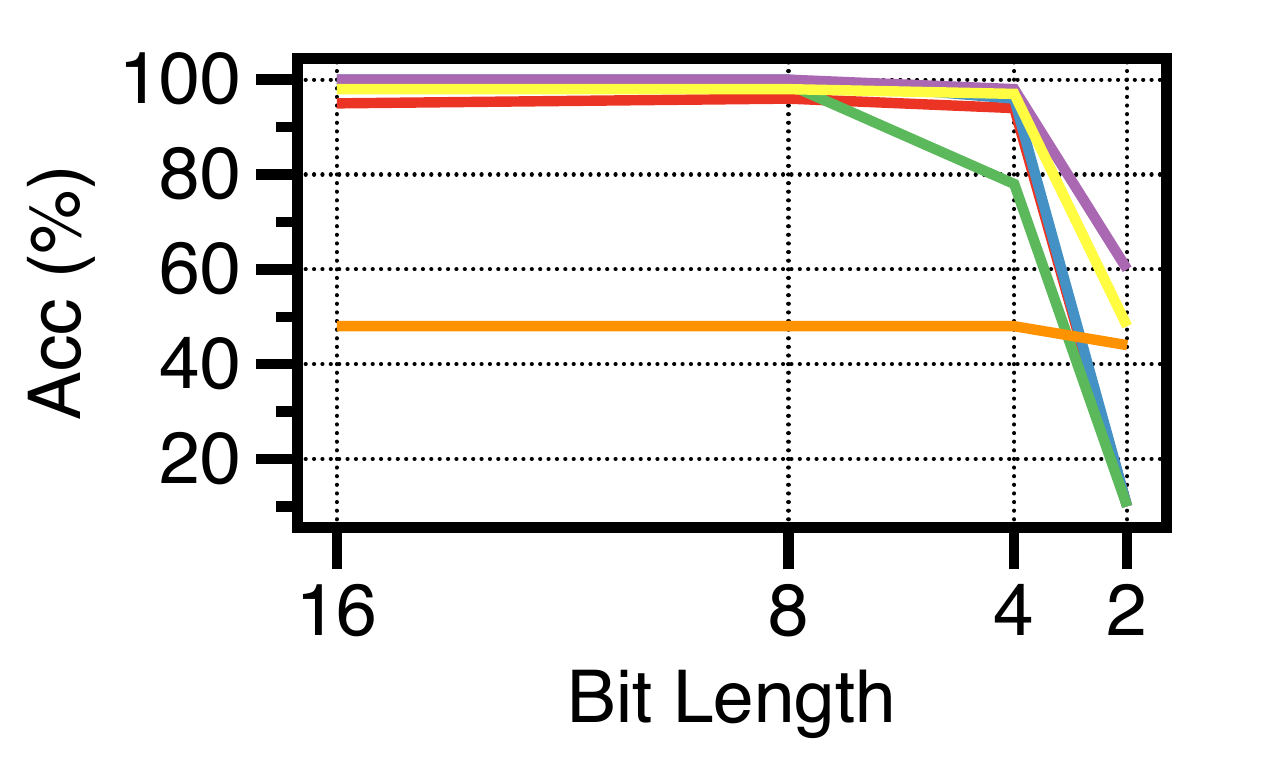} 
    \end{minipage}
    \begin{minipage}[c]{0.02\columnwidth}
     	\centering
     	\rotatebox{90}{\tiny{\textbf{Fashion-MNIST}}}
    \end{minipage}%
    \begin{minipage}[c]{0.32\columnwidth} 
        \centering
        \includegraphics[width=\columnwidth]{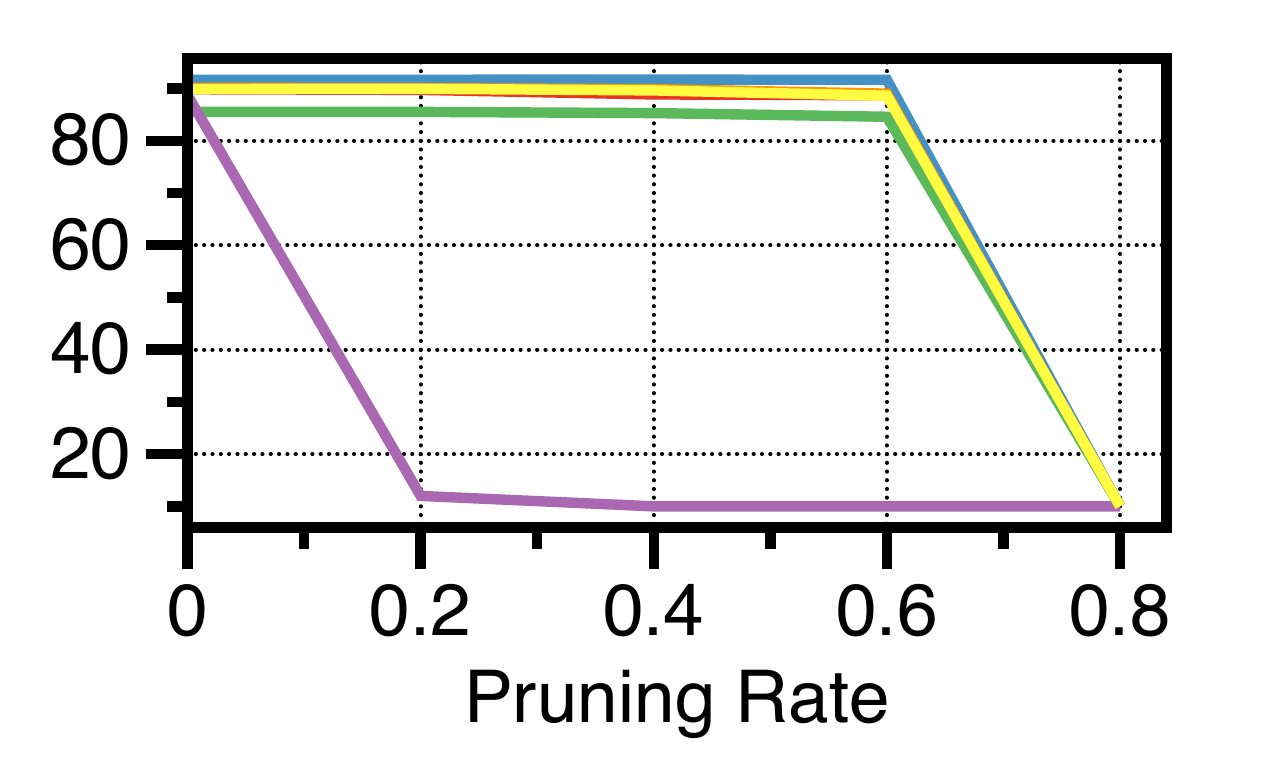} 
    \end{minipage}
    \begin{minipage}[c]{0.32\columnwidth} 
        \centering
        \includegraphics[width=\columnwidth]{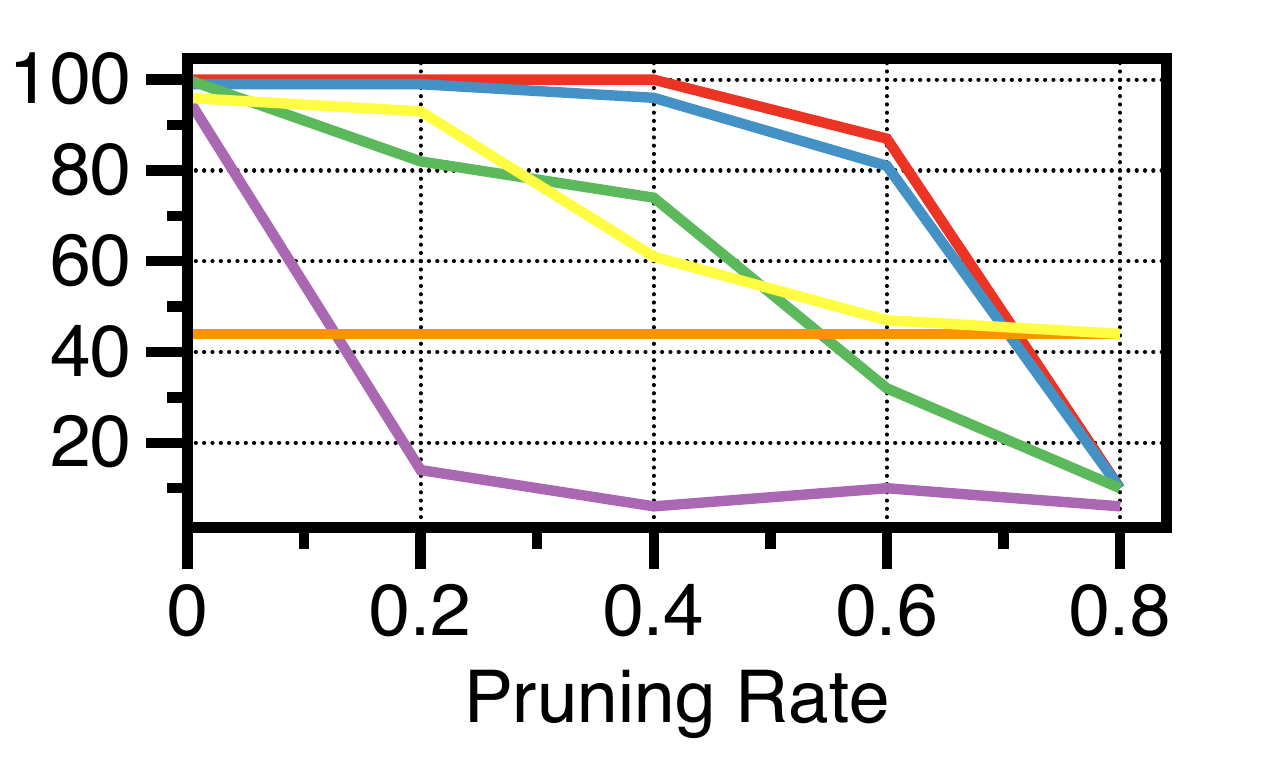} 
    \end{minipage}
    \begin{minipage}[c]{0.32\columnwidth} 
        \centering
        \includegraphics[width=\columnwidth]{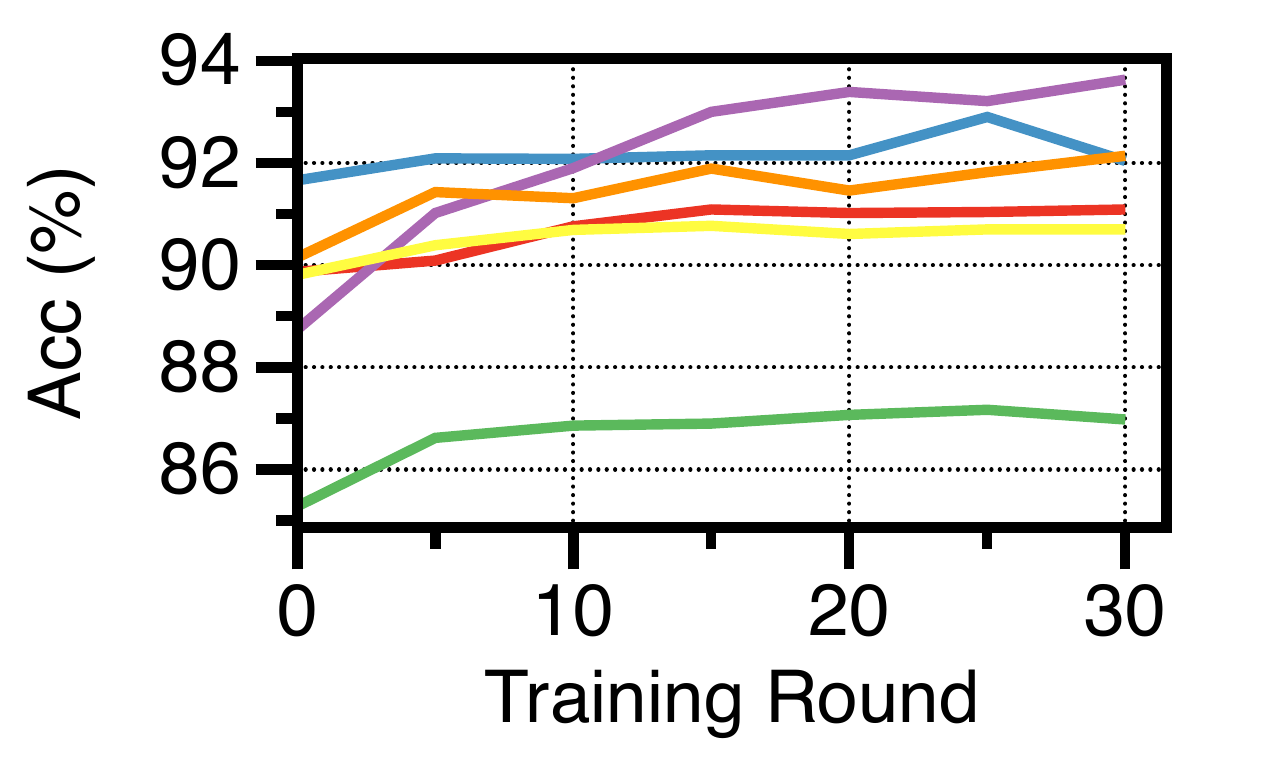} 
    \end{minipage}
    \begin{minipage}[c]{0.32\columnwidth} 
        \centering
        \includegraphics[width=\columnwidth]{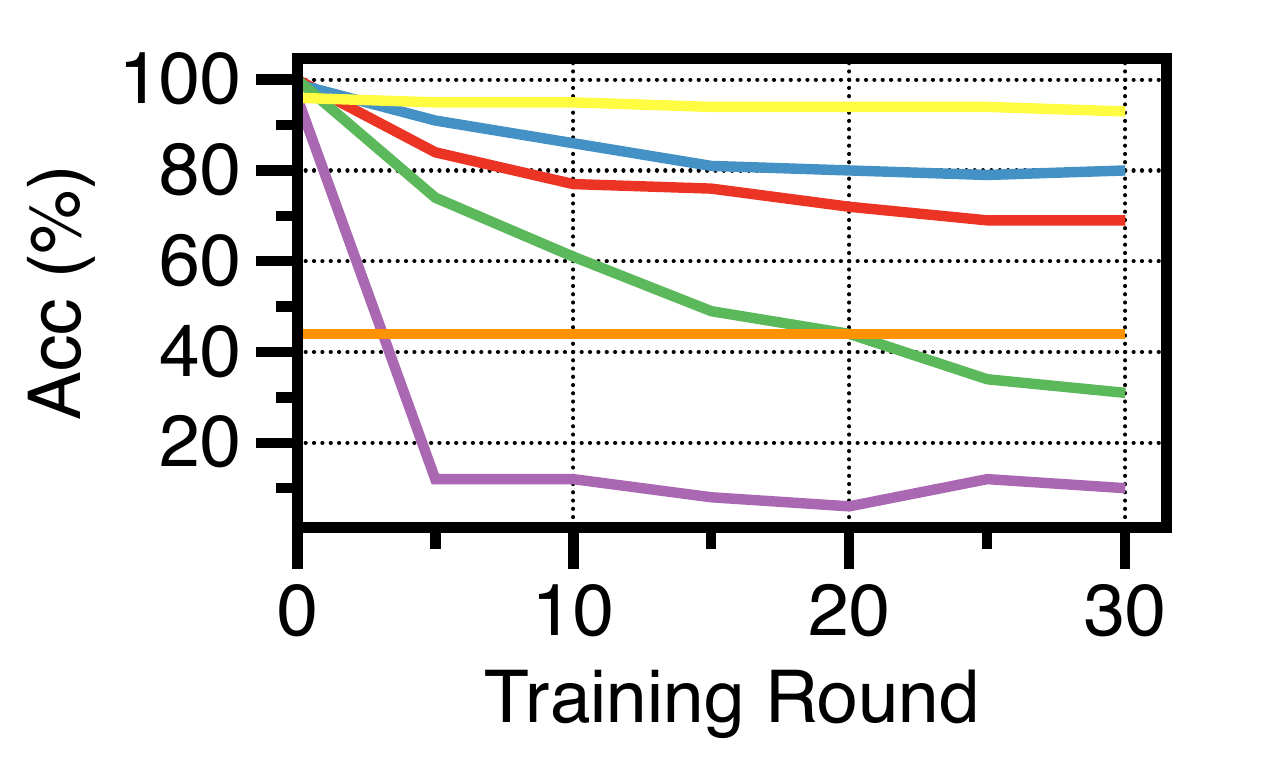} 
    \end{minipage}
    \begin{minipage}[c]{0.32\columnwidth} 
        \centering
        \includegraphics[width=\columnwidth]{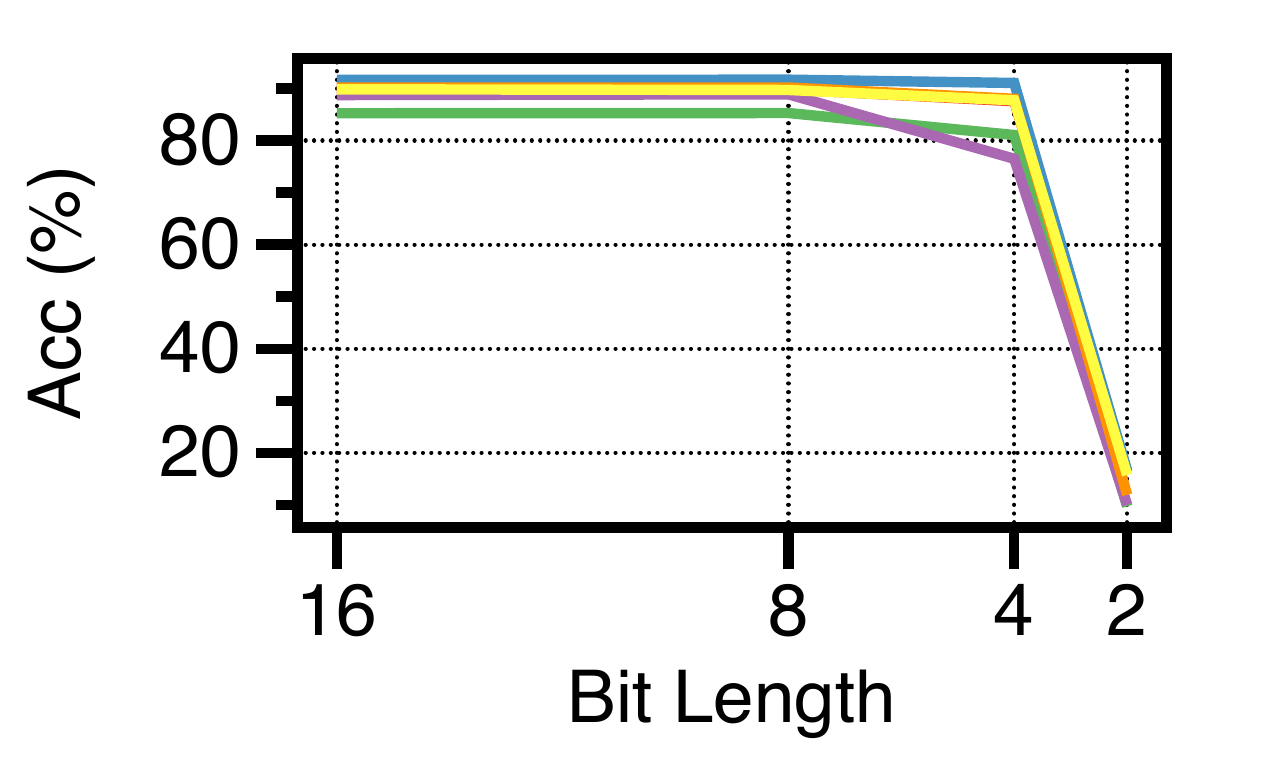} 
    \end{minipage}
    \begin{minipage}[c]{0.32\columnwidth} 
        \centering
        \includegraphics[width=\columnwidth]{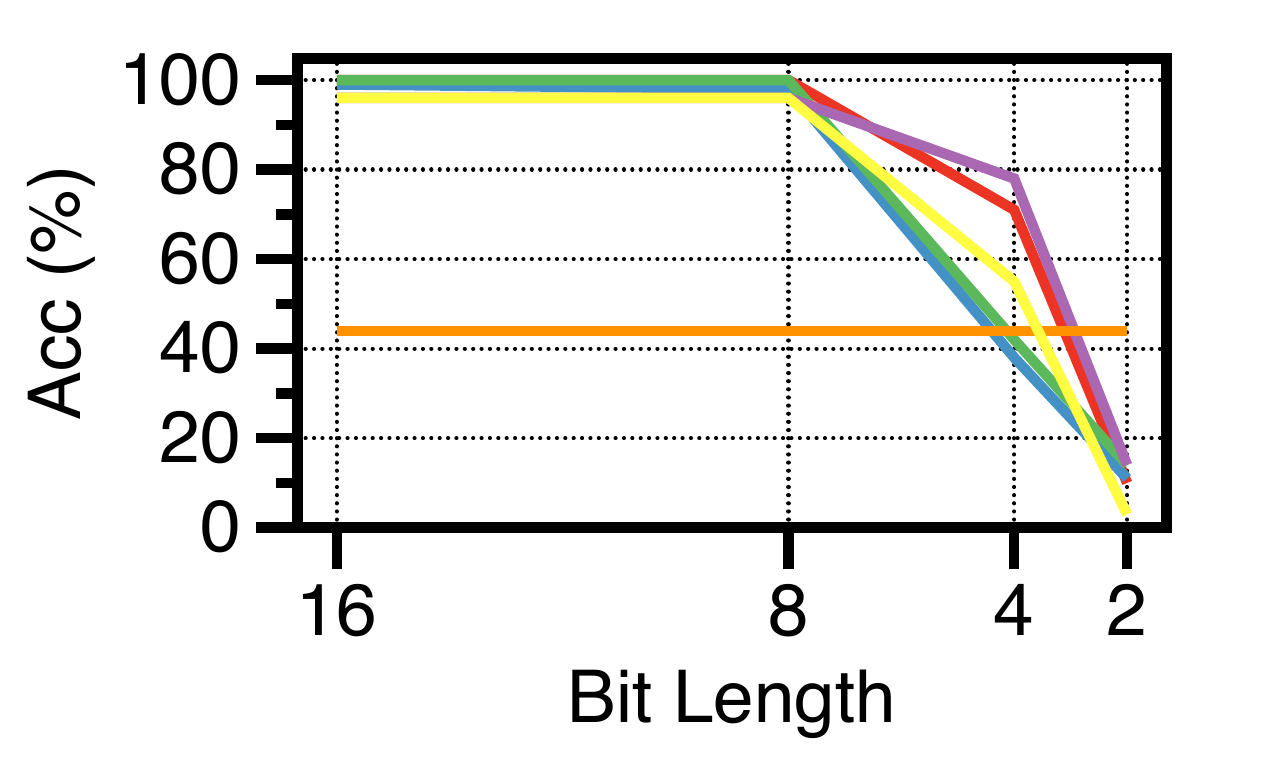} 
    \end{minipage}
    \begin{minipage}[c]{0.02\columnwidth}
     	\centering
     	\rotatebox{90}{\tiny{\textbf{CIFAR-10}}}
    \end{minipage}%
    \begin{minipage}[c]{0.32\columnwidth} 
        \centering
        \includegraphics[width=\columnwidth]{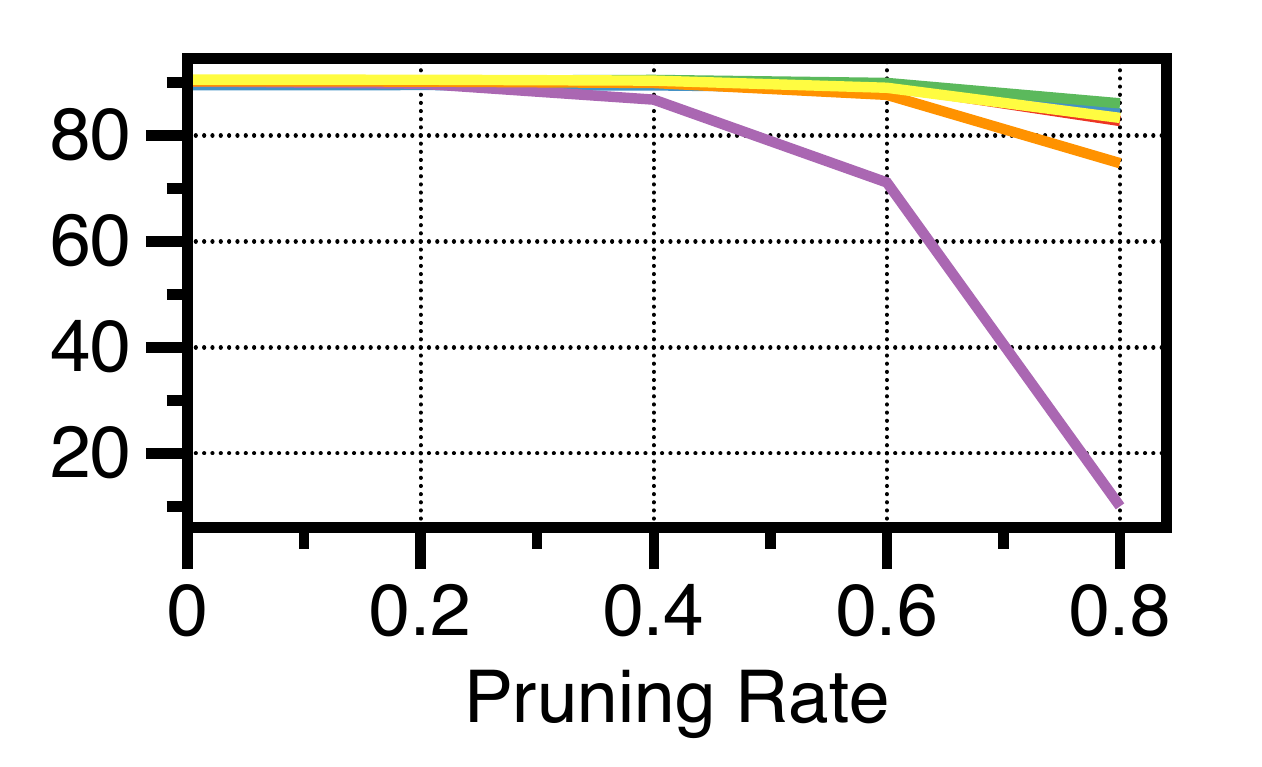} 
    \end{minipage}
    \begin{minipage}[c]{0.32\columnwidth} 
        \centering
        \includegraphics[width=\columnwidth]{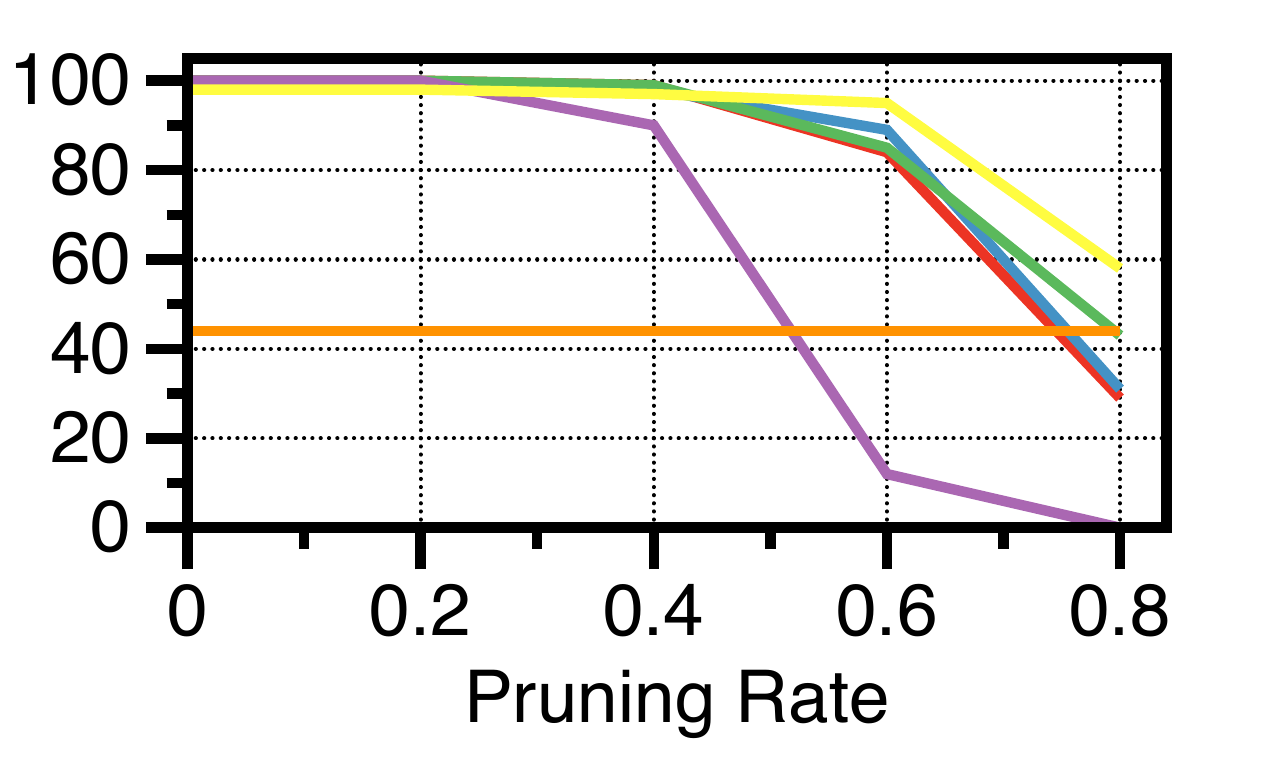} 
    \end{minipage}
    \begin{minipage}[c]{0.32\columnwidth} 
        \centering
        \includegraphics[width=\columnwidth]{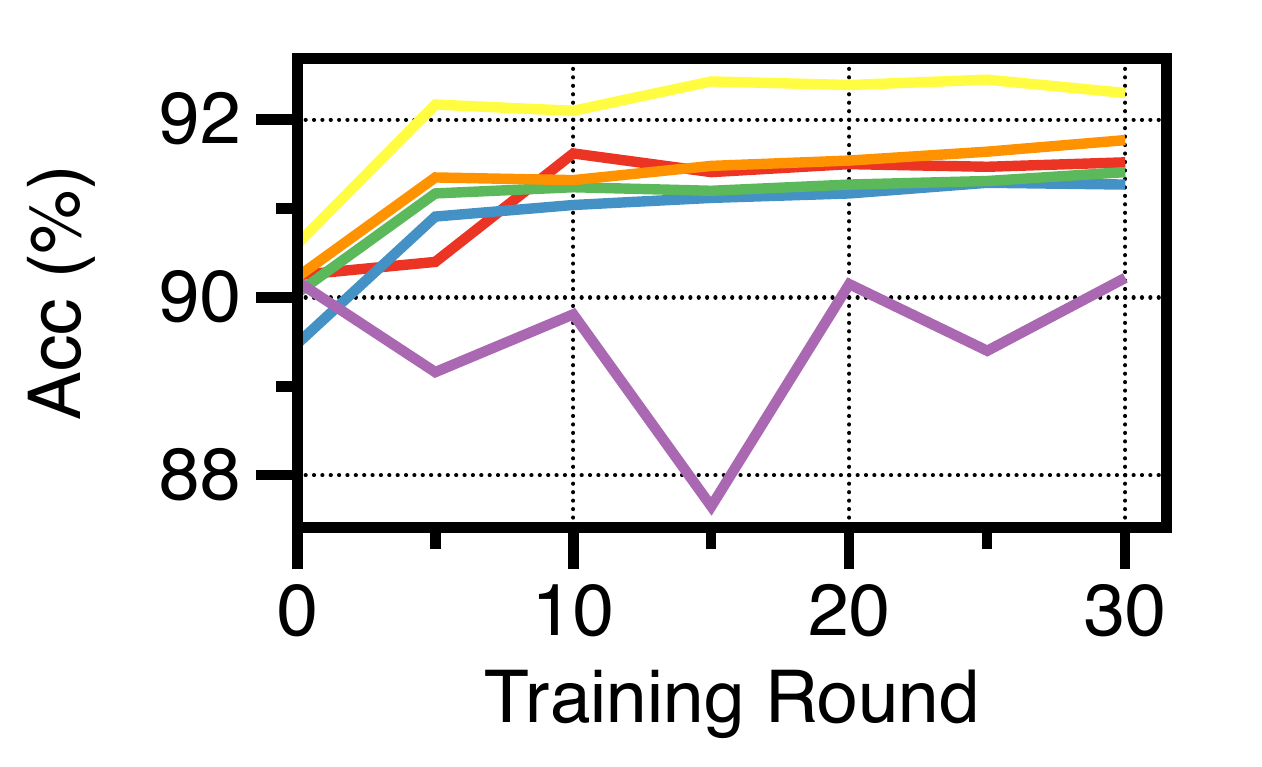} 
    \end{minipage}
    \begin{minipage}[c]{0.32\columnwidth} 
        \centering
        \includegraphics[width=\columnwidth]{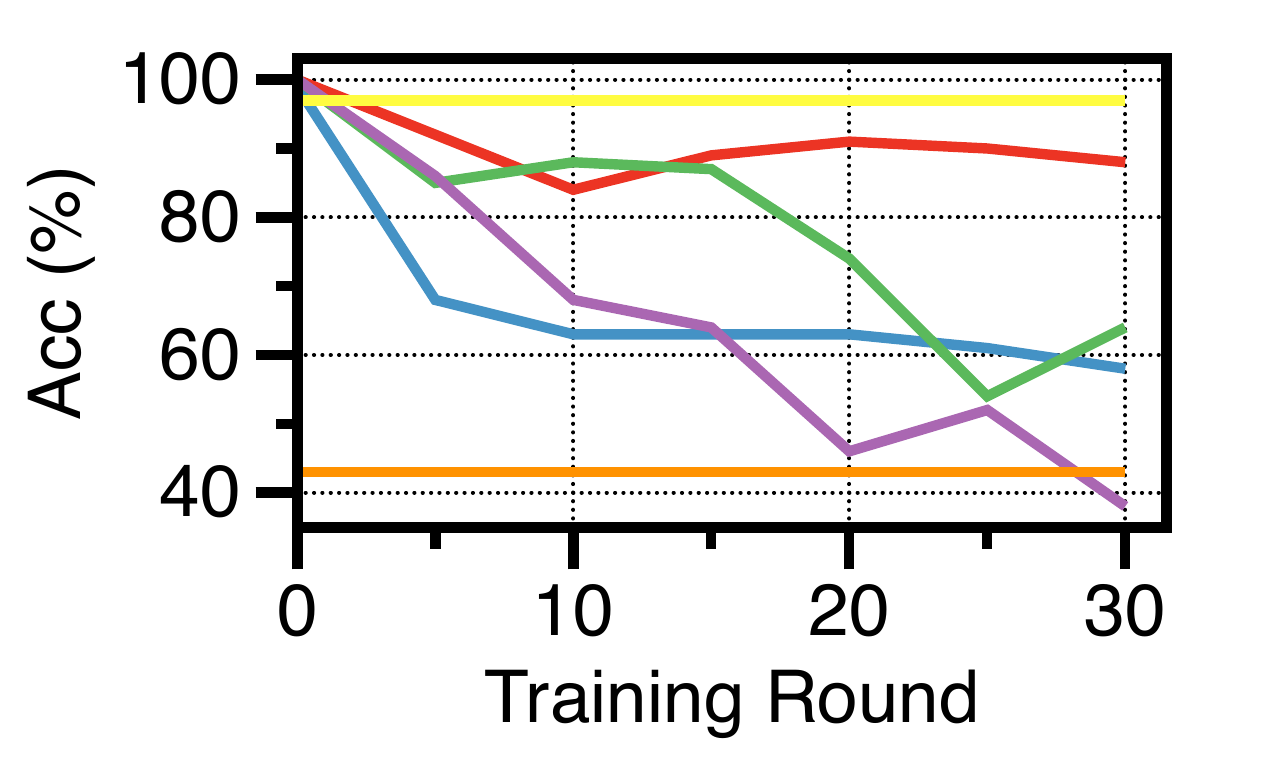} 
    \end{minipage}
    \begin{minipage}[c]{0.32\columnwidth} 
        \centering
        \includegraphics[width=\columnwidth]{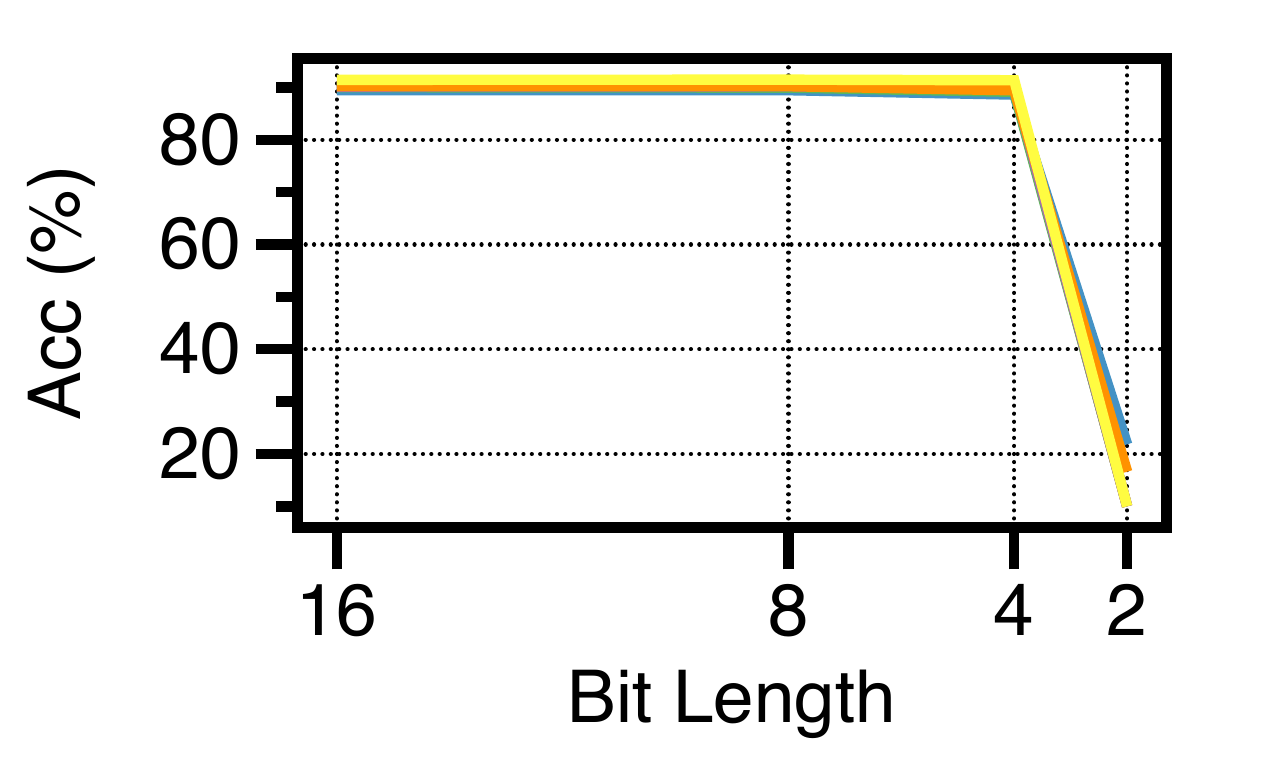} 
    \end{minipage}
    \begin{minipage}[c]{0.32\columnwidth} 
        \centering
        \includegraphics[width=\columnwidth]{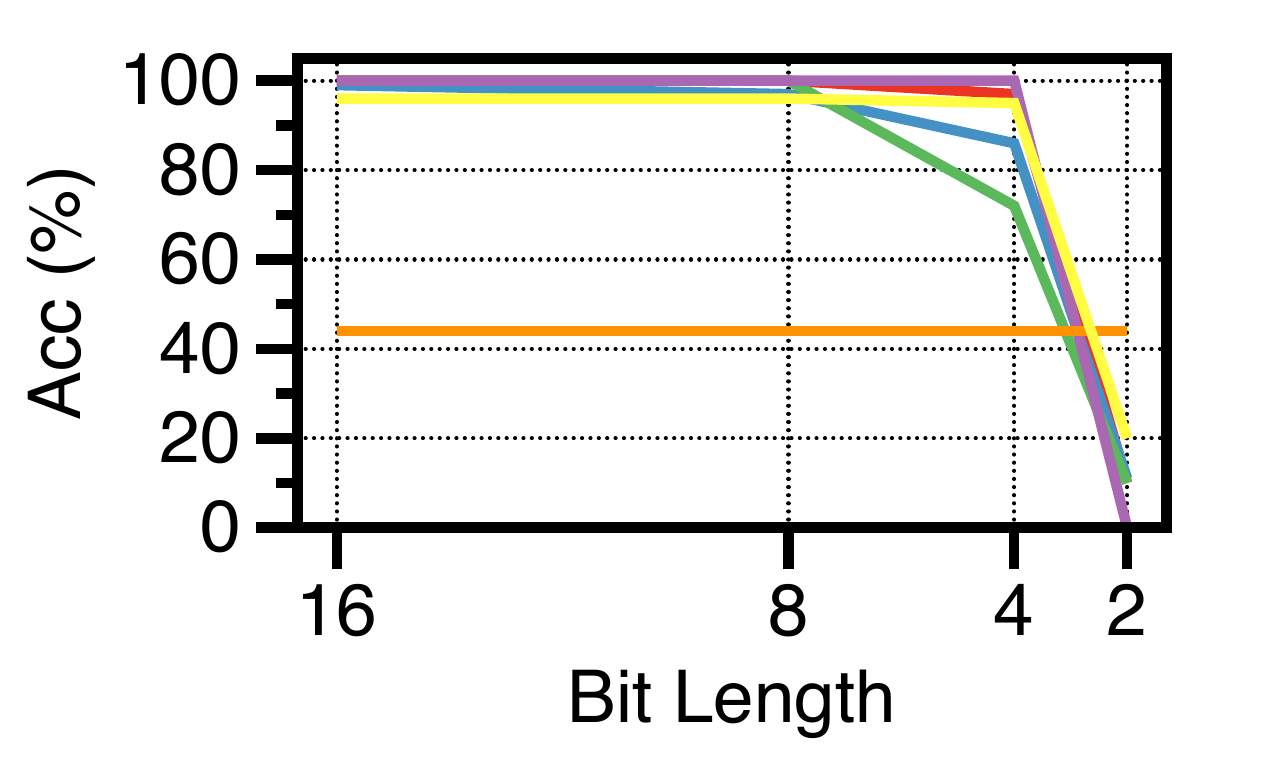} 
    \end{minipage}
    \begin{minipage}[c]{0.02\columnwidth}
     	\centering
     	\rotatebox{90}{\tiny{\textbf{CIFAR-100}}}
    \end{minipage}%
    \begin{minipage}[c]{0.32\columnwidth} 
        \centering
        \includegraphics[width=\columnwidth]{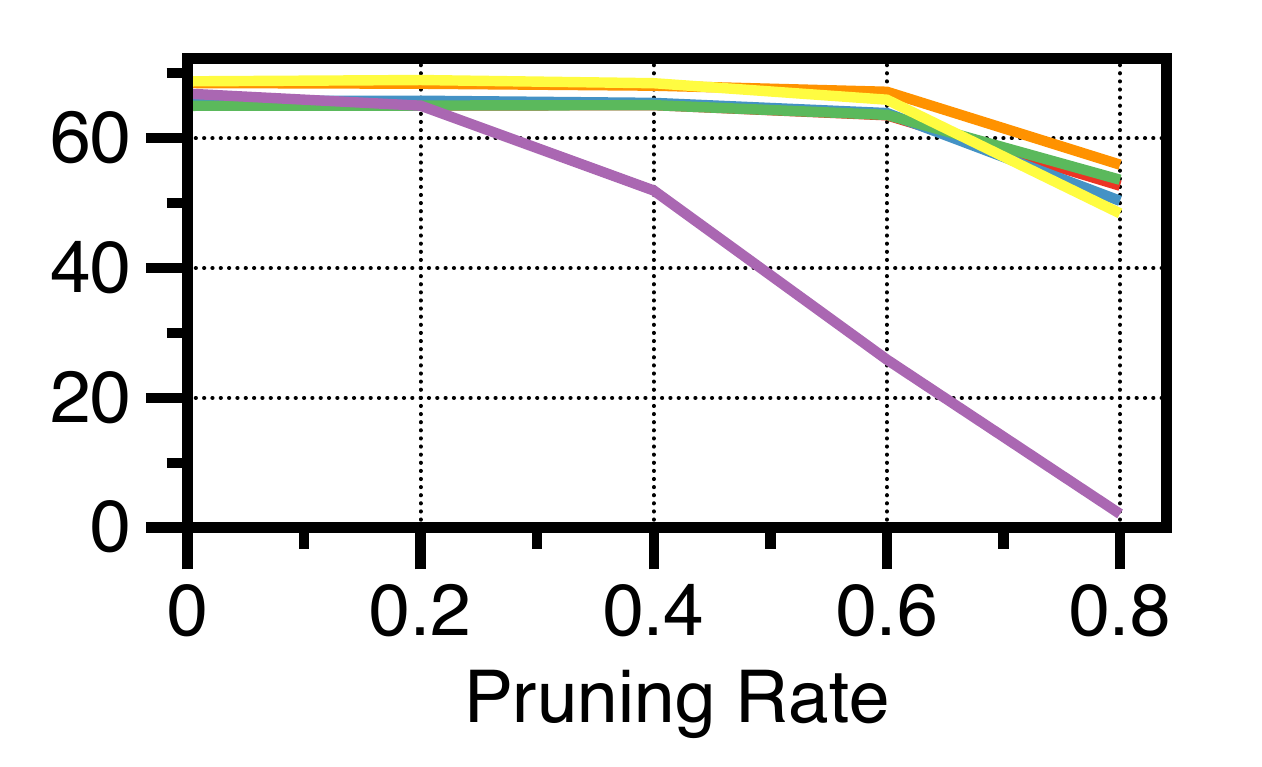} 
    \end{minipage}
    \begin{minipage}[c]{0.32\columnwidth} 
        \centering
        \includegraphics[width=\columnwidth]{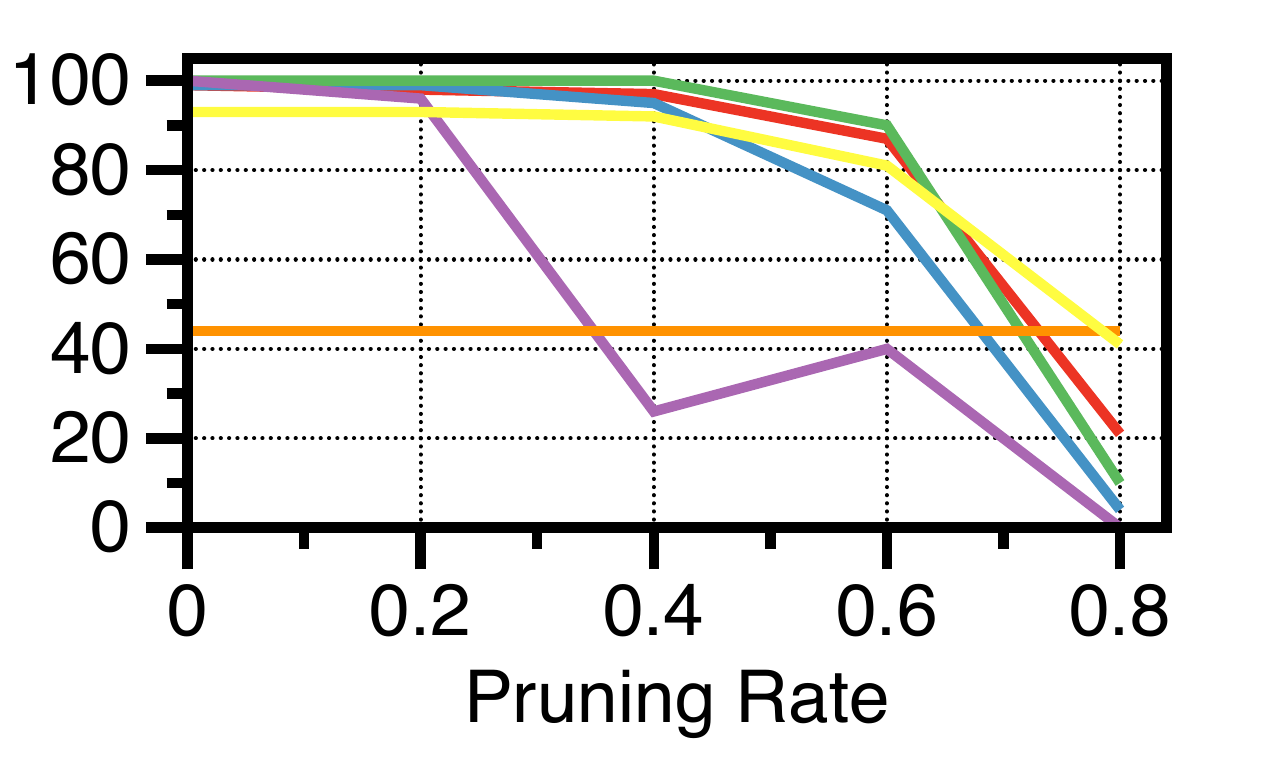} 
    \end{minipage}
    \begin{minipage}[c]{0.32\columnwidth} 
        \centering
        \includegraphics[width=\columnwidth]{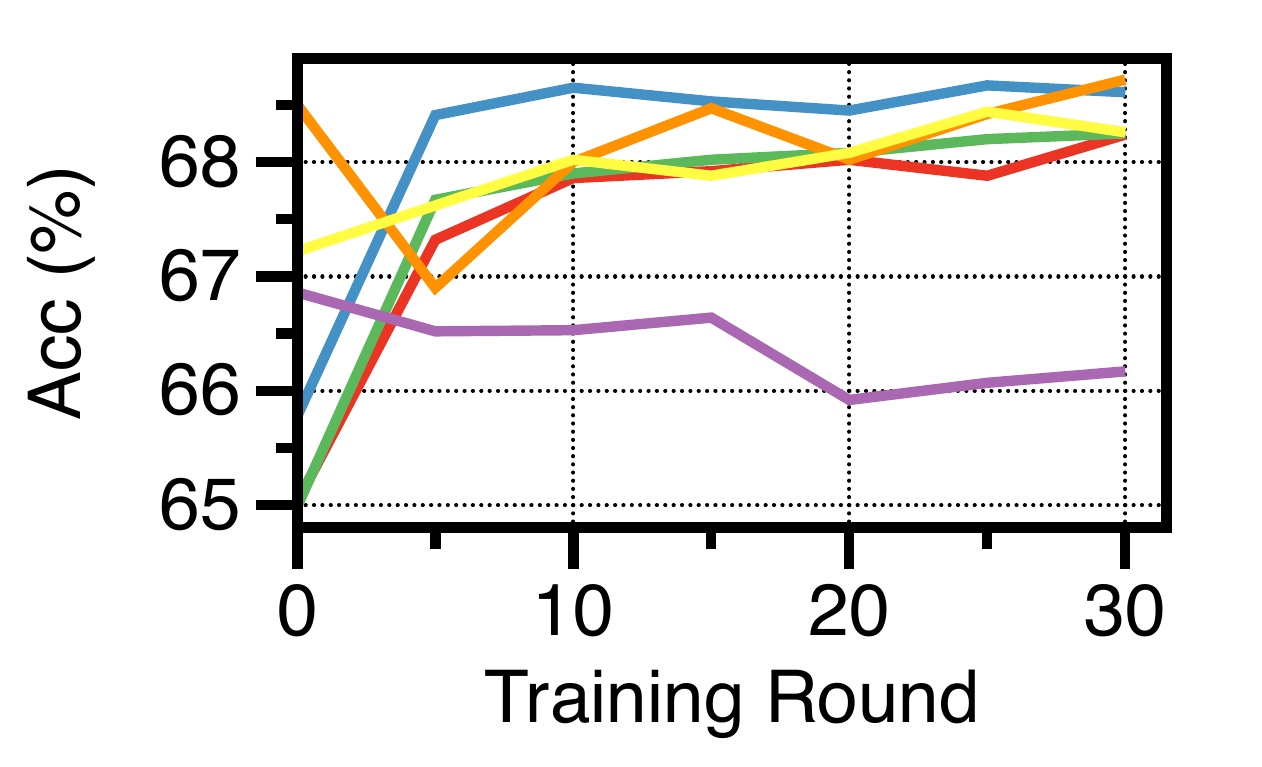} 
    \end{minipage}
    \begin{minipage}[c]{0.32\columnwidth} 
        \centering
        \includegraphics[width=\columnwidth]{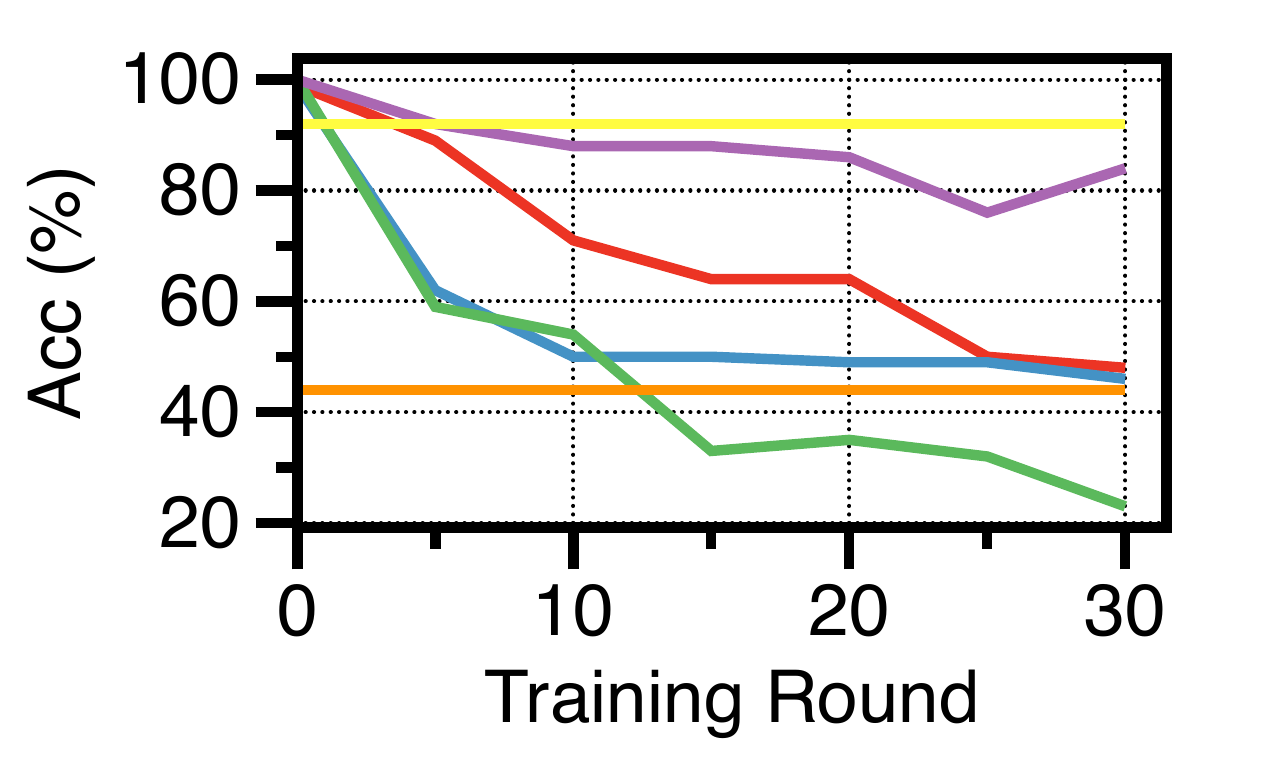} 
    \end{minipage}
    \begin{minipage}[c]{0.32\columnwidth} 
        \centering
        \includegraphics[width=\columnwidth]{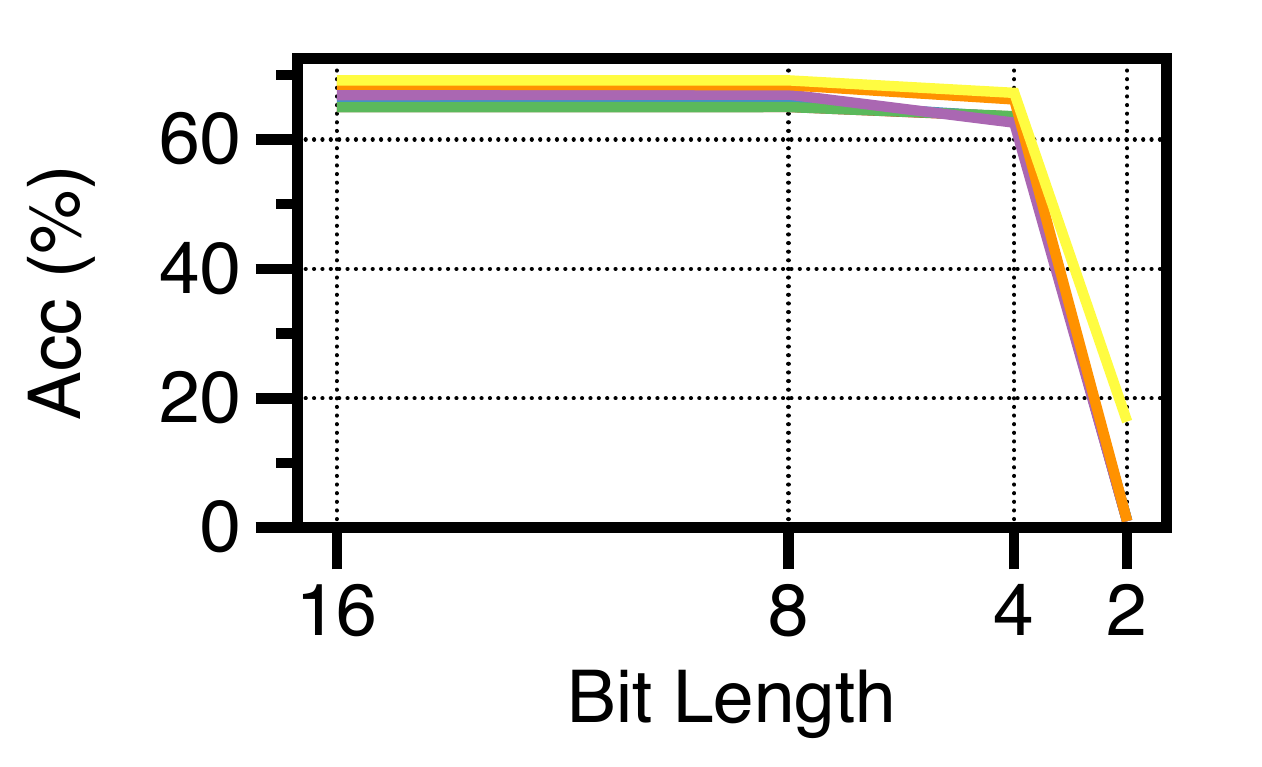} 
    \end{minipage}
    \begin{minipage}[c]{0.32\columnwidth} 
        \centering
        \includegraphics[width=\columnwidth]{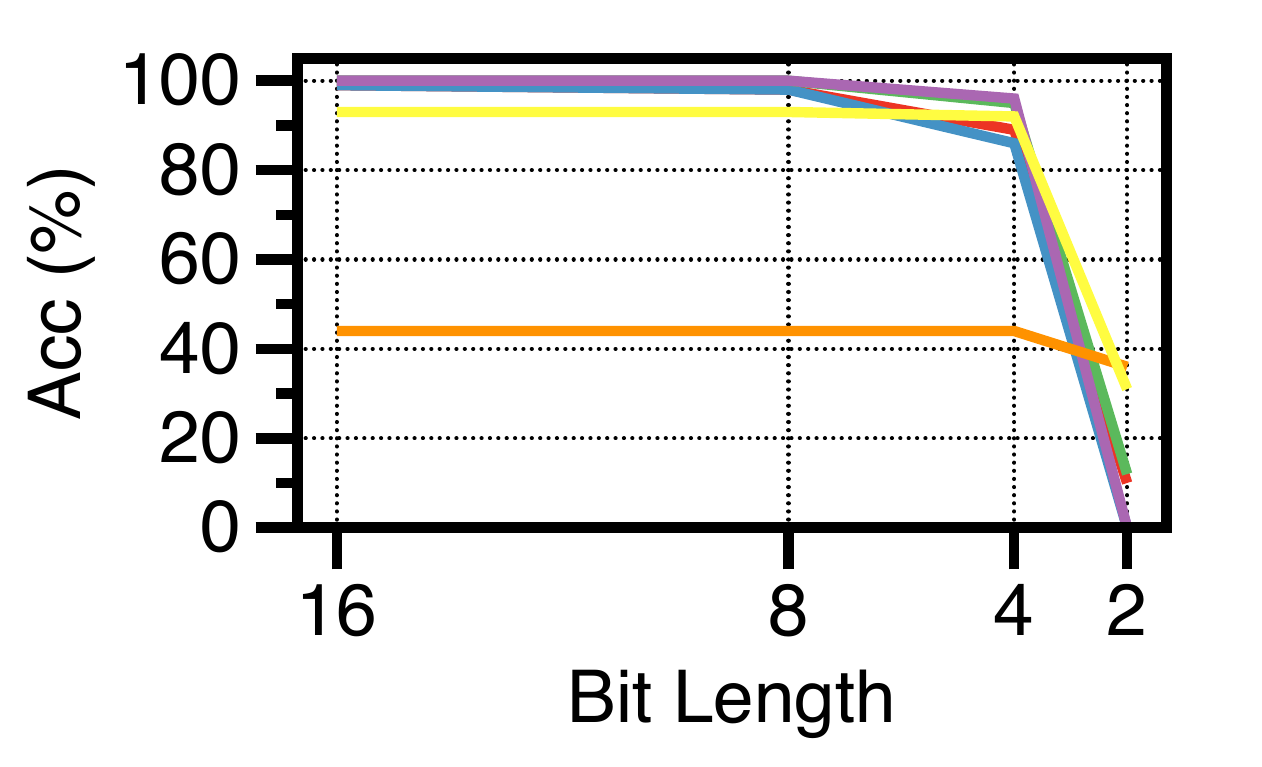} 
    \end{minipage}
   \begin{subfigure}{1.2\columnwidth}
     \centering
     \includegraphics[width=\columnwidth]{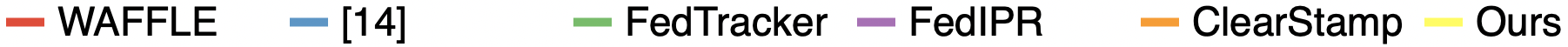}
   \end{subfigure}
\caption{Comparison results under modification attacks}
\label{fig:comp-ma}
\end{figure*}

The core component of \texttt{FLClear} is the integration of contrastive learning. We first evaluate its effect on model accuracy and watermark fidelity through ablation experiments conducted with and without contrastive learning. As shown in \Cref{fig:acc-ssim-abl}, incorporating contrastive learning introduces only a negligible effect on both model accuracy and watermark fidelity without attacks. In both cases, model accuracy remains nearly identical, with differences of less than 1\%. Similarly, SSIM values of reconstructed watermarks show minimal variation.

We further investigate whether model accuracy and watermark fidelity are preserved under model modification attacks in \Cref{tab:ma-abl}. Across all datasets and attack types, incorporating contrastive learning consistently improves watermark resilience without compromising model performance. For instance, under pruning and quantization, SSIM improves markedly from 0.46 to 0.61 on Fashion-MNIST and from 0.76 to 0.92 on CIFAR-100. Similarly, fine-tuning and overwriting attacks show substantial SSIM gains. The minimal change in accuracy ($<$1\%) across all settings demonstrates that contrastive learning introduces negligible task interference. 

As discussed earlier, the contrastive loss is introduced to explicitly counter forgery attacks. To evaluate its effectiveness, \Cref{tab:asr-nocl} reports the ASR for watermark forgery attacks when the contrastive learning is removed. The parameters are set to $(m=0.1, y=0.1, num=250)$. Compared to \Cref{tab:asr}, the results clearly show that removing contrastive learning substantially increases ASR across all datasets, particularly under low thresholds (\(\tau \leq 0.3\)). For instance, ASR for targeted attacks rises by up to \(+99.3\%\) on CIFAR-10 and \(+99.7\%\) on CIFAR-100. 
Similar behavior is observed for untargeted attacks, where the ASR increases to as high as 94.7\% under small values of $\tau$. These results confirm that the contrastive objective effectively regularizes the latent feature space by enforcing intra-watermark compactness and inter-watermark separation, thereby jointly suppressing forgery attempts even under relatively low verification thresholds.

\subsection{Comparison with Baselines}
\label{sec:comparison}

We compare \texttt{FLClear} against existing watermarking methods under three model modification attacks: pruning, fine-tuning, and quantization. Because overwriting and forgery attacks in this paper are specifically tailored to target the \texttt{FLClear} framework, other watermarking schemes are not directly comparable under these conditions. Two metrics of model accuracy (Acc) and watermark accuracy (WM-Acc)\footnote{The compared schemes employ different metrics (e.g., SSIM and BER) to evaluate watermark performance. Since these metrics are ranged from 0 to 1, we report results using a unified metric WM-Acc for consistency and simplicity.} are utilized to assess utility preservation and watermark robustness. 

The results in \Cref{fig:comp-ma} show that, across all attack types, our method consistently maintains high model accuracy and watermark accuracy, demonstrating strong resilience to structural perturbations and parameter readjustments. Under pruning attacks, most baseline methods experience a sharp drop in watermark accuracy once the pruning rate exceeds 0.4, indicating severe degradation of embedded watermark features. In contrast, \texttt{FLClear} sustains stable watermark recognition even at 0.8 pruning rate. During fine-tuning, our approach achieves stable watermark accuracy performance while preserving model accuracy. Under quantization, although all methods suffer inevitable performance drops at extreme bit-length compression (e.g., 2-bit), \texttt{FLClear} retains competitive watermark fidelity and task accuracy.

It is noteworthy that ClearStamp attains high model accuracy across attacks but maintains consistently low and stable watermark accuracy (40\%). This occurs because ClearStamp was originally designed for centralized watermarking. When applied in the FL setting, distributed training and aggregation prevent the watermark from being effectively constructed, resulting in nearly constant watermark accuracy. This observation further indicates that \texttt{FLClear} is orthogonal to ClearStamp, although both employ a transposed model for watermark embedding. Please refer to Appendix \ref{sec:cs-wm} for the visualization of the watermarks and explanation.

\subsection{Capacity}
\label{sec:capacity}
It is challenging to accurately evaluate the watermark capacity of \texttt{FLClear} from a visual perspective. To enable a quantitative evaluation, we convert random bit strings into images via a dot-code representation~\cite{1057351}. The image is divided into square patches corresponding to the number of bits in the encoded string. For grayscale images, each patch encodes one bit, with black and white representing 0 and 1, respectively. Using color images increases capacity because each patch contains three independent RGB channels. Each RGB patch encodes 3 bits (1 bit per channel), which is three times the capacity of grayscale.

For datasets used in this paper, grayscale images such as those in the MNIST and Fashion-MNIST (with resolution $28 \times 28 \times 1$) can encode a maximum watermark capacity of 784 bits per image. For color datasets like CIFAR-10 and CIFAR-100 (with resolution $32 \times 32 \times 3$), the maximum capacity is 3072 bits per image. These results indicate that the watermark capacity is inherently determined not by \texttt{FLClear} itself but by the spatial dimensions of the input images used for main-task training. We also evaluate the bit error rate (BER) of extracted watermarks to quantify decoding reliability. BER is defined as the ratio of incorrect bits to the total number of bits. As shown in \Cref{tab:ber}, BER values remain consistently low (0.65\%-2.44\%), demonstrating the robustness and reliability of the embedded bit sequences.
 
\begin{table}[t]
\centering
\caption{BER Results}
\label{tab:ber}
\renewcommand{\arraystretch}{1.1}
\setlength{\tabcolsep}{6pt}
\resizebox{\columnwidth}{!}{
\begin{tabular}{ccccccc}
\toprule
\multirow{2}{*}{ } & \multirow{2}{*}{\textbf{MNIST}} & \multirow{2}{*}{\textbf{Fashion-MNIST}} & \textbf{CIFAR-10} & \textbf{CIFAR-10} & \textbf{CIFAR-100} & \textbf{CIFAR-100} \\
                         &                        &                                & \textbf{(grayscale)}   & \textbf{(color)}   & \textbf{(grayscale)}    & \textbf{(color)}   \\
\midrule
\textbf{BER (\%)} & 0.82 & 1.60 & 1.43 & 0.65 & 1.17 & 2.44 \\
\bottomrule
\end{tabular}
}
\end{table}

\subsection{Overhead}
\label{sec:overhead}

\begin{figure}[t]
    \centering
    \begin{subfigure}{0.48\columnwidth}
        \centering
        \includegraphics[width=\textwidth]{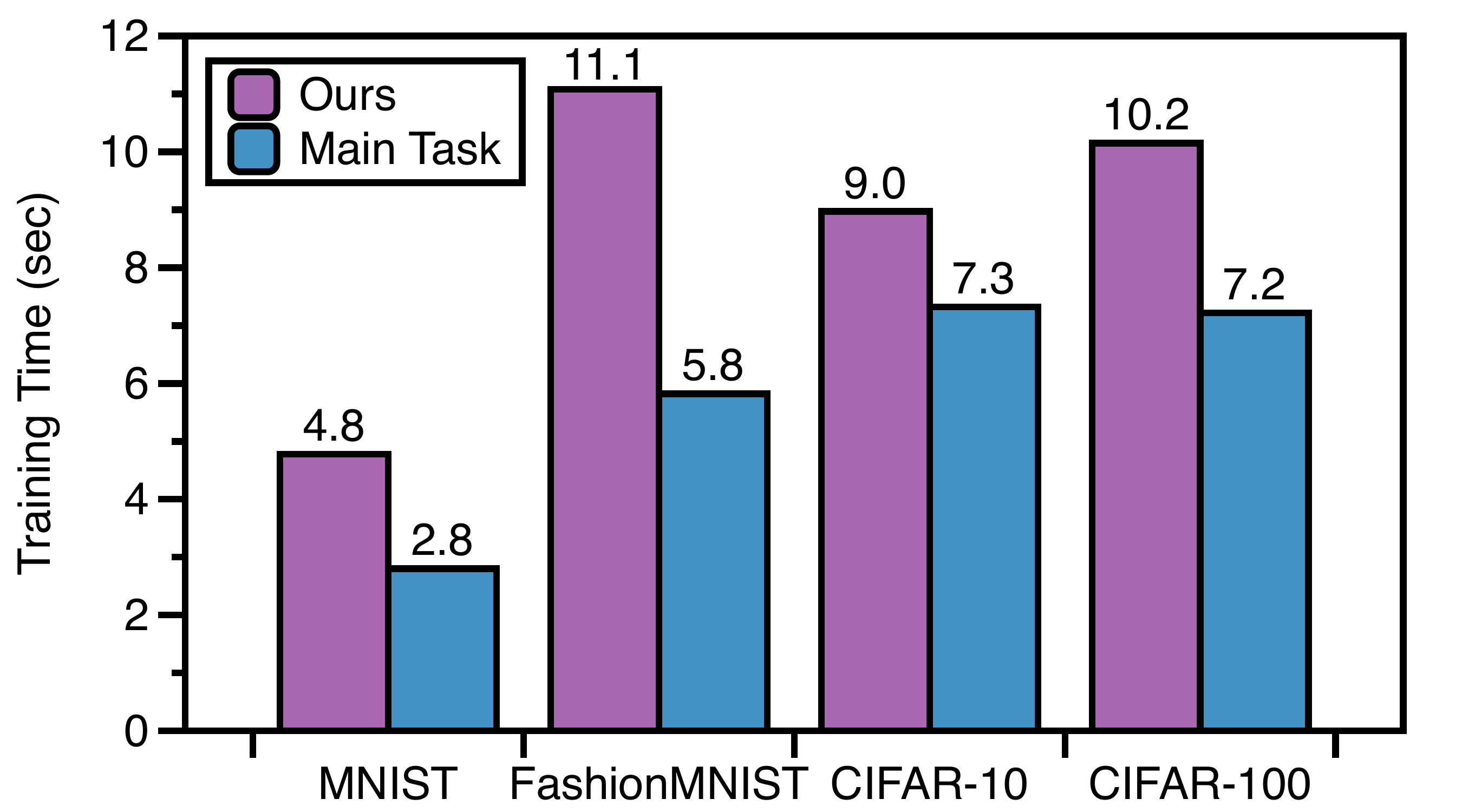}
        \caption{Time Cost}
        \label{fig:tc}
    \end{subfigure}
    \begin{subfigure}{0.48\columnwidth}
        \centering
        \includegraphics[width=\textwidth]{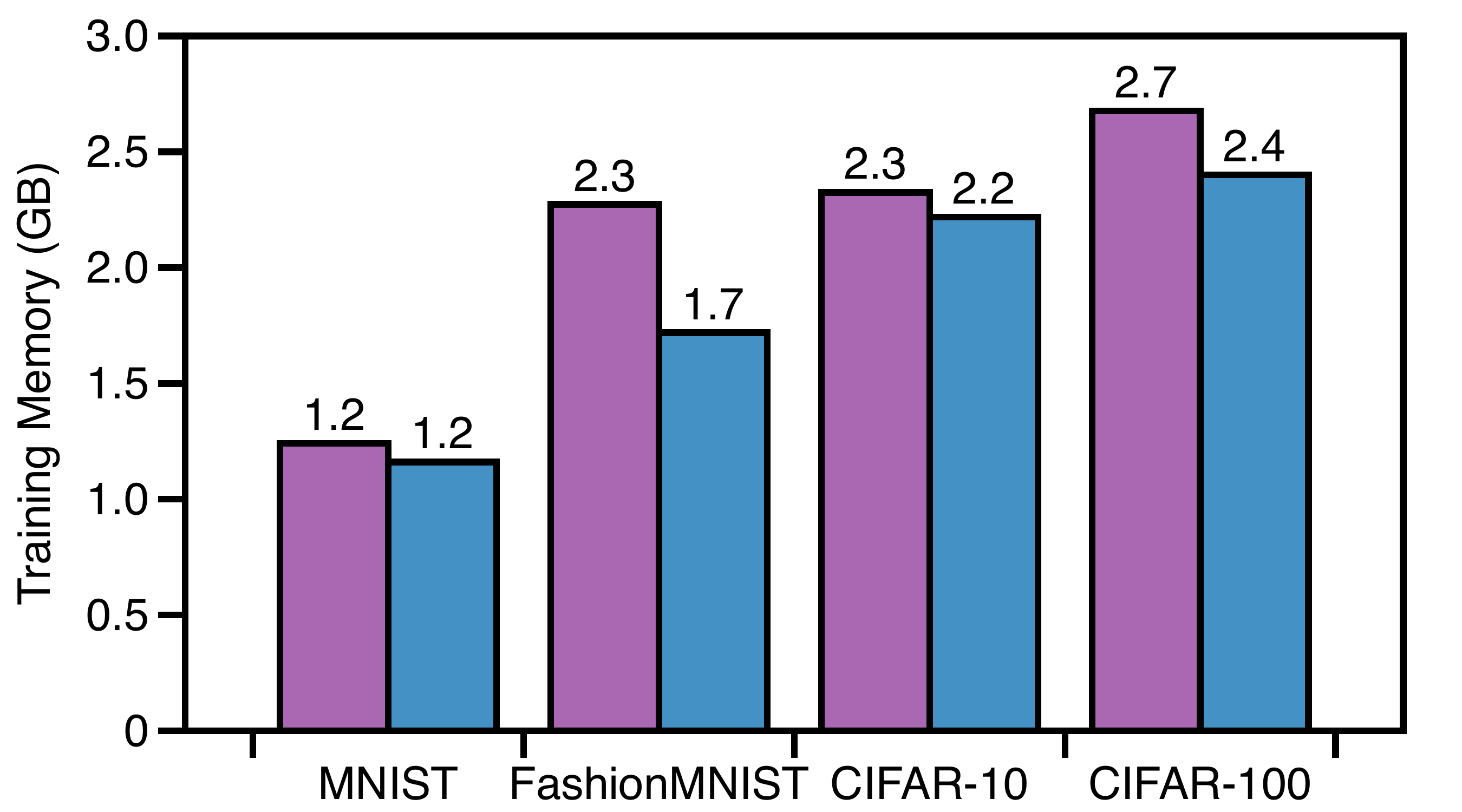}
        \caption{Memory Cost}
        \label{fig:mc}
    \end{subfigure}
    \caption{Overhead of each training round}
    \label{fig:overhead}
\end{figure}

Finally, we present the overhead of our framework in terms of training time and memory usage per round in \Cref{fig:overhead}. As shown in \Cref{fig:tc}, \texttt{FLClear} causes a moderate increase in training time compared with standalone main-task training, with increments of 2.0\,s, 5.3\,s, 1.7\,s, and 3.0\,s on MNIST, Fashion-MNIST, CIFAR-10, and CIFAR-100, respectively. This overhead primarily results from joint optimization of watermark and task objectives. Despite the increased computational complexity, the additional time cost remains within 1.5$\times$ of the baseline across all datasets.
Moreover, memory usage, as shown in \Cref{fig:mc}, differs minimally between our method and the baseline, remaining nearly identical on MNIST (1.2 GB) and within 0.1–0.6 GB across larger datasets. 

\section{Discussion}
\label{sec:discussion}

\noindent \textbf{Limitations.}
The first limitation of our work is the convergence speed. The transposed model training responsible for watermarking requires more training rounds to achieve convergence compared to the main-task model. This occurs because jointly optimizing the main and watermarking tasks is inherently more challenging than optimizing the main task alone. The joint optimization objective may introduce gradient conflicts and increase optimization complexity across tasks, thereby slowing convergence. Moreover, the selection of watermark images is constrained by the main-task training, as the transposed model reverses the input–output mapping of the main task. Consequently, the dimensions of the watermark image must match those of the input images used for training the main-task model. This implies that if high-resolution or large watermark images are employed, they will be automatically resized to align with the input size of the main task.

\noindent \textbf{Orthogonal Works.} Several prior works have inspired but remain orthogonal to \texttt{FLClear}. Our proposed framework builds upon the concept of the transposed model, originally motivated by deconvolutions \cite{8618415}, which approximate the reverse operation of convolutions. Drawing further inspiration from Siamese Networks \cite{koch2015siamese}, we adopt the principle of weight sharing as the foundation of transposed training. However, our method extends beyond existing transposed model designs by enabling watermark-aware learning within federated environments. 

The work most closely related to ours is ClearStamp~\cite{10.5555/3698900.3699195} (USENIX ’24), which embeds a human-visible watermark by training a transposed model that jointly learns the watermark and the main task. However, our work differs from ClearStamp in several key aspects: (i) \emph{implementation mechanism.} \texttt{FLClear} introduces a novel contrastive learning–based watermark embedding mechanism, whereas ClearStamp employs a watermark hardening strategy; (ii) \emph{design goal.} \texttt{FLClear} simultaneously achieves collision-free, secure, and visually verifiable watermarking, while ClearStamp addresses only part of these objectives; (iii) \emph{application scenario.} \texttt{FLClear} is specifically designed for FL environments involving model aggregation, whereas ClearStamp is a centralized watermarking approach. Our experimental results in \Cref{fig:comp-ma} demonstrate that directly applying ClearStamp in FL settings fails to construct valid watermark images, highlighting its incompatibility with decentralized training scenarios. 


\section{Related Work}
\label{sec:rw}

\begin{table}[t]
\centering
\caption{An overview of state-of-the-art multi-client FL watermarking schemes}
\label{tab:rw-overview}
\resizebox{\columnwidth}{!}{
\begin{threeparttable}
\renewcommand{\arraystretch}{1.15}
\setlength{\tabcolsep}{4pt}
{\large 
\begin{tabular}{ccccccc}
\toprule
\multicolumn{1}{c}{\multirow{2}{*}{\begin{tabular}[c]{@{}c@{}}Client-level\\ Watermarking\end{tabular}}} &
\multicolumn{1}{c}{\multirow{2}{*}{Method}} &
\multirow{2}{*}{\begin{tabular}[c]{@{}c@{}}Collision-free without\\ Restriction\end{tabular}} &
\multicolumn{2}{c}{Robustness \& Security} & \multicolumn{2}{c}{Verification} \\ 
\cmidrule(lr){4-5} \cmidrule(lr){6-7}
& & & Anti-Modification & Anti-Forgery & Qualitative & Quantitative \\
\midrule

\multirow{3}{*}{Weighted-based} 
 & \cite{10630980} & \CIRCLE & \CIRCLE & \CIRCLE & \Circle & \CIRCLE \\
 & \cite{liang2023fedcipfederatedclientintellectual} & \Circle & \CIRCLE & \Circle & \Circle & \CIRCLE \\
 & \cite{xu2024robwe} & \Circle & \CIRCLE & \Circle & \Circle & \CIRCLE \\
\midrule

\multirow{4}{*}{Trigger-based}
 & \cite{10.1016/j.eswa.2024.123776} & \Circle & \CIRCLE & \Circle & \Circle & \CIRCLE \\
 & \cite{li2025infightingdarkmultilabelbackdoor} & \CIRCLE & \CIRCLE & \Circle & \Circle & \CIRCLE \\
 & \cite{luo2025harmlessbackdoorbasedclientsidewatermarking} & \Circle & \CIRCLE & \Circle & \Circle & \CIRCLE \\
 & \cite{gai2025mfl} & \Circle & \CIRCLE & \Circle & \Circle & \CIRCLE \\
\midrule
\multirow{3}{*}{Hybrid-based}
 & \cite{9847383} & \Circle & \CIRCLE & \Circle & \Circle & \CIRCLE \\
 & \cite{10840113} & \Circle & \CIRCLE & \Circle & \Circle & \CIRCLE \\
 & \cite{10504977} & \Circle & \CIRCLE & \Circle & \Circle & \CIRCLE \\
 
\multicolumn{2}{c}{\textbf{\texttt{FLClear} (Ours)}} & \CIRCLE & \CIRCLE & \CIRCLE & \CIRCLE & \CIRCLE \\
\bottomrule
\end{tabular}
}
\begin{tablenotes}
\large
\item \CIRCLE: The property is supported; \Circle: The property is not supported.	
\end{tablenotes}
\end{threeparttable}
}
\end{table}

The FL watermarks can be employed to protect the IPR of the server, and individual clients, which are respectively referred to as server-side watermarks, and client-side watermarks. 

\emph{Server-side watermarks}~\cite{9603498, CHEN2023103504,10888286,10993411,cao2024fedsmw,LUO2024121161,11036162, yu2023leaked} are embedded by the central server after the model aggregation phase to protect the ownership of the global model. These watermarking approaches can be further categorized into two subtypes: weight-based and trigger-based. Weight-based server-side methods~\cite{10993411,LUO2024121161,11036162} embed ownership information directly into model parameters by inserting unique signatures or controlled perturbations into the weights. Trigger-based server-side watermarking~\cite{9603498,CHEN2023103504,10888286,cao2024fedsmw,yu2023leaked} relies on predefined input triggers or backdoor patterns that induce distinct model responses for ownership verification, typically without requiring access to model internals. For example, Tekgul \emph{et al.} \cite{9603498} employ data-agnostic trigger sets to retrain the model and embed black-box watermarks. 

\emph{Client-side watermarks}~\cite{10630980,10504977,10840113,10.1016/j.eswa.2024.123776,9847383,liang2023fedcipfederatedclientintellectual,10.1145/3630636,gai2025mfl,xu2024robwe,luo2025harmlessbackdoorbasedclientsidewatermarking,li2025infightingdarkmultilabelbackdoor} are embedded by one or more clients to uniquely mark their contributions to the collaboratively trained global model. Single-client watermarking is typically implemented by the model owner or FL initiator. For instance, Yang \emph{et al.} \cite{10.1145/3630636} embed backdoor trigger sets via gradient enhancement.
Multi-client watermarking \cite{10630980,10.1016/j.eswa.2024.123776,9847383,liang2023fedcipfederatedclientintellectual,gai2025mfl,xu2024robwe,luo2025harmlessbackdoorbasedclientsidewatermarking,li2025infightingdarkmultilabelbackdoor,10504977,10840113} allows multiple clients to independently embed distinctive watermarks into their local models, which are subsequently aggregated into the global model. 
Among these approaches, methods such as~\cite{10630980,liang2023fedcipfederatedclientintellectual,xu2024robwe} directly manipulate model parameters to encode ownership information and are therefore classified as weight-based client-side watermarking.  
In contrast, works such as \cite{10.1016/j.eswa.2024.123776,li2025infightingdarkmultilabelbackdoor,luo2025harmlessbackdoorbasedclientsidewatermarking,gai2025mfl} adopt backdoor triggers to embed verifiable behavioral patterns, representing trigger-based client-side watermarking. 
A third line of research \cite{10504977,10840113,9847383} integrate both parameter-level and trigger-based mechanisms to improve watermark robustness and flexibility, thus classified as hybrid-based client-side watermarking. For instance, Shao \emph{et al.}~\cite{10504977} propose a dual-layer mechanism that combines global watermarks with client-specific fingerprints.

We compare \texttt{FLClear} with state-of-the-art multi-client FL watermarking schemes in \Cref{tab:rw-overview}. The results show that our framework is the only one that simultaneously achieves all targeted functionalities.

\section{Conclusion}
\label{sec:conclusion}
This work addresses the challenges of FL watermarking by proposing \texttt{FLClear}, a framework that achieves collision-free, secure, and visually verifiable watermarking. \texttt{FLClear} employs a transposed model to embed watermarks into the network and jointly optimizes it with the main-task model through contrastive learning. Extensive experiments across diverse models, datasets, aggregation schemes, and attack scenarios demonstrate the scalability and effectiveness of \texttt{FLClear}.
Currently, our framework operates within an image-based DNN setting, where static images serve as the form of watermark information. In future work, we plan to extend this paradigm to videos by embedding watermarks in a visually perceptible and temporally coherent manner. This direction may introduces a more dynamic application scenario and presents new technical challenges, requiring the watermarking scheme to remain fidelity and robust under both spatial and temporal perturbations.

\section*{Acknowledgment}
This work was supported in part by Open Fund of Anhui Province Key Laboratory of Cyberspace Security Situation Awareness and Evaluation under Grant No. CSSAE-2023-005.

\bibliographystyle{IEEEtran}
\bibliography{ref}

@inproceedings{lukas2022sok,
  title={Sok: How robust is image classification deep neural network watermarking?},
  author={Lukas, Nils and Jiang, Edward and Li, Xinda and Kerschbaum, Florian},
  booktitle={2022 IEEE Symposium on Security and Privacy (SP)},
  pages={787--804},
  year={2022},
  //organization={IEEE}
}

@inproceedings{koch2015siamese,
  title={Siamese neural networks for one-shot image recognition},
  author={Koch, Gregory and Zemel, Richard and Salakhutdinov, Ruslan and others},
  booktitle={ICML deep learning workshop},
  volume={2},
  number={1},
  pages={1--30},
  year={2015},
  //organization={Lille}
}

@article{8618415,
  title={Pixel transposed convolutional networks},
  author={Gao, Hongyang and Yuan, Hao and Wang, Zhengyang and Ji, Shuiwang},
  journal={IEEE transactions on pattern analysis and machine intelligence},
  volume={42},
  number={5},
  pages={1218--1227},
  year={2019},
  //publisher={IEEE}
}

@inproceedings{sandler2018mobilenetv2,
  title={Mobilenetv2: Inverted residuals and linear bottlenecks},
  author={Sandler, Mark and Howard, Andrew and Zhu, Menglong and Zhmoginov, Andrey and Chen, Liang-Chieh},
  booktitle={Proceedings of the IEEE conference on computer vision and pattern recognition},
  pages={4510--4520},
  year={2018}
}

@article{xiao2017fashion,
  title={Fashion-mnist: a novel image dataset for benchmarking machine learning algorithms},
  author={Xiao, Han and Rasul, Kashif and Vollgraf, Roland},
  journal={arXiv preprint arXiv:1708.07747},
  year={2017}
}

@inproceedings{hu2024learn,
  title={Learn what you want to unlearn: Unlearning inversion attacks against machine unlearning},
  author={Hu, Hongsheng and Wang, Shuo and Dong, Tian and Xue, Minhui},
  booktitle={2024 IEEE Symposium on Security and Privacy (SP)},
  pages={3257--3275},
  year={2024},
  organization={IEEE}
}

@misc{LeCun:1998:MnistDatabaseHandwritten,
  title = {The Mnist Database of Handwritten Digits},
  author = {LeCun, Yann},
  year = {1998}
}

@article{hsu2019measuring,
  title={Measuring the effects of non-identical data distribution for federated visual classification},
  author={Hsu, Tzu-Ming Harry and Qi, Hang and Brown, Matthew},
  journal={arXiv preprint arXiv:1909.06335},
  year={2019}
}

@article{10.1145/3298981,
  title={Federated machine learning: Concept and applications},
  author={Yang, Qiang and Liu, Yang and Chen, Tianjian and Tong, Yongxin},
  journal={ACM Transactions on Intelligent Systems and Technology (TIST)},
  volume={10},
  number={2},
  pages={1--19},
  year={2019}
}

@inproceedings{10.1145/3078971.3078974,
  title={Embedding watermarks into deep neural networks},
  author={Uchida, Yusuke and Nagai, Yuki and Sakazawa, Shigeyuki and Satoh, Shin'ichi},
  booktitle={Proceedings of the 2017 ACM on international conference on multimedia retrieval},
  pages={269--277},
  year={2017}
}

@inproceedings{10.5555/3277203.3277324,
  title={Turning your weakness into a strength: Watermarking deep neural networks by backdooring},
  author={Adi, Yossi and Baum, Carsten and Cisse, Moustapha and Pinkas, Benny and Keshet, Joseph},
  booktitle={27th USENIX security symposium (USENIX Security 18)},
  pages={1615--1631},
  year={2018}
}

@article{2021A,
  title={A survey of deep neural network watermarking techniques},
  author={Li, Yue and Wang, Hongxia and Barni, Mauro},
  journal={Neurocomputing},
  volume={461},
  pages={171--193},
  year={2021},
  //publisher={Elsevier}
}

@inproceedings{9603498,
  title={Waffle: Watermarking in federated learning},
  author={Tekgul, Buse GA and Xia, Yuxi and Marchal, Samuel and Asokan, N},
  booktitle={40th International Symposium on Reliable Distributed Systems (SRDS)},
  pages={310--320},
  year={2021},
  //organization={IEEE}
}

@inproceedings{10888286,
  title={Collusion-resistant Black-box Watermarking in Federated Learning through Weight Relevance Analysis},
  author={Rodr{\'\i}guez-Lois, Elena and P{\'e}rez-Gonz{\'a}lez, Fernando},
  booktitle={ICASSP 2025-2025 IEEE International Conference on Acoustics, Speech and Signal Processing (ICASSP)},
  pages={1--5},
  year={2025},
  //organization={IEEE}
}

@article{kong2024asia,
  title={ASIA: A federated boosting tree model against sequence inference attacks in financial networks},
  author={Kong, Yubo and Li, Zhong and Jiang, Changjun},
  journal={IEEE Transactions on Information Forensics and Security},
  year={2024},
  //publisher={IEEE}
}

@article{CHEN2023103504,
  title={Fedright: An effective model copyright protection for federated learning},
  author={Chen, Jinyin and Li, Mingjun and Cheng, Yao and Zheng, Haibin},
  journal={Computers \& Security},
  volume={135},
  pages={103504},
  year={2023},
  //publisher={Elsevier}
}

@article{10288131,
  title={Federated learning for healthcare applications},
  author={Chaddad, Ahmad and Wu, Yihang and Desrosiers, Christian},
  journal={IEEE internet of things journal},
  volume={11},
  number={5},
  pages={7339--7358},
  year={2023},
  //publisher={IEEE}
}

@article{taik2022clustered,
  title={Clustered vehicular federated learning: Process and optimization},
  author={Taik, Afaf and Mlika, Zoubeir and Cherkaoui, Soumaya},
  journal={IEEE Transactions on Intelligent Transportation Systems},
  volume={23},
  number={12},
  pages={25371--25383},
  year={2022},
  //publisher={IEEE}
}

@article{LI2020106854,
  title={A review of applications in federated learning},
  author={Li, Li and Fan, Yuxi and Tse, Mike and Lin, Kuo-Yi},
  journal={Computers \& Industrial Engineering},
  volume={149},
  pages={106854},
  year={2020},
  //publisher={Elsevier}
}

@article{10504977,
  title={Fedtracker: Furnishing ownership verification and traceability for federated learning model},
  author={Shao, Shuo and Yang, Wenyuan and Gu, Hanlin and Qin, Zhan and Fan, Lixin and Yang, Qiang and Ren, Kui},
  journal={IEEE Transactions on Dependable and Secure Computing},
  volume={22},
  number={1},
  pages={114--131},
  year={2024},
 //publisher={IEEE}
}

@INPROCEEDINGS{10840113,
  author={Rodríguez-Lois, Elena and Pérez-González, Fernando},
  booktitle={2024 2nd International Conference on Federated Learning Technologies and Applications (FLTA)}, 
  title={Exploring Federated Learning Dynamics for Black-and-White-Box DNN Traitor Tracing}, 
  year={2024},
  pages={282-289},
  }

@article{liang2023fedcipfederatedclientintellectual,
  title={Fedcip: Federated client intellectual property protection with traitor tracking},
  author={Liang, Junchuan and Wang, Rong},
  journal={arXiv preprint arXiv:2306.01356},
  year={2023}
}

@inproceedings{10630980,
  title={Fedmark: Large-capacity and robust watermarking in federated learning},
  author={Zhang, Lan and Tang, Chen and Liu, Huiqi and Yu, Haikuo and Zhuang, Xirong and Zhao, Qi and Wang, Lei and Fang, Wenjing and Li, Xiang-Yang},
  booktitle={2024 IEEE 44th International Conference on Distributed Computing Systems (ICDCS)},
  pages={821--832},
  year={2024},
  //organization={IEEE}
}

@article{9847383,
  title={FedIPR: Ownership verification for federated deep neural network models},
  author={Li, Bowen and Fan, Lixin and Gu, Hanlin and Li, Jie and Yang, Qiang},
  journal={IEEE Transactions on Pattern Analysis and Machine Intelligence},
  volume={45},
  number={4},
  pages={4521--4536},
  year={2022},
  //publisher={IEEE}
}

@inproceedings{li2025infightingdarkmultilabelbackdoor,
  title={Infighting in the Dark: Multi-Label Backdoor Attack in Federated Learning},
  author={Li, Ye and Zhao, Yanchao and Zhu, Chengcheng and Zhang, Jiale},
  booktitle={Proceedings of the Computer Vision and Pattern Recognition Conference},
  pages={25770--25779},
  year={2025}
}

@inproceedings{luo2025harmlessbackdoorbasedclientsidewatermarking,
  title={Harmless backdoor-based client-side watermarking in federated learning},
  author={Luo, Kaijing and Chow, Ka-Ho},
  booktitle={2025 IEEE 10th European Symposium on Security and Privacy (EuroS\&P)},
  pages={1002--1020},
  year={2025},
  //organization={IEEE}
}

@article{10.1016/j.eswa.2024.123776,
  title={Fedcrmw: Federated model ownership verification with compression-resistant model watermarking},
  author={Nie, Hewang and Lu, Songfeng},
  journal={Expert Systems with Applications},
  volume={249},
  pages={123776},
  year={2024},
  //publisher={Elsevier}
}

@inproceedings{10.5555/3698900.3699195,
author = {Krau\ss{}, Torsten and Stang, Jasper and Dmitrienko, Alexandra},
title = {ClearStamp: a human-visible and robust model-ownership proof based on transposed model training},
booktitle = {Proceedings of the 33rd USENIX Conference on Security Symposium},
year = {2024},
}

@INPROCEEDINGS{5539957,
  author={Zeiler, Matthew D. and Krishnan, Dilip and Taylor, Graham W. and Fergus, Rob},
  booktitle={2010 IEEE Computer Society Conference on Computer Vision and Pattern Recognition}, 
  title={Deconvolutional networks}, 
  year={2010},
  pages={2528-2535},
}

@article{10.1145/3630636,
  title={Watermarking in secure federated learning: A verification framework based on client-side backdooring},
  author={Yang, Wenyuan and Shao, Shuo and Yang, Yue and Liu, Xiyao and Liu, Ximeng and Xia, Zhihua and Schaefer, Gerald and Fang, Hui},
  journal={ACM Transactions on Intelligent Systems and Technology},
  volume={15},
  number={1},
  pages={1--25},
  year={2023},
  //publisher={ACM New York, NY}
}

@article{10993411,
  title={FedMLC: White-box Model Watermarking for Copyright Protection in Federated Learning for IoT Environment},
  author={Chen, Weitong and Zhang, Wei and Wu, Di and Keskinarkaus, Anja and Sepp{\"a}nen, Tapio and Zhang, Jiale and Gao, Longxiang and Luan, Tom H},
  journal={IEEE Internet of Things Journal},
  year={2025},
  //publisher={IEEE}
}

@article{11036162,
  title={Robust Watermarking for Federated Diffusion Models with Unlearning-Enhanced Redundancy},
  author={Pan, Zijie and Ying, Zuobin and Wang, Yajie and Wang, Yani and Zhang, Zijian and Zhou, Wanlei and Zhu, Liehuang},
  journal={IEEE Transactions on Dependable and Secure Computing},
  year={2025},
  //publisher={IEEE}
}

@article{LUO2024121161,
  title={Copyright protection framework for federated learning models against collusion attacks},
  author={Luo, Yuling and Li, Yuanze and Qin, Sheng and Fu, Qiang and Liu, Junxiu},
  journal={Information Sciences},
  volume={680},
  pages={121161},
  year={2024},
  //publisher={Elsevier}
}

@inproceedings{cao2024fedsmw,
  title={FedSMW: Server-Side Model Watermark Framework for Model Ownership Verification in Federated Learning},
  author={Cao, Yang and Fang, Hao and Chen, Bin and Wang, Xuan and Xia, Shu-Tao},
  booktitle={2024 16th International Conference on Wireless Communications and Signal Processing (WCSP)},
  pages={18--23},
  year={2024},

}

@inproceedings{gai2025mfl,
  title={MFL-owner: ownership protection for multi-modal federated learning via orthogonal transform watermark},
  author={Gai, Keke and Wang, Dongjue and Yu, Jing and Wang, Mohan and Zhu, Liehuang and Wu, Qi},
  booktitle={Proceedings of the AAAI Conference on Artificial Intelligence},
  volume={39},
  number={3},
  pages={3049--3058},
  year={2025}
}

@article{xu2024robwe,
  title={Robwe: Robust watermark embedding for personalized federated learning model ownership protection},
  author={Xu, Yang and Tan, Yunlin and Zhang, Cheng and Chi, Kai and Sun, Peng and Yang, Wenyuan and Ren, Ju and Jiang, Hongbo and Zhang, Yaoxue},
  journal={arXiv preprint arXiv:2402.19054},
  year={2024}
}

@article{yu2023leaked,
  title={Who leaked the model? tracking ip infringers in accountable federated learning},
  author={Yu, Shuyang and Hong, Junyuan and Zeng, Yi and Wang, Fei and Jia, Ruoxi and Zhou, Jiayu},
  journal={arXiv preprint arXiv:2312.03205},
  year={2023}
}

@INPROCEEDINGS{8695386,
  author={Sakazawa, Shigeyuki and Myodo, Emi and Tasaka, Kazuyuki and Yanagihara, Hiromasa},
  booktitle={2019 IEEE Conference on Multimedia Information Processing and Retrieval (MIPR)}, 
  title={Visual Decoding of Hidden Watermark in Trained Deep Neural Network}, 
  year={2019},
  pages={371-374},
}

@article{khosla2020supervised,
  title={Supervised contrastive learning},
  author={Khosla, Prannay and Teterwak, Piotr and Wang, Chen and Sarna, Aaron and Tian, Yonglong and Isola, Phillip and Maschinot, Aaron and Liu, Ce and Krishnan, Dilip},
  journal={Advances in neural information processing systems},
  volume={33},
  pages={18661--18673},
  year={2020}
}

@article{zhang2021survey,
  title={A survey on multi-task learning},
  author={Zhang, Yu and Yang, Qiang},
  journal={IEEE transactions on knowledge and data engineering},
  volume={34},
  number={12},
  pages={5586--5609},
  year={2021},
  //publisher={IEEE}
}

@inproceedings{wang2019neural,
  title={Neural cleanse: Identifying and mitigating backdoor attacks in neural networks},
  author={Wang, Bolun and Yao, Yuanshun and Shan, Shawn and Li, Huiying and Viswanath, Bimal and Zheng, Haitao and Zhao, Ben Y},
  booktitle={2019 IEEE symposium on security and privacy (SP)},
  pages={707--723},
  year={2019},
}

@inproceedings{qiu2024belt,
  title={BELT: Old-School Backdoor Attacks can Evade the State-of-the-Art Defense with Backdoor Exclusivity Lifting},
  author={Qiu, Huming and Sun, Junjie and Zhang, Mi and Pan, Xudong and Yang, Min},
  booktitle={2024 IEEE Symposium on Security and Privacy (SP)},
  pages={2124--2141},
  year={2024},
}

@inproceedings{karimireddy2020scaffold,
  title={Scaffold: Stochastic controlled averaging for federated learning},
  author={Karimireddy, Sai Praneeth and Kale, Satyen and Mohri, Mehryar and Reddi, Sashank and Stich, Sebastian and Suresh, Ananda Theertha},
  booktitle={International conference on machine learning},
  pages={5132--5143},
  year={2020},
}

@article{li2020federated,
  title={Federated optimization in heterogeneous networks},
  author={Li, Tian and Sahu, Anit Kumar and Zaheer, Manzil and Sanjabi, Maziar and Talwalkar, Ameet and Smith, Virginia},
  journal={Proceedings of Machine learning and systems},
  volume={2},
  pages={429--450},
  year={2020}
}

@inproceedings{tartaglione2021delving,
  title={Delving in the loss landscape to embed robust watermarks into neural networks},
  author={Tartaglione, Enzo and Grangetto, Marco and Cavagnino, Davide and Botta, Marco},
  booktitle={2020 25th International Conference on Pattern Recognition (ICPR)},
  pages={1243--1250},
  year={2021},
  //organization={IEEE}
}

@inproceedings{reisizadeh2020fedpaq,
  title={Fedpaq: A communication-efficient federated learning method with periodic averaging and quantization},
  author={Reisizadeh, Amirhossein and Mokhtari, Aryan and Hassani, Hamed and Jadbabaie, Ali and Pedarsani, Ramtin},
  booktitle={International conference on artificial intelligence and statistics},
  pages={2021--2031},
  year={2020},
}

@inproceedings{mcmahan2017communication,
  title={Communication-efficient learning of deep networks from decentralized data},
  author={McMahan, Brendan and Moore, Eider and Ramage, Daniel and Hampson, Seth and y Arcas, Blaise Aguera},
  booktitle={Artificial intelligence and statistics},
  pages={1273--1282},
  year={2017},
  //organization={PMLR}
}

@article{krizhevsky2012imagenet,
  title={Imagenet classification with deep convolutional neural networks},
  author={Krizhevsky, Alex and Sutskever, Ilya and Hinton, Geoffrey E},
  journal={Advances in neural information processing systems},
  volume={25},
  year={2012}
}

@article{simonyan2014very,
  title={Very deep convolutional networks for large-scale image recognition},
  author={Simonyan, Karen and Zisserman, Andrew},
  journal={arXiv preprint arXiv:1409.1556},
  year={2014}
}

@inproceedings{he2016deep,
  title={Deep residual learning for image recognition},
  author={He, Kaiming and Zhang, Xiangyu and Ren, Shaoqing and Sun, Jian},
  booktitle={Proceedings of the IEEE conference on computer vision and pattern recognition},
  pages={770--778},
  year={2016}
}

@article{krizhevsky2009learning,
  title={Learning multiple layers of features from tiny images},
  author={Krizhevsky, Alex and Hinton, Geoffrey and others},
  year={2009},
  //publisher={Toronto, ON, Canada}
}

@article{yao2025hashed,
  title={Hashed Watermark as a Filter: Defeating Forging and Overwriting Attacks in Weight-based Neural Network Watermarking},
  author={Yao, Yuan and Song, Jin and Jin, Jian},
  journal={arXiv preprint arXiv:2507.11137},
  year={2025}
}

@article{li2021fedbn,
  title={Fedbn: Federated learning on non-iid features via local batch normalization},
  author={Li, Xiaoxiao and Jiang, Meirui and Zhang, Xiaofei and Kamp, Michael and Dou, Qi},
  journal={arXiv preprint arXiv:2102.07623},
  year={2021}
}

@inproceedings{DBLP:conf/iclr/ReddiCZGRKKM21,
  author       = {Sashank J. Reddi and
                  Zachary Charles and
                  Manzil Zaheer and
                  Zachary Garrett and
                  Keith Rush and
                  Jakub Kone{\v{c}}n{\'y} and
                  Sanjiv Kumar and
                  Hugh Brendan McMahan},
  title        = {Adaptive Federated Optimization},
  booktitle    = {9th International Conference on Learning Representations,},
  year         = {2021},
}

@inproceedings{ioffe2015batch,
  title={Batch normalization: Accelerating deep network training by     reducing internal covariate shift},
  author={Ioffe, Sergey and Szegedy, Christian},
  booktitle={International conference on machine learning},
  pages={448-456},
  year={2015},
}

@article{han2015learning,
  title={Learning both weights and connections for efficient neural     network},
  author={Han, Song and Pool, Jeff and Tran, John and Dally, William},
  journal={Advances in neural information processing systems},
  volume={28},
  year={2015},
}

@article{fan2019rethinking,
  title={Rethinking deep neural network ownership verification: Embedding passports to defeat ambiguity attacks},
  author={Fan, Lixin and Ng, Kam Woh and Chan, Chee Seng},
  journal={Advances in neural information processing systems},
  volume={32},
  year={2019}
}

@article{yang2024fedgmark,
  title={Fedgmark: Certifiably robust watermarking for federated graph learning},
  author={Yang, Yuxin and Li, Qiang and Hong, Yuan and Wang, Binghui},
  journal={Advances in Neural Information Processing Systems},
  volume={37},
  pages={48971--48995},
  year={2024}
}

@article{pang2025modelshield,
  title={Modelshield: Adaptive and robust watermark against model extraction attack},
  author={Pang, Kaiyi and Qi, Tao and Wu, Chuhan and Bai, Minhao and Jiang, Minghu and Huang, Yongfeng},
  journal={IEEE Transactions on Information Forensics and Security},
  year={2025},
  //publisher={IEEE}
}

@article{1057351,
  title={Two-dimensional dot codes for product identification},
  author={Van Gils, W},
  journal={IEEE transactions on information theory},
  volume={33},
  number={5},
  pages={620--631},
  year={1987},
  //publisher={IEEE}
}

\appendices

\section{Additional Details on Experimental Setup}
All experiments were conducted on a server equipped with an Intel(R) Xeon(R) Gold 6330 CPU @ 2.00 GHz (28 cores per socket, 2 threads per core) and an NVIDIA GeForce RTX 4090 GPU. Each parameter in the vector is randomly sampled from range $[-1, 1]$. The similarity threshold to separate positive and negative vectors was set to 0.95, where vectors with similarity below this threshold were treated as negative vectors and those exceeding it as positive vectors. This high threshold enhances robustness against forgery attacks by ensuring that only highly correlated samples are considered valid.

\subsection{Datasets and Models}
\noindent \textbf{Datasets.}
The MNIST~\cite{LeCun:1998:MnistDatabaseHandwritten} dataset includes 60,000 28 $\times$ 28 grayscale handwritten digit images (0-9, totaling 10 classes) for training and 10,000 images for testing. Fashion-MNIST~\cite{xiao2017fashion} dataset consists of 70,000 grayscale images of fashion items from 10 categories (such as T-shirts, trousers, sneakers, bags, etc.). The CIFAR-10~\cite{krizhevsky2009learning} dataset contains 50,000 32×32 color images for training and 10,000 images for testing, covering 10 classes. The CIFAR-100~\cite{krizhevsky2009learning} dataset comprises 50,000 32×32 color images for training and 10,000 images for testing, encompassing 100 fine-grained classes.

\noindent \textbf{Non-IID data distribution.}
We adopt the Non-IID data partitioning scheme described in \cite{hsu2019measuring}, where data heterogeneity among clients is simulated using a Dirichlet distribution with concentration parameter $\alpha$. Smaller values of $\alpha$ produce more skewed distributions, causing each client’s data to concentrate in fewer classes and thereby increasing heterogeneity. Conversely, larger $\alpha$ values yield a more uniform class distribution across clients, approximating an IID setting. In our experiments, we set $\alpha = 0.8$ to introduce moderate heterogeneity, providing a realistic simulation of Non-IID conditions in FL environments.

\noindent \textbf{DNN model structures.}
\Cref{tab:model_structures} summarizes the network architectures employed for each dataset to ensure fair comparisons across varying data complexities. All models incorporate batch normalization and ReLU activation for training stability, max pooling for spatial downsampling, and fully connected layers for classification. By employing standardized architectures, we ensure that observed performance variations arise from the proposed watermarking and learning strategies rather than differences in model capacity, thereby reinforcing the validity of cross-dataset evaluations.

\begin{table}[t]
    \centering
    \caption{Model structures for various datasets}
    \label{tab:model_structures}
    \resizebox{\linewidth}{!}{
    \begin{tabular}{ccc}
    \toprule
         \textbf{Model} &\textbf{Layer}  &\textbf{Parameter} \\ \midrule
         \multirow{7}{*}{AlexNet (on MNIST)} & Conv + BN + ReLU & $3 \times 3 \times 64$ \\
                             & Max Pooling & $2 \times 2$ \\
                             & Conv + BN + ReLU & $3 \times 3 \times 192$ \\
                             & Max Pooling & $2 \times 2$ \\
                             & \textbf{$3 \times$} Conv + BN + ReLU  & $3 \times 3 \times 384$, $3 \times 3 \times 384$, $3 \times 3 \times 256$ \\
                             & \textbf{$2 \times$} Fully Connected & $4096$, $512$, $128$, $10$\\ \midrule
         \multirow{10}{*}{MobileNetV2 (on Fashion-MNIST)} & Conv + BN + ReLU & $3 \times 3 \times 32$ (s=1) \\
                             & Inverted Residual Block (IRB) & $t=1, c=16, n=1, s=1$ \\
                             & IRB Group 2 ($n=2$) & $t=6, c=24, s=2$ \\
                             & IRB Group 3 ($n=3$) & $t=6, c=32, s=1$ \\
                             & IRB Group 4 ($n=4$) & $t=6, c=64, s=2$ \\
                             & IRB Group 5 ($n=3$) & $t=6, c=96, s=1$ \\
                             & IRB Group 6 ($n=3$) & $t=6, c=160, s=2$ \\
                             & IRB Group 7 ($n=1$) & $t=6, c=320, s=1$ \\
                             & Fully Connected & $5120$, $512$, $10$\\ \midrule
        \multirow{9}{*}{VGG13 (on CIFAR-10)} & \textbf{$2 \times$} (Conv + BN + ReLU) & $3 \times 3 \times 64$ \\
                             & Max Pooling & $2 \times 2$ \\
                             & \textbf{$2 \times$} (Conv + BN + ReLU) & $3 \times 3 \times 128$ \\
                             & Max Pooling & $2 \times 2$ \\
                             & \textbf{$2 \times$} (Conv + BN + ReLU) & $3 \times 3 \times 256$ \\
                             & \textbf{$4 \times$} (Conv + BN + ReLU) & $3 \times 3 \times 512$ \\
                             & Max Pooling & $2 \times 2$ \\
                             & \textbf{$3 \times$} Fully Connected & $8192$, $1028$, $512$, $10$ \\ \midrule
         \multirow{7}{*}{ResNet18 (on CIFAR-100)} & Conv + BN + ReLU  & $3 \times 3 \times 64$ (s=1) \\
                             & Layer 1 ($\mathbf{2 \times}$ Basic Block) & $c=64$, s=1, no Shortcut Conv \\
                             & Layer 2 ($\mathbf{2 \times}$ Basic Block) & $c=128$, s=2, $1 \times 1$ Shortcut Conv \\
                             & Layer 3 ($\mathbf{2 \times}$ Basic Block) & $c=256$, s=2, $1 \times 1$ Shortcut Conv \\
                             & Layer 4 ($\mathbf{2 \times}$ Basic Block) & $c=512$, s=2, $1 \times 1$ Shortcut Conv \\
                             & \textbf{$3 \times$} Fully Connected & $8192$, $1028$, $512$, $100$\\
                             
    \bottomrule
    \end{tabular}}
\end{table}

\subsection{Attack Methods}
We describe the attack methods and parameter settings used in our experiments below.
\begin{itemize}[left=0pt]
    \item Pruning: The model is compressed by removing weights with lower contributions to predictions. We tested different pruning ratios including 20\%, 40\%, 60\%, and 80\% to assess watermark robustness after compression.
    \item Fine-tuning: The watermark model is additionally trained on an independent holdout dataset to simulate post-deployment model reuse or update scenarios. We fine-tune the model for up to 30 training rounds with a learning rate of 0.001. We set the maximum tuning round is 30.
    \item Quantization: We converted the model's weights from high-precision floating-point representations to different low-precision formats including Int2, Int4, Int8, and Int16.
    \item Overwriting: The adversary applies a random vector and its own watermark to tune the transposed model following the \texttt{FLClear} training procedure, optimizing the vector accordingly. The maximum training round is set to 100.
    \item Forgery Attacks: We implemented the untargeted forgery attack using random unrelated watermark images as targets and the targeted forgery attack using the true watermark image (i.e., the client watermark image) as the target, generating forged trigger vectors through iterative optimization.
\end{itemize}

\subsection{Baselines}
\noindent \textbf{Aggregation Schemes.}
We introduce the aggregation schemes employed in our paper. These methods are designed to tackle challenges including data heterogeneity, communication efficiency, client resource constraints, and client drift in FL.
\begin{itemize}[left=0pt]
	\item FedAvg~\cite{mcmahan2017communication} is the foundational aggregation algorithm in FL, where the server computes a weighted average of the local models to update the global model. This approach significantly reduces communication costs compared to synchronous gradient descent.
	\item FedProx~\cite{li2020federated} extends FedAvg by introducing a proximal term in the local optimization objective, which constrains the local model updates to remain close to the global model. This modification alleviates the divergence problem caused by data heterogeneity across clients.
	\item FedPAQ~\cite{reisizadeh2020fedpaq} enhances the communication efficiency of FL by combining periodic averaging and quantization of model updates. Instead of sending updates after every local iteration, clients communicate less frequently, and their model updates are quantized to reduce transmission overhead.
	\item FedADAM~\cite{DBLP:conf/iclr/ReddiCZGRKKM21} is an adaptive optimization algorithm that extends the Adam optimizer to the FL setting. It maintains first- and second-moment estimates of client updates to adaptively adjust learning rates during global aggregation. 
	\item SCAFFOLD~\cite{karimireddy2020scaffold} addresses client drift by introducing control variates that correct for gradient bias in local updates. Each client maintains a control variable that captures the difference between local and global gradients, which is used to adjust local optimization.
\end{itemize}

\noindent \textbf{The State-of-the-art Schemes.}
We describe watermarking approaches used for comparison in this paper. These approaches cover both parameter-based and trigger-set-based watermarking schemes.
\begin{itemize}[left=0pt]
    \item WAFFLE~\cite{9603498} is a trigger-set-based watermarking scheme that introduces a post-aggregation retraining phase at the server to encode watermarks into the model's input-output mapping, allowing ownership verification solely through black-box API queries.
	\item \cite{10888286} is also a trigger-set-based watermarking scheme that assigns trigger sets to each data owner and employs weight correlation analysis to enhance the robustness of black-box watermarks.
    \item FedTracker~\cite{10504977} is a hybrid watermarking scheme that establishes global model ownership via a shared trigger-set watermark while embedding unique identifier watermarks in each client's model copy.
	\item FedIPR~\cite{9847383} is a hybrid watermarking scheme that synchronously embeds private feature-based watermarks and trigger-set backdoors into each client's local model, enabling dual-mode white-box/black-box ownership verification.
	\item ClearStamp~\cite{10.5555/3698900.3699195} is a visual watermarking method for centralized training that employs a transposed model training architecture to diffusely embed human-perceptible watermark information.
\end{itemize}

\subsection{Visual Evaluation Metrics}
We briefly describe the four visual evaluation metrics in the paper.
\begin{itemize}[left=0pt]
	\item Mean Squared Error (MSE)  measures the average squared difference between the original image $I$ and the reconstructed image $\hat{I}$. It quantifies pixel-wise similarity, where a lower value indicates higher fidelity. 
		\begin{equation}
			\text{MSE} = \frac{1}{N} \sum_{i=1}^{N} (I_i - \hat{I}_i)^2.
		\end{equation}
		
	\item Structural Similarity Index (SSIM) evaluates perceptual image quality by comparing luminance, contrast, and structural information between $I$ and $\hat{I}$. Its value ranges from $-1$ to $1$, where a higher value indicates greater similarity. 
		\begin{equation}
			\text{SSIM}(I, \hat{I}) = \frac{(2\mu_I \mu_{\hat{I}} + C_1)(2\sigma_{I\hat{I}} + C_2)}{(\mu_I^2 + \mu_{\hat{I}}^2 + C_1)(\sigma_I^2 + \sigma_{\hat{I}}^2 + C_2)}.
		\end{equation}
	\item Peak Signal-to-Noise Ratio (PSNR) expresses the ratio between the maximum possible power of a signal and the power of noise affecting its representation. It is derived from MSE, with higher PSNR values indicating better image quality. 
		\begin{equation}
		\text{PSNR} = 10 \log_{10} \left( \frac{L^2}{\text{MSE}} \right),
		\end{equation}
		where $L$ is the maximum possible pixel value (e.g., 255 for 8-bit images).
	\item Learned Perceptual Image Patch Similarity (LPIPS) \cite{hu2024learn} measures perceptual similarity using deep network features rather than raw pixels. It computes the distance between image features extracted from pretrained networks (e.g., VGG), where lower values denote more perceptually similar images. 
		\begin{equation}
		\text{LPIPS}(I, \hat{I}) = \sum_l \frac{1}{H_l W_l} \sum_{h,w} \| w_l \odot (\phi_l(I)_{hw} - \phi_l(\hat{I})_{hw}) \|_2^2,
		\end{equation}
		where $\phi_l(\cdot)$ denotes the feature map at layer $l$ and $w_l$ are learned weights.
\end{itemize}

\section{Additional Experimental Results}

\begin{table*}[htbp]
\centering
\caption{Comprehensive ASR results under forgery attacks}
\label{tab:asr-more}
\resizebox{\textwidth}{!}{
\begin{tabular}{cccccccccccccccccc}
\toprule
\multirow{3}{*}{\textbf{Dataset}} &
\multirow{3}{*}{\shortstack{\textbf{Attack}\\ \textbf{Type}}} &
\multirow{3}{*}{\textbf{Threshold $\tau$}} &
\multicolumn{5}{c}{\textbf{m=0.3, num=500}} &
\multicolumn{5}{c}{\textbf{y=0.3, num=500}} &
\multicolumn{5}{c}{\textbf{m=0.3, y=0.3}} \\

& & & \multicolumn{5}{c}{\textbf{m (Margin)}} &
\multicolumn{5}{c}{\textbf{y (Weight)}} &
\multicolumn{5}{c}{\textbf{num (Number of Input Vectors)}} \\

\cmidrule(lr){4-8} \cmidrule(lr){9-13} \cmidrule(lr){14-18}
& & & 0.1 & 0.3 & 0.5 & 0.7 & 0.9 &
0.1 & 0.3 & 0.5 & 0.7 & 0.9 &
250 & 500 & 750 & 1000 & 1250 \\
\midrule

\multirow{10}{*}{MNIST}
& \multirow{5}{*}{\textit{Target (\%)}} 
& 0.1 & 31.30 & 98.30 & 100.00 & 100.00 & 100.00 & 100.00 & 96.00 & 94.30 & 96.30 & 99.30 & 99.00 & 95.90 & 95.30 & 97.00 & 98.30 \\
& & 0.3 & 18.00 & 42.30 & 85.30 & 92.30 & 93.00 & 40.30 & 49.00 & 48.30 & 54.00 & 62.30 & 48.70 & 46.00 & 43.70 & 43.70 & 54.00 \\
& & 0.5 & 13.70 & 24.00 & 45.30 & 54.00 & 55.00 & 16.70 & 30.70 & 29.00 & 40.30 & 42.00 & 26.30 & 28.70 & 27.00 & 25.70 & 32.00 \\
& & 0.7 & 0.00 & 12.30 & 27.70 & 36.00 & 36.70 & 0.00 & 10.30 & 15.00 & 25.30 & 25.70 & 8.70 & 10.30 & 12.70 & 11.30 & 18.00 \\
& & 0.9 & 0.00 & 0.00 & 0.00 & 0.00 & 0.00 & 0.00 & 0.00 & 0.00 & 0.00 & 0.00 & 0.00 & 0.00 & 0.00 & 0.00 & 0.00 \\
\cmidrule(lr){2-18}
& \multirow{5}{*}{\textit{Untarget (\%)}} 
& 0.1 & 7.00 & 84.30 & 100.00 & 100.00 & 100.00 & 98.30 & 88.00 & 85.70 & 84.30 & 98.30 & 98.30 & 88.20 & 87.70 & 88.00 & 88.30 \\
& & 0.3 & 0.00 & 0.00 & 0.00 & 0.00 & 0.00 & 0.00 & 0.00 & 0.00 & 0.00 & 0.00 & 0.00 & 0.00 & 0.00 & 0.00 & 0.00 \\
& & 0.5 & 0.00 & 0.00 & 0.00 & 0.00 & 0.00 & 0.00 & 0.00 & 0.00 & 0.00 & 0.00 & 0.00 & 0.00 & 0.00 & 0.00 & 0.00 \\
& & 0.7 & 0.00 & 0.00 & 0.00 & 0.00 & 0.00 & 0.00 & 0.00 & 0.00 & 0.00 & 0.00 & 0.00 & 0.00 & 0.00 & 0.00 & 0.00 \\
& & 0.9 & 0.00 & 0.00 & 0.00 & 0.00 & 0.00 & 0.00 & 0.00 & 0.00 & 0.00 & 0.00 & 0.00 & 0.00 & 0.00 & 0.00 & 0.00 \\
\midrule

\multirow{10}{*}{Fashion-MNIST}
& \multirow{5}{*}{\textit{Target (\%)}} 
& 0.1 & 1.30 & 100.00 & 100.00 & 100.00 & 100.00 & 100.00 & 100.00 & 100.00 & 100.00 & 100.00 & 99.30 & 100.00 & 100.00 & 100.00 & 100.00 \\
& & 0.3 & 0.00 & 0.00 & 99.70 & 0.00 & 100.00 & 0.00 & 0.00 & 0.00 & 0.00 & 36.70 & 0.00 & 0.00 & 0.00 & 26.00 & 84.70 \\
& & 0.5 & 0.00 & 0.00 & 0.00 & 0.00 & 0.00 & 0.00 & 0.00 & 0.00 & 0.00 & 0.00 & 0.00 & 0.00 & 0.00 & 0.00 & 1.30 \\
& & 0.7 & 0.00 & 0.00 & 0.00 & 0.00 & 0.00 & 0.00 & 0.00 & 0.00 & 0.00 & 0.00 & 0.00 & 0.00 & 0.00 & 0.00 & 0.00 \\
& & 0.9 & 0.00 & 0.00 & 0.00 & 0.00 & 0.00 & 0.00 & 0.00 & 0.00 & 0.00 & 0.00 & 0.00 & 0.00 & 0.00 & 0.00 & 0.00 \\
\cmidrule(lr){2-18}
& \multirow{5}{*}{\textit{Untarget (\%)}} 
& 0.1 & 0.00 & 91.30 & 100.00 & 100.00 & 100.00 & 57.00 & 91.30 & 86.70 & 83.00 & 92.00 & 37.30 & 91.30 & 13.00 & 97.00 & 100.00 \\
& & 0.3 & 0.00 & 0.00 & 0.00 & 0.00 & 0.00 & 0.00 & 0.00 & 0.00 & 0.00 & 0.00 & 0.00 & 0.00 & 0.00 & 0.00 & 0.00 \\
& & 0.5 & 0.00 & 0.00 & 0.00 & 0.00 & 0.00 & 0.00 & 0.00 & 0.00 & 0.00 & 0.00 & 0.00 & 0.00 & 0.00 & 0.00 & 0.00 \\
& & 0.7 & 0.00 & 0.00 & 0.00 & 0.00 & 0.00 & 0.00 & 0.00 & 0.00 & 0.00 & 0.00 & 0.00 & 0.00 & 0.00 & 0.00 & 0.00 \\
& & 0.9 & 0.00 & 0.00 & 0.00 & 0.00 & 0.00 & 0.00 & 0.00 & 0.00 & 0.00 & 0.00 & 0.00 & 0.00 & 0.00 & 0.00 & 0.00 \\
\midrule

\multirow{10}{*}{CIFAR-10}
& \multirow{5}{*}{\textit{Target (\%)}} 
& 0.1 & 13.30 & 78.00 & 100.00 & 100.00 & 100.00 & 95.00 & 76.00 & 97.00 & 93.00 & 99.00 & 78.30 & 76.70 & 99.00 & 99.70 & 100.00 \\
& & 0.3 & 0.00 & 0.00 & 81.00 & 100.00 & 100.00 & 0.00 & 0.00 & 0.00 & 0.30 & 0.00 & 0.00 & 0.00 & 1.00 & 1.30 & 34.70 \\
& & 0.5 & 0.00 & 0.00 & 0.00 & 11.70 & 10.70 & 0.00 & 0.00 & 0.00 & 0.00 & 0.00 & 0.00 & 0.00 & 0.00 & 0.00 & 10.00 \\
& & 0.7 & 0.00 & 0.00 & 0.00 & 0.00 & 0.00 & 0.00 & 0.00 & 0.00 & 0.00 & 0.00 & 0.00 & 0.00 & 0.00 & 0.00 & 0.00 \\
& & 0.9 & 0.00 & 0.00 & 0.00 & 0.00 & 0.00 & 0.00 & 0.00 & 0.00 & 0.00 & 0.00 & 0.00 & 0.00 & 0.00 & 0.00 & 0.00 \\
\cmidrule(lr){2-18}
& \multirow{5}{*}{\textit{Untarget (\%)}} 
& 0.1 & 0.00 & 12.00 & 100.00 & 100.00 & 100.00 & 48.30 & 13.30 & 60.70 & 64.30 & 43.70 & 13.30 & 10.70 & 87.30 & 89.30 & 100.00 \\
& & 0.3 & 0.00 & 0.00 & 0.00 & 0.00 & 0.00 & 0.00 & 0.00 & 0.00 & 0.00 & 0.00 & 0.00 & 0.00 & 0.00 & 0.00 & 0.00 \\
& & 0.5 & 0.00 & 0.00 & 0.00 & 0.00 & 0.00 & 0.00 & 0.00 & 0.00 & 0.00 & 0.00 & 0.00 & 0.00 & 0.00 & 0.00 & 0.00 \\
& & 0.7 & 0.00 & 0.00 & 0.00 & 0.00 & 0.00 & 0.00 & 0.00 & 0.00 & 0.00 & 0.00 & 0.00 & 0.00 & 0.00 & 0.00 & 0.00 \\
& & 0.9 & 0.00 & 0.00 & 0.00 & 0.00 & 0.00 & 0.00 & 0.00 & 0.00 & 0.00 & 0.00 & 0.00 & 0.00 & 0.00 & 0.00 & 0.00 \\
\midrule

\multirow{10}{*}{CIFAR-100}
& \multirow{5}{*}{\textit{Target (\%)}} 
& 0.1 & 77.30 & 100.00 & 100.00 & 100.00 & 100.00 & 100.00 & 100.00 & 100.00 & 100.00 & 100.00 & 99.70 & 100.00 & 100.00 & 62.70 & 100.00 \\
& & 0.3 & 0.00 & 0.00 & 0.30 & 100.00 & 100.00 & 0.00 & 0.00 & 0.70 & 0.30 & 8.70 & 0.00 & 0.00 & 0.00 & 0.00 & 0.00 \\
& & 0.5 & 0.00 & 0.00 & 0.00 & 0.00 & 53.70 & 0.00 & 0.00 & 0.00 & 0.00 & 0.00 & 0.00 & 0.00 & 0.00 & 0.00 & 0.00 \\
& & 0.7 & 0.00 & 0.00 & 0.00 & 0.00 & 0.30 & 0.00 & 0.00 & 0.00 & 0.00 & 0.00 & 0.00 & 0.00 & 0.00 & 0.00 & 0.00 \\
& & 0.9 & 0.00 & 0.00 & 0.00 & 0.00 & 0.00 & 0.00 & 0.00 & 0.00 & 0.00 & 0.00 & 0.00 & 0.00 & 0.00 & 0.00 & 0.00 \\
\cmidrule(lr){2-18}
& \multirow{5}{*}{\textit{Untarget (\%)}} 
& 0.1 & 100.00 & 100.00 & 100.00 & 100.00 & 100.00 & 100.00 & 100.00 & 100.00 & 100.00 & 100.00 & 100.00 & 100.00 & 100.00 & 100.00 & 100.00 \\
& & 0.3 & 0.00 & 0.00 & 0.00 & 0.00 & 0.00 & 0.00 & 0.00 & 0.00 & 0.00 & 0.00 & 0.00 & 0.00 & 0.00 & 0.00 & 0.00 \\
& & 0.5 & 0.00 & 0.00 & 0.00 & 0.00 & 0.00 & 0.00 & 0.00 & 0.00 & 0.00 & 0.00 & 0.00 & 0.00 & 0.00 & 0.00 & 0.00 \\
& & 0.7 & 0.00 & 0.00 & 0.00 & 0.00 & 0.00 & 0.00 & 0.00 & 0.00 & 0.00 & 0.00 & 0.00 & 0.00 & 0.00 & 0.00 & 0.00 \\
& & 0.9 & 0.00 & 0.00 & 0.00 & 0.00 & 0.00 & 0.00 & 0.00 & 0.00 & 0.00 & 0.00 & 0.00 & 0.00 & 0.00 & 0.00 & 0.00 \\
\bottomrule
\end{tabular}
}
\end{table*}

\begin{table}[htbp]
\centering
\caption{Performance under vector augmentation}
\label{tab:noise}
\renewcommand{\arraystretch}{1.1} 
\setlength{\tabcolsep}{3pt}    
\resizebox{\columnwidth}{!}{
\begin{tabular}{ccccccc}
\toprule
\multirow{2}{*}{\textbf{Dataset}} & 
\multicolumn{2}{c}{\textbf{Laplace}} & 
\multicolumn{2}{c}{\textbf{Uniform}} & 
\multicolumn{2}{c}{\textbf{Gaussian (Ours)}} \\
\cmidrule(lr){2-3} \cmidrule(lr){4-5} \cmidrule(lr){6-7}
 & \textbf{Acc (\%)} & \textbf{SSIM} 
 & \textbf{Acc (\%)} & \textbf{SSIM} 
 & \textbf{Acc (\%)} & \textbf{SSIM} \\
\midrule
MNIST         & 99.51 & 0.99 & 99.55 & 0.99 & 99.36 & 0.98 \\
Fashion-MNIST & 89.12 & 0.98 & 89.08 & 0.99 & 89.70 & 0.96 \\
CIFAR-10      & 89.53 & 0.98 & 89.49 & 0.98 & 91.44 & 0.98 \\
CIFAR-100     & 69.08 & 0.91 & 68.33 & 0.91 & 68.76 & 0.93 \\
\bottomrule
\end{tabular}
}
\end{table}

\subsection{More ASR Results}
\label{app:asr}
\Cref{tab:asr} presents a comprehensive analysis of ASR under various parameter settings, including margin $m$, sample weight $y$, and input vector numbers $num$. The results indicate that ASR is highly sensitive to these hyperparameters, especially under targeted attacks. For instance, when $m \geq 0.5$, the ASR rapidly approaches 100\% on all datasets, reflecting that larger margins amplify feature separability, making forged watermarks easier to align with genuine ones. Conversely, smaller margins ($m \leq 0.3$) or higher thresholds ($\tau \geq 0.7$) substantially suppress attack success, with ASR often dropping to zero. Similarly, increasing $y$ or the number of forged vectors initially improves ASR but quickly saturates beyond moderate values, suggesting diminishing returns due to feature redundancy. Across datasets, CIFAR-100 remains the most resistant, achieving only 77.3\% ASR at $m=0.1$, $y=0.3$, and $\tau=0.1$, highlighting the robustness of high-dimensional representations. Overall, these results demonstrate that proper hyperparameter calibration is crucial for maintaining watermark integrity and reducing vulnerability to adaptive forgery attacks.

\subsection{Impact of Vector Augmentation}
We employed vector augmentation by adding noise sampled from Gaussian distributions. To examine the impact of different noise distributions on performance and watermark fidelity, we compared Gaussian, Laplace, and Uniform augmentations across four datasets. As shown in \Cref{tab:noise}, Our augmentation strategy achieves consistently high model accuracy (e.g., 91.44\% on CIFAR-10, 68.76\% on CIFAR-100) while preserving strong watermark fidelity (SSIM $\geq$ 0.93). Laplace and Uniform augmentations provide slightly higher accuracy on simpler datasets (e.g., MNIST, Fashion-MNIST) but exhibit reduced stability on more complex ones. The results indicate that differences in vector augmentation have minimal impact on performance.
 
\subsection{ClearStamp Watermark}
\label{sec:cs-wm}

\begin{figure}[t]
    \centering
    \begin{subfigure}{0.32\columnwidth}
        \centering
        \includegraphics[width=\textwidth]{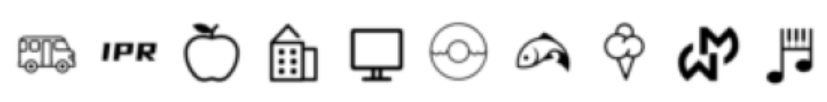}
        \caption{Original}
        \label{fig:cs-o}
    \end{subfigure}
    \begin{subfigure}{0.32\columnwidth}
        \centering
        \includegraphics[width=\textwidth]{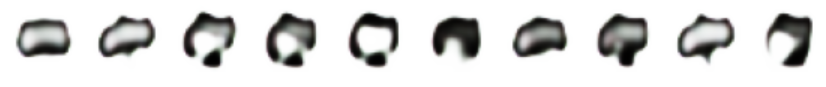}
        \caption{MNIST}
        \label{fig:cs-f}
    \end{subfigure}
    \begin{subfigure}{0.32\columnwidth}
        \centering
        \includegraphics[width=\textwidth]{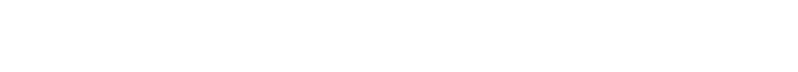}
        \caption{Other Datasets}
        \label{fig:cs-v}
    \end{subfigure}
    \caption{ClearStamp watermark in FL}
    \label{fig:cs}
\end{figure}

\Cref{fig:cs} presents the visual results of watermarks generated using the ClearStamp scheme across different datasets under various attack settings. The originally designated watermark set is shown in \Cref{fig:cs-o}, while \Cref{fig:cs-f} depicts the optimal watermarks reconstructed on the MNIST dataset, and \Cref{fig:cs-v} displays results for other datasets. The reconstructed watermark on MNIST fails to preserve recognizable structural or stylistic features. Moreover, due to dataset heterogeneity, the watermarks produced on other datasets appear completely unstructured, suggesting that ClearStamp fails to extract or verify meaningful watermark information in federated settings. Please note that, as shown in \Cref{fig:comp-ma}, ClearStamp achieves approximately 40\% watermark accuracy not because the watermark is successfully constructed, but rather because the original watermark images contain white-background regions. ClearStamp inadvertently reproduces similar white areas, leading to artificially inflated WM-Acc scores even though the watermark itself is not actually encoded.

\end{document}